\documentclass{article}

\usepackage{hyperref}

\usepackage{amsmath}
\usepackage{amssymb}
\usepackage{mathtools}
\usepackage{amsthm}
\usepackage{xurl} % To break urls when they exceed the line length

\usepackage[accepted]{icml2023}

\usepackage[capitalize,noabbrev]{cleveref}

\theoremstyle{plain}

\theoremstyle{definition}

\theoremstyle{remark}

\usepackage[textsize=tiny]{todonotes}

\usepackage{microtype}
\usepackage{booktabs} 
\usepackage{subcaption}
\usepackage{latexsym}
\usepackage{graphicx}
\usepackage{multicol,multirow}
\usepackage{rotating}
\usepackage{appendix}
\usepackage{ifpdf}
\usepackage[T1]{fontenc}
\usepackage{times}
\usepackage{sourcesanspro}
\usepackage{newtxmath}
\usepackage{textcomp}
\usepackage{xcolor}
\usepackage{bm}
\usepackage{placeins}
\usepackage{xcolor}
\usepackage{tikz}
\usetikzlibrary{spy}
\usetikzlibrary{bayesnet}
\usetikzlibrary{arrows}
\usepackage{makecell}
\UseRawInputEncoding

\begin{document}

\twocolumn[
\icmltitle{Disentangled Generative Models for Robust Prediction of System Dynamics}



\icmlsetsymbol{equal}{*}

\begin{icmlauthorlist}
\icmlauthor{Stathi Fotiadis}{bioeng}
\icmlauthor{Mario Lino}{aero}
\icmlauthor{Shunlong Hu}{bioeng}
\icmlauthor{Stef Garasto}{greenwich}
\icmlauthor{Chris D Cantwell}{aero}
\icmlauthor{Anil Anthony Bharath }{bioeng}
\end{icmlauthorlist}
\icmlaffiliation{bioeng}{Department of Bioengineering, Imperial College London, UK}
\icmlaffiliation{aero}{Department of Aeronautics, Imperial College London, UK}
\icmlaffiliation{greenwich}{School of Computing and Mathematical Sciences, University of Greenwich, London, UK}

% \begin{icmlauthorlist}
% \icmlauthor{Stathi Fotiadis}{icl}
% \icmlauthor{Mario Lino Valencia}{icl}
% \icmlauthor{Shunlong Hu}{icl}
% \icmlauthor{Stef Garasto}{greenwich}
% \icmlauthor{Chris D Cantwell}{icl}
% \icmlauthor{Anil Anthony Bharath }{icl}
% \end{icmlauthorlist}
% \icmlaffiliation{icl}{Imperial College London, UK}
% \icmlaffiliation{greenwich}{University of Greenwich, London, UK}

\icmlcorrespondingauthor{Stathi Fotiadis}{s.fotiadis19@imperial.ac.uk}

\icmlkeywords{Machine Learning, ICML, dynamical systems, prediction, forecasting, generative models, vae, system dynamics}
\vskip 0.3in
]


\printAffiliationsAndNotice{}  % leave blank if no need to mention equal contribution

\begin{abstract}
The use of deep neural networks for modelling system dynamics is increasingly popular, but long-term prediction accuracy and out-of-distribution generalization still present challenges. In this study, we address these challenges by considering the parameters of dynamical systems as factors of variation of the data and leverage their ground-truth values to disentangle the representations learned by generative models. Our experimental results in phase-space and observation-space dynamics, demonstrate the effectiveness of latent-space supervision in producing disentangled representations, leading to improved long-term prediction accuracy and out-of-distribution robustness.

%In this work, we treat the parameters of dynamical systems as factors of variation in the data and use the ground-truth values of those parameters to disentangle the representations of generative models. Our experiments in phase-space and observation-space dynamics indicate that latent-space supervision can effectively produce disentangled model representations, leading to more accurate long-term prediction both in- and out-of-distribution.

\end{abstract}

\section{Introduction}
The robust prediction of the behaviour of dynamical systems remains an open question in machine learning, and engineering in general. The ability to make robust predictions is important not only for forecasting systems of interest like weather \citep{garg2021weatherbench}, but also because it supports innovations in fields like system control, autonomous planning \citep{Hafner2018LearningPixels} and computer-aided engineering \citep{Brunton2020MachineMechanics}. In this context, the use of deep generative models has recently gained significant traction for sequence modelling \citep{Girin2020DynamicalReview}. The robustness of machine learning models can be considered along two axes: (1) long-term and (2) out-of-distribution (OOD) performance. Accurate long-term prediction can be notoriously difficult in many dynamical systems because error accumulation can cause divergence in finite time \citep{Zhou2020Informer:Forecasting, Raissi2019Physics-informedEquations}, a problem that even traditional solvers can suffer from. At the same time, machine learning techniques are known to suffer from poor OOD performance \citep{Goyal2020InductiveCognition}, when they are employed in a setting they had not encountered in their training phase. 

When it comes to modelling dynamics, training data  contain both the dynamics themselves and the dynamical system parameters. However, many approaches to learning fail to distinguish between the two, which could result in entangled representations, leading to overfitting and thus poorer forecasts \citep{bengio2013}. Here, our aim is to investigate generative models whose latent space is disentangled in such a way that the parameters and the dynamics are distinctly represented. More specifically, we explore dynamical systems modelled by ordinary differential equations (ODEs) and their respective parameters.

Our method builds on two elements. First, the inherent ability of Variational Autoencoders (VAEs) \cite{Kingma2014Auto-encodingBayes} to produce disentangled representations in an unsupervised way \cite{Higgins2017Beta-VAE:FRAMEWORK}, a feature that has been applied in the context of image and scene modelling \citep{Kim2018DisentanglingFactorising}. Second, latent space supervision with ground-truth factors has been found to produce more disentangled representations in image modelling \cite{Locatello2019DisentanglingLabels}. We motivate the use of disentangled representations through a theoretical analysis of the emission process through the scope of dynamical systems. In practice, we treat the parameters of a dynamical system as factors of variation of the data distribution and use the ground-truth values of these parameters to improve the latent space disentanglement. While various assumptions, like domain stationarity, have been used to improve the disentanglement in the prediction of dynamical systems in an unsupervised way \citep{Li2018DisentangledAutoencoder, Miladinovic2019DisentangledRepresentations}, to the best of our knowledge, this is the first attempt to use supervised disentanglement for system dynamics. 
Furthermore, contrary to system-identification techniques that require knowledge of the full underlying system to be computationally effective \citep{ayad2019systid}, our technique only needs to be aware of the system parameters.

\textbf{Contributions}
Our work is the first, to the best of our knowledge, that uses ground-truth information of the dynamical system parameters to disentangle the latent space of generative models. We provide a theoretical motivation for disentangled representations in dynamical system prediction and, practically, encourage latent space disentanglement through supervision. We conduct experiments with VAEs trained with noisy observations of the phase-space of 3 dynamical systems. We also apply our method to a state-of-the-art generative model (Recurrent State Space Model \cite{Hafner2018LearningPixels}) trained on image sequences of a swinging pendulum. We propose a definition of OOD data in the context of system dynamics and evaluate the performance of models in- and out-of-distribution. We demonstrate that models with a disentangled latent space can better capture the variability of dynamical systems and produce more accurate long-term predictions, both in- and out-of-distribution. %Our method is applicable to problems where synthetic data can be used for training, such as computer-aided engineering and robotics. 
However, the practical importance of our method is currently limited by the labelling cost. It would be worth assessing the model in the semi-supervised setting, as that would be better suited for real-world application. All the necessary code to reproduce our experiments is provided at \url{https://github.com/stathius/sd-vae}.

\section{Related Work}
 
\textbf{VAEs and disentanglement.} Disentanglement aims to produce representations where each latent variable captures a different factor of variation of the data distribution. This can also be seen as identifying the true causal model of the data-generating process \cite{Scholkopf2019CausalityLearning}. While supervised disentanglement is a long-standing idea \citep{Mathieu2016DisentanglingTraining}, information-theoretic properties can be leveraged to allow unsupervised disentanglement in VAEs  \citep{Higgins2017Beta-VAE:FRAMEWORK, Kim2018DisentanglingFactorising}. Recent findings \citep{Locatello2020AEvaluation} emphasize the vital role of inductive biases from models or data for useful disentanglement, leading to semi- and weakly-supervised disentanglement approaches \citep{Locatello2019DisentanglingLabels, Locatello2020Weakly-SupervisedCompromises}. In the field of physical sciences, hierarchical priors have been proposed to learn disentangled representations of high-dimensional spatial fields \citep{Jacobsen2022DisentanglingAutoencoders}. To assess the strength of disentanglement, simulated datasets are usually used, because simulations give access to the ground-truth factors of variation (i.e., color or shape of an object in image data). Various metrics have been proposed to quantify disentanglement, both predictor-based \cite{Eastwood2018AREPRESENTATIONS, Kumar2017VariationalObservations} and information-theoretic ones \cite{Chen2018IsolatingVAEs} but the task still presents challenges \cite{Carbonneau2020MeasuringMetrics}. 

\textbf{Disentanglement in sequence modelling.} While disentanglement methods are often tested in a static (image) setting, there is a growing interest in applying disentanglement to sequence dynamics. Using a bottleneck corresponding to the degrees of freedom of the physical system, \citet{Iten2018DiscoveringNetworks} learn an interpretable representation using a VAE. However, their model gives physically inconsistent predictions in OOD data \citep{Barber2021JointSystems}. Disentangling content from dynamics has also been tried in deep state-space models (SSMs) \citep{Fraccaro2017ALearning, Li2018DisentangledAutoencoder}, but these methods focus mostly on modelling variations in the appearance of moving objects, failing to take dynamics into account. Unsupervised techniques have also been proposed. Assuming domain stationarity, \citet{Miladinovic2019DisentangledRepresentations} separate the dynamics from sequence-wide properties in dynamical systems like Lotka-Volterra but they do not fully evaluate the OOD performance of their model. \citet{yeo2021variational} suggest that learning hierarchy of semantic concepts leads to feature abstraction and enhanced disentanglement, while \cite{Li2023GenerativeDisentanglement} propose a model for time-series generation whose representation is disentangled by minimizing the pairwise total correlation between the latent variables. While unsupervised methods have their advantages, they also dismiss a wealth of information that can be cheaply collected from simulated data, a gap that our method tries to fill. 

\textbf{VAEs for sequence modelling.} Dynamical VAEs \cite{Girin2020DynamicalReview} have long been used to model sequence dynamics. Combining VAEs with physics-informed neural networks \citep{Raissi2019Physics-informedEquations} can also be used to model stochastic differential equations \citep{Zhong2022PI-VAE:Equations}. Feed-forward VAEs have also attracted a lot of interest in modelling physical systems. There are two main motivations for this. First, VAEs offer various ways to incorporate the inductive biases obtained from prior knowledge of the physical system. Second, since their latent space is relatively simple, one can easily assess if those inductive biases result in more interpretable representations. Methods to incorporate inductive biases include (i) bottlenecks based on the degrees of freedom of the physical system \cite{Iten2018DiscoveringNetworks}, (ii) the use of geometric and topological information of the dynamical system responses to shape the manifold of the latent representations \cite{Lopez2022GD-VAEs:Reductions}, and (iii) using physics-informed priors \citep{Takeishi2021Physics-IntegratedModeling}. Furthermore, feed-forward VAEs can be combined with recurrent neural networks (RNN) to improve accuracy  while at the same time learning highly-disentangled representations of dynamical systems \citep{yeo2021variational}.

\section{Problem formulation}
\subsection{Dynamical systems}
\label{sec:dynamical_systems_definition}
Let $\textbf{u} \in \mathbb{R}^d$ be the state of a system. We consider system dynamics that are governed by a set of differential equations (DEs):
\begin{equation}
\frac{d\mathbf{u}}{dt} = \mathcal{F}(\textbf{u}, \boldsymbol\xi)
\label{eq:governing_equations}
\end{equation}
where $\mathcal{F}$ describes the governing equations and $\boldsymbol\xi \in \mathbb{R}^{N_\xi}$ denotes the parameters of these DEs. While these equations describe how the system state evolves over time, there is a limited number of real-world problems where they can be solved analytically. Hence, most often, the time evolution of a system is acquired by numerical methods, given the governing equations and some initial state. In experimental data, observations $\mathbf{\tilde{u}}_t$ contain some noise: $\mathbf{\tilde{u}}_t = \textbf{u}_t  + {\boldsymbol\epsilon}_t$,  where  ${\boldsymbol\epsilon}_t \in \mathbb{R}^d$ is the stochastic uncorrelated noise. In our computational experiments, the data are corrupted with white Gaussian noise to simulate observation noise.
%In general long term prediction in non-trivial. As numerical errors accrue at some point predictions diverge from the ground truth, it is a matter of when, not if.

In the experimental section, we are concerned with dynamical systems governed by Ordinary DEs (ODEs) but our methods could in principle apply to Partial DEs as well. The three dynamical systems we study are the swinging pendulum, the Lotka-Volterra system used to model prey-predator populations, and the planar 3-body system. The governing equations are the following:
\begin{align} 
\textbf{Simple pendulum: } &\ddot{\theta} = - \frac{g}{\ell} \sin \theta \\
\textbf{Lotka-Volterra: } &\dot{x}=\alpha x-\beta x y  \nonumber\\ 
&\dot{y}=\delta x y-\gamma y  \\
\textbf{3-body system: } & \dot{\textbf{v}}_{i}= \frac{K_{1}}{m_{i}} \sum_{j} \frac{ m_{i} m_{j}}{|\textbf{r}_{i j}|^3} \textbf{r}_{i j} \nonumber \\ &\dot{\textbf{x}}_{i}=K_{2} \textbf{v}_{i} \\ 
& \textbf{v}_{i}, \textbf{x}_{i} \in \mathbb{R}^2, i\in[1,2,3] \nonumber
\end{align}

Where $\theta$ is the length of the pendulum, $g$ is the acceleration due to gravity and $\ell$ is its length. Since the two parameters appear in ratio we keep gravity constant and only vary the length of the pendulum, i.e., $\boldsymbol\xi=[\ell]$. In Lotka-Volterra, $x,y$ are the prey and predator populations while the 4 parameters $\boldsymbol\xi=[\alpha, \beta, \gamma, \delta]$ describe the interaction of the two species.
In the 3-body, $\textbf{x}_i , \textbf{v}_i$ are the positions and velocities of the bodies and the 4 parameters $\boldsymbol\xi=[K_1, m_1, m_2, m_3]$ represent the gravity constant and masses. Overall, these systems are characterized by a varied number of degrees of freedom, governing equations and number of parameters. We also refer to \cref{sec:app:theoretical} for more details.

\subsection{Theoretical motivation for disentanglement}

The problem setup that involves inferring the evolution of a system state, up to some time in future, $t+n$, given a number of previous (observed) states up to a point in time $t$. The system dynamics are defined by the form of the differential equation (DE), the parameters of it $\bm\xi$, and the initial conditions $\bm{I}$. Since the DE is deterministic, if parameters and current state is known then next step can be calculated using numerical methods with a high precision (bound by the numerical precision of the computational method). We can consider the simplified setting where the conditional distribution of the next state $P(\mathbf{x}_{t:t+n} | \mathbf{x}_{<t}; \bm\xi_C, \bm{I}_C)$ is characterized only by the noise of the observations, assuming there is no other type of uncertainty. In the absence of noise the distribution becomes Dirac's delta function. In practice we often do not have access to $\bm\xi, \bm{I}$. There are two options in this case a) assign priors on $\bm\xi_C, \bm{I}_C$, and marginalize over them to obtain an estimate of the marginal, and b) estimate $\bm\xi_C$ and $\bm{I}_C$ and directly model the conditional. Given the wide nature of divergence in the trajectories of a system for different $\bm\xi_C, \bm{I}_C$, it is hard to both  assign a proper prior and efficiently marginalize. On the other hand if we can derive good estimates for $\bm\xi_C, \bm{I}_C$, then the second modelling choice becomes more appealing and this is where disentanglement can be beneficial. For simplicity, we consider the pendulum where its length $\bm\xi_C=l$ varies between trajectories, and all other parameters and initial conditions are constant and known. In this case, the marginal $P(\mathbf{x}_{t:t+n}|\mathbf{x}_{<t})$ remains unknown, the conditional, $P(\mathbf{x}_{t:t+n}|\mathbf{x}_{<t}, l)$ is just characterized by the observational noise, as described earlier. In VAEs this procedure is modelled by the decoder as $P( \mathbf{x}_{t+n}| \mathbf{z}_{<t})$. Disentanglement allows the separation of the latent vector in two parts i) $\mathbf{z}_{<t}$ that captures the dynamics and ii) $\mathbf{z}_l$ that encodes the pendulum length. This leads to a conditional distribution $P(\mathbf{x}_{t:t+n}| \mathbf{z}_{<t}, \mathbf{z}_l)$ which better resembles the functional structure of the real conditional distribution above.% Assuming that the model is able to capture well the deterministic dynamics this should be a better modelling choice and increase prediction performance.

\begin{figure}
    \centering
\includegraphics[width=1\linewidth]{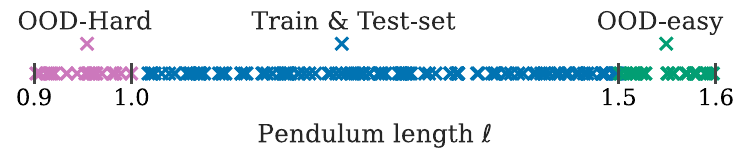}
    \caption{Samples from the parameter distributions of pendulum length $\ell$. Each trajectory in the train, validation \& test sets is simulated with length drawn from $\ell \sim \mathcal{U} (1.0, 1.5)$, while OOD-Easy and OOD-Hard have disjoint distributions. Note that predicting the trajectory of a shorter pendulum is harder for our models because it swings faster.}
    \label{fig:pendulum_length_distribution}
\end{figure}

\subsection{Definition of OOD dataset}

The evolution of a dynamical system is defined by far and foremost by the form of $\mathcal{F}$ in \cref{eq:governing_equations}. Considering $\mathcal{F}$ given, the next most important factor that characterizes the distribution of trajectories in the system is the values of the parameters $\boldsymbol\xi$. If these parameters come from a distribution $\boldsymbol\xi \sim P(\boldsymbol\xi)$, the trajectories of states that the system can follow will form another distribution $
\textbf{u}_{\leq t} \sim P(\textbf{u}_{\leq t} \mid \bm{\xi})$, where $\textbf{u}_{\leq t}$ is the evolution of the system states from the start-up to time $t$. Given the nature of dynamical systems, different parameters can produce widely different trajectories in state space \cite{lai1994extreme} so it is reasonable to assume that changes in the parameter distribution will affect the trajectory distribution as well. For our systems, we additionally verify this by visually inspecting the trajectories produced by each parameter distribution (see \cref{sec:dataset_phase_space}). Here, we define an OOD dataset as a dataset comprising a set of trajectories derived from a parameter distribution that is disjoint from the one used to generate the trajectories of the training dataset. For each system, we draw the parameters $\boldsymbol\xi$ from a uniform distribution which is the same for the training, validation and test sets. These three datasets are considered in-distribution. Furthermore, we create two additional datasets using different parameter distributions. The support of these distributions is disjoint from the previous distribution and with each other. We name these datasets OOD-Easy and OOD-Hard. \cref{fig:pendulum_length_distribution} illustrates the distribution of lengths of the pendulum datasets.

Capturing the whole distribution of trajectories in a single training set is unrealistic \citep{Fotiadis2020ComparingPropagation} and for learning models with robust OOD prediction, some extra inductive biases are needed \citep{Bird2019CustomizingSystems, Barber2021JointSystems}. In our method, this inductive bias comes by leveraging the ground-truth parameters to disentangle the latent representation. For the observation-space pendulum experiments, we extend the notion of parameters to include the initial conditions (boundary conditions could also be added). A detailed description of the datasets is provided in \cref{tab:datasets} of the Appendix.

\begin{figure}[t]
\centering
  \includegraphics[width=0.95\linewidth]{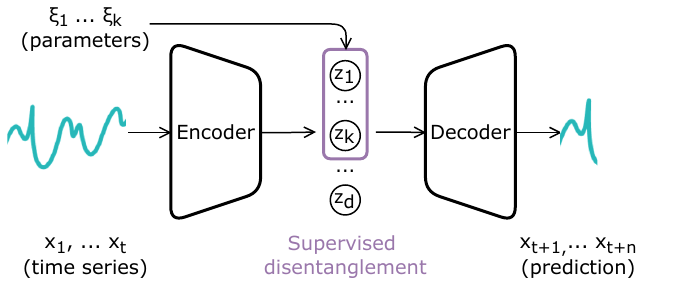}
    \caption{\textbf{The SD-VAE model} Taking as input $t$ observations of the phase space, the model predicts the state in future time-steps in one pass. The model is trained with an additional loss term over some of the latent, which makes the representation more disentangled.}
\label{fig:model_sdvae}
\end{figure}

\section{Methods}

\subsection{Variational Autoencoders (VAEs)}
VAEs \citep{Kingma2014Auto-encodingBayes} offer a principled approach to latent variable modeling. It combines an encoder $Q_{\phi}(\bm{z}|\bm{x})$ which takes the data $x$ as input and infers the latent representation $z$, with a generative decoder $P_{\theta}(\bm{x}|\bm{z})$ that project the representation back to the data space. The encoder and decoder are parameterized by neural networks which makes the computation of the marginal likelihood prohibitively expensive. Training is, thus, done with approximate inference, i.e., maximizing the evidence lower bound (ELBO) of the marginal over the data.
\begin{equation}
\label{eq:vae_loss}
\begin{aligned} 
\mathcal{L}_{\phi, \theta}(\mathbf{x})&=\mathbb{E}_{Q_{\phi}(\mathbf{z} \mid \mathbf{x})}\left[\log P_{\theta}(\mathbf{x} \mid \mathbf{z})\right]\\
&-D_{KL}\left(Q_{\phi}(\mathbf{z} \mid \mathbf{x})|| P(\mathbf{z})\right)
\end{aligned}
\end{equation}
In the standard formulation, the ELBO consists of a reconstruction loss and the Kullback-Leibler divergence between the  posterior distribution $Q_{\phi}(\mathbf{z} \mid \mathbf{x})$ and a prior  $P(\textbf{z)}$.

 \begin{figure}[t]
\centering
  \includegraphics[width=0.85\linewidth]{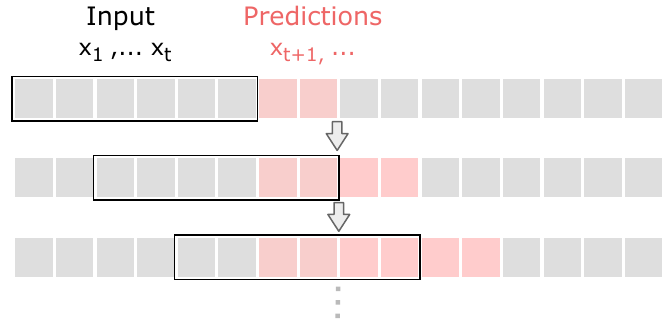}
    \caption{\textbf{Autoregressive prediction} By using model autoregressively on its own predictions we can derive an arbitrarily long horizon.}
\label{fig:autoregressive}
\end{figure}

\subsection{Disentangling VAEs for modelling dynamics}
\label{sec:methods_vaes_for_modelling_dynamics}
In theory, disentanglement in VAEs can be also achieved in an unsupervised way. Choices include using a prior with uncorrelated variables like the standard normal, adding  a weighting factor on the KL divergence term of the loss \cite{Higgins2017SCAN:Concepts} or constraining the size of the latent space to coincide with the factors of variation in the data \cite{Iten2020DiscoveringNetworks}. As \citet{Locatello2020AEvaluation} has shown, unsupervised disentanglement only works if there are biases in the data to exploit. Supervised disentanglement is possible when information about the factors of variation of the data is available. Using the ground-truth values of simulated images to disentangle VAE representations improves image generation quality \cite{Locatello2019DisentanglingLabels}. 

We extend the idea of supervised disentanglement to the context of modelling dynamics. In our datasets, each trajectory is accompanied by the parameters $\boldsymbol\xi$ of the ODE that were used to produce it. We treat those parameters as factors of variation in the data and use them to enforce a structure on the latent space using constrained optimization. Under the Karush-Kuhn-Tucker conditions, we can rewrite the constraint in the Langragian form and obtain the regularization term $\mathcal{L}_{\bm{\xi}}(\mathbf{z}_{1:N_{\bm{\xi}}}, \bm{\xi})$, between the ground truth parameters $\bm{\xi} \in \mathbb{R}^{N_{\xi}}$ and the output of the first $N_{\xi}$ latents of the VAE,  $\mathbf{z}_{1:N_{\xi}}$. We discuss the choice of the regression term $\mathcal{L}_{\bm{\xi}}$ in \cref{sec:experiments_sdvae_vs_vae}. Extending the original VAE to generate predictions instead of reconstructions, is also needed. To accommodate for this, we change the reconstruction term in \cref{eq:vae_loss} to a prediction term $\log P_{\theta}\left(\mathbf{x}_{t<,\leq t+n}\mid\mathbf{z}\right)$, leading to the the final training objective:

\begin{equation}
\label{eq:sdvae_loss}
\begin{aligned} 
\mathcal{L}_{\phi, \theta}(\mathbf{x_{\leq t}})&= \mathbb{E}_{Q_{\phi}(\mathbf{z} \mid \mathbf{x}_{\leq t})}\left[\log P_{\theta}(\mathbf{x}_{t<,\leq t+n}\mid \mathbf{z})\right. \\
&+ \delta \left.\mathcal{L}_{\bm{\xi}}(\mathbf{z}_{1:N_{\bm{\xi}}}, \bm{\xi})\right] \\
&-\beta D_{KL}\left(Q_{\phi}(\mathbf{z} \mid \mathbf{x}_{\leq t})|| P(\mathbf{z})\right)
\end{aligned}
\end{equation}

Where $t$ is the length of the input and $n$ of the predicted output and we drop the dependence of $\textbf{z}$ on $\textbf{t}$ to simplify the notation. To allow more flexibility between prediction and disentanglement, both the KLD and regression terms are weighted by tunable parameters ($\beta$ and $\delta$ respectively). Weighting the prediction term can also be seen as tuning the decoder variance. A more elaborate derivation of the objective can be found in \cref{sec:vae_loss}  We refer to this model as \textbf{SD-VAE}. A schematic of the architecture can be seen in \cref{fig:model_sdvae}.

Both the VAE and SD-VAE can produce arbitrarily-long predictions by re-feeding the model predictions back as input (\cref{fig:autoregressive}). This autoregressive approach has been shown to work well in problems like wave propagation and weather forecasting \citep{Fotiadis2020ComparingPropagation, lam2022graphcast}.

\subsection{Disentanglement of dynamics in observation-space}\label{sec:methods-rssm}

We investigate how disentanglement affects modelling of dynamics when the state of the system is not accessible directly but it inferred from high-dimensional observations like image sequences. 
In this case, the state of the system $\textbf{u}_t \in R^d$ is mapped to a high-dimensional rendering $\textbf{f}_t$. In our dataset $\textbf{f}_t \in \mathbb{R}_{\geq0}^{64\times 64}$ and $t\in \mathbb{N}$. The model for this dataset is the Recurrent State Space Model (RSSM) \citep{Hafner2018LearningPixels}. RSSM has been successfully used for planning from pixels and is considered state-of-the-art model in long-term spatiotemporal prediction \citep{Saxena2021ClockworkAutoencoders}. Furthermore, RSSM is a hybrid model combining deterministic and stochastic components, and this allows us to assess disentanglement outside VAEs. We use the same formulation of the loss function as in the original paper, with the addition of the supervised disentanglement loss, similarly to what we do in \cref{eq:sdvae_loss}. Since the RSSM has latent variables for each time-step, we apply a disentanglement loss on all of them. The SD-RSSM loss function can be found in \cref{sec:sd-rssm-loss}.

\section{Disentangling for system dynamics}
\label{sec:experiments_sdvae_vs_vae}
In this section we compare models trained to predict the evolution of dynamical systems. The main goal of our experiments is to assess whether supervised disentanglement of VAEs  improves the prediction accuracy and if the improvement also transfers to OOD data. To achieve this we compare VAEs with our proposed SD-VAE. Additionally, using quantitative and qualitative approaches, we analyze how latent space supervision affects the representation of VAEs. Next, we try to see if supervised disentanglement works also in deterministic autoencoders (AEs). Lastly, we conduct experiments with LSTMs, a popular recurrent method for low dimensional sequence modelling \citep{yu2020lstm}. Overall, we train and compare VAE, SD-VAE, AE, SD-AE and LSTM models.

\subsection{Datasets}
\label{sec:experiments_sdvae_vs_vae_datasets}
To create the datasets, we use an adaptive Runge-Kutta integrator with a timestep of $0.01$ seconds. For every simulated sequence we draw a different combination of parameters. For the pendulum simulations we randomly draw the initial angle $\theta$ from a uniform distribution $10^{\circ}-170^{\circ}$, the angular velocity $\omega$ is always $0$. For the Lotka-Volterra and 3-body system, the initial conditions are always the same to avoid pathological trajectories. Dataset details can be found in \cref{sec:dataset_phase_space}.

\begin{figure*}[ht]
  \centering
\begin{subfigure}[t]{\linewidth}
  \includegraphics[width=\linewidth]{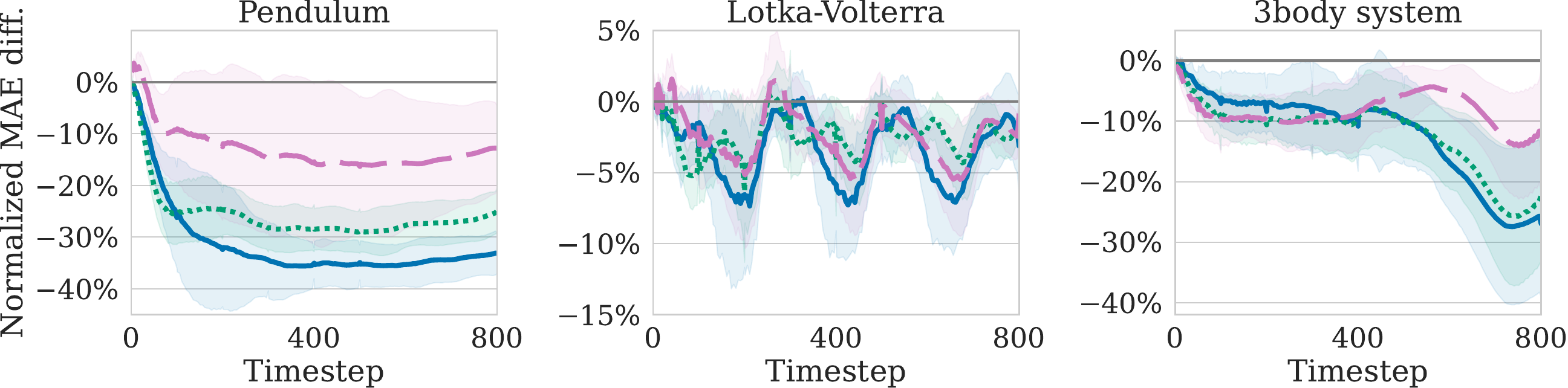}
  \end{subfigure}
% \vspace{-3.07\baselineskip}
    \begin{subfigure}[t]{0.55\linewidth}
  \includegraphics[width=\linewidth]{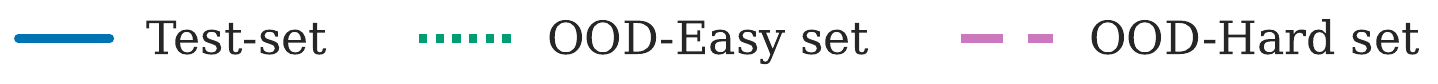}
\end{subfigure}

  \caption{\textbf{Disentangled VAE (SD-VAE) vs VAE}. Normalized MAE difference between the SD-VAE and the plain VAE. Negative values indicate that SD-VAE has lower error. We plot the mean and one standard deviation interval of the best 5 models (selected based on the cummulative MAE over a validation set)}
  \label{fig:mae-vae_vs_sdvae}
\end{figure*}

%%%%%% TABLES
\begin{table*}[ht]
\centering
\caption{\textbf{MAE averaged over 800 steps.} Mean of the best 5 models that were selected by validation MAE. SD-VAE outperforms VAE and the other models. LSTM diverged during testing on Lotka-Volterra.}
\label{tab:results_mae_all_models}
\begin{tabular}{lccc|ccc|ccc}
\toprule
{} & \multicolumn{3}{c}{Pendulum} & \multicolumn{3}{c}{Lotka-Volterra} & \multicolumn{3}{c}{3-body system} \\
{} &          Test-set & \thead{OOD-Easy} & \thead{OOD-Hard} &          Test-set & \thead{OOD-Easy} & \thead{OOD-Hard} &          Test-set & \thead{OOD-Easy} & \thead{OOD-Hard} \\
\midrule
LSTM   &           $0.829$ &           $1.318$ &           $1.985$ &             $-$ &             $-$ &             $-$ &           $0.061$ &           $0.082$ &           $0.099$ \\
MLP    &           $0.635$ &           $1.097$ &           $1.420$ &           $0.103$ &           $0.140$ &           $0.157$ &           $0.064$ &           $0.079$ &           $0.093$ \\
SD-MLP &           $0.687$ &           $1.088$ &           $1.442$ &           $0.104$ &           $0.141$ &           $0.157$ &           $0.053$ &           $0.067$ &           $0.084$ \\
VAE    &           $0.673$ &           $1.128$ &           $1.386$ &           $0.104$ &           $0.142$ &           $0.159$ &           $0.060$ &           $0.075$ &           $0.089$ \\
SD-VAE &  $\mathbf{0.443}$ &  $\mathbf{0.819}$ &  $\mathbf{1.185}$ &  $\mathbf{0.100}$ &  $\mathbf{0.138}$ &  $\mathbf{0.155}$ &  $\mathbf{0.048}$ &  $\mathbf{0.062}$ &  $\mathbf{0.080}$ \\
\bottomrule
\end{tabular}
\end{table*}

\subsection{Models and training}

\textbf{Choices for the VAE models} We use the same model choices for both VAE and SD-VAE. Our prior is an isotropic Gaussian  $P(\bm{z}) =\mathcal{N}(\bm{z} \mid \mathbf{0}, \bm{I})$ which helps to disentangle the learned representation \citep{Higgins2017Beta-VAE:FRAMEWORK}. To get a closed form KL-divergence term, we use a Gaussian with diagonal covariance as the approximate posterior distribution $q_{\phi}(\bm{z} \mid \bm{x}) =\mathcal{N}\left(\bm{z} \mid \bm{\mu}_{z}, \bm{\Sigma}_{z}\right)$, a common practical choice \cite{Kingma2014Auto-encodingBayes}. The decoder has a Laplace distribution $p_{\theta}(\bm{x} \mid \bm{z})=\mathrm{Laplace}\left(\bm{x} \mid \bm{\mu}_{x}, \bm{I}\right)$ which is equivalent to using a $L_1$ prediction loss. Preliminary experiments showed that $L_1$ loss works better than $L_2$. This is not unexpected, since $L_1$ is known to provide crisper results in image modelling \citep{Mathieu2018DisentanglingAutoencoders} and has also been used in time-series forecasting \cite{tang2021probanalysis}.  The covariance of the decoder is constant and isotropic. No scaling of the covariance is needed since we weight the KLD term in \cref{eq:sdvae_loss}.
% The parameters $\bm{\mu}_{z} \equiv \bm{\mu}_{z}(\bm{x} ; \phi), \bm{\Sigma}_{z} \equiv \operatorname{diag}\left[\bm{\sigma}_{z}(\bm{x} ; \phi)\right]^{2}$, and $\bm{\mu}_{x} \equiv \bm{\mu}_{x}(\bm{z} ; \theta)$ are computed via feed-forward neural networks. 

\textbf{Choices for the supervised disentanglement term} For the regression loss we chose the $L_1$ loss, corresponding to a Laplacian prior with mean $\bm{\xi}_i$ and unitary covariance. This choice was driven by preliminary experiments with various loss functions. Using a standard normal prior pulls the latents to be close to $0$ but this comes at odds with the disentanglement loss term because the target parameters can have larger values. To alleviate this issue we scale the parameters  $\bm{\xi} \in [0,1]$. We find that linear scaling offers some small improvement and we use it throughout our  experiments (details in \cref{sec:scaling_parameters}).

\textbf{Training } Early experiments revealed significant variance in the performance of the models, depending on hyperparameters. With this in mind, we take various steps to make model comparisons as fair as possible. Firstly, all models have similar capacity of neurons. Both the VAE and AE have an encoder with two hidden layers of sizes 400 and 200 respectively and a reverse decoder. The LSTM model has two stacked LSTM cells with a hidden size of 100, which results in an equivalent number of learned parameters. We tune the hyperparameters of each method using grid-search and train the same number of models for each method to avoid favouring one over the others by chance. To further reduce statistical chance, we conduct large-scale experiments training overall 1200 models which required more than $5,000$ CPU-hours. Details for the hyperparameters and number of experiments can be found in \cref{sec:hyperparameters_phase_space}.

\subsection{Long-term and OOD prediction accuracy}

We compare the prediction accuracy of VAE and SD-VAE on the three dynamical systems described in \cref{sec:dynamical_systems_definition} and for each system we compare on three datasets: the in-distribution test-set, which shares the same parameter distribution with the training set, and the OOD-Easy and OOD-Hard sets which represent an increasing distribution shift from the training data. Models are compared using the Mean Absolute Error (MAE) between prediction and ground truth, a widely used metrics for sequence prediction problems \cite{Girin2020DynamicalReview}, that was also used for training. Models are used in an autoregressive manner (\cref{sec:methods_vaes_for_modelling_dynamics}) to produce long-term predictions of 800 timesteps. We consider this to be sufficiently long-term since it is 20 times longer than the output of a forward pass. We predict up to 800 timesteps because the simulated trajectories are of 1000 steps long and we reserve the first 200 timesteps to randomly select a starting point for the input. To account for the variability in model training, we provide the mean and standard deviation computed for the 3 best models of each method. The best models are selected based on average validation MAE.

Results (\cref{fig:mae-vae_vs_sdvae} \& \cref{tab:results_mae_all_models}) indicate that SD-VAE offers a substantial and consistent improvement over the VAE across all 3 dynamical systems and datasets with a reduction in error that surpasses $30\%$ in the pendulum system. In both the pendulum and 3-body system the improvement is mostly increasing for long-term predictions indicating that SD-VAE captures better the system dynamics. While the accuracy of both models deteriorates in the OOD-Easy and OOD-Hard set (details in  \cref{fig:results_mae_separate} of the Appendix), SD-VAE still outperforms the VAE. This is an  indication that the disentanglement of domain parameters can be a useful inductive bias for OOD generalization. Overall, results show that SD-VAE forecasts more accurately both long-term and OOD, indicating that supervised disentanglement helps the model to better capture the system dynamics.

\begin{table*}[t]
\caption{\textbf{SD-VAE exhibits stronger disentanglement properties than the plain VAE according to widely used metrics}. Scores are averages over the best 3 models (selected by validation accuracy).}
\label{tab:resunts_disentanglement_metrics}
\centering
\begin{tabular}{lccc|cc|cc}
\toprule
    & {} & \multicolumn{2}{c}{Pendulum} & \multicolumn{2}{c}{Lotka-Volterra} & \multicolumn{2}{c}{3 body system} \\
    & {} &                VAE &             SD-VAE &                VAE &             SD-VAE &                VAE &             SD-VAE \\
\midrule
Disentanglement & {} &    - &   - &        $0.27$ &  $\mathbf{0.53}$ &        $0.20$ &  $\mathbf{0.90}$ \\
Completeness & {} &   $0.17$ &  $\mathbf{0.90}$ &        $0.20$ &  $\mathbf{0.57}$ &        $0.13$ &  $\mathbf{0.90}$ \\
Informativeness & {} &   $0.94$ &  $\mathbf{0.99}$ &        $\mathbf{1.00}$ &  $\mathbf{1.00}$ &        $\mathbf{1.00}$ &  $\mathbf{1.00}$ \\
SAP & {} &   $0.03$ &  $\mathbf{0.87}$ &        $0.04$ &  $\mathbf{0.21}$ &        $0.01$ &  $\mathbf{0.67}$ \\
MIG & {} &   $0.01$ &  $\mathbf{0.17}$ &        $0.00$ &  $\mathbf{0.03}$ &        $0.00$ &  $\mathbf{0.08}$ \\
\bottomrule
\end{tabular}
\end{table*}

\begin{figure}[t]
\centering
\includegraphics[width=0.95\hsize]{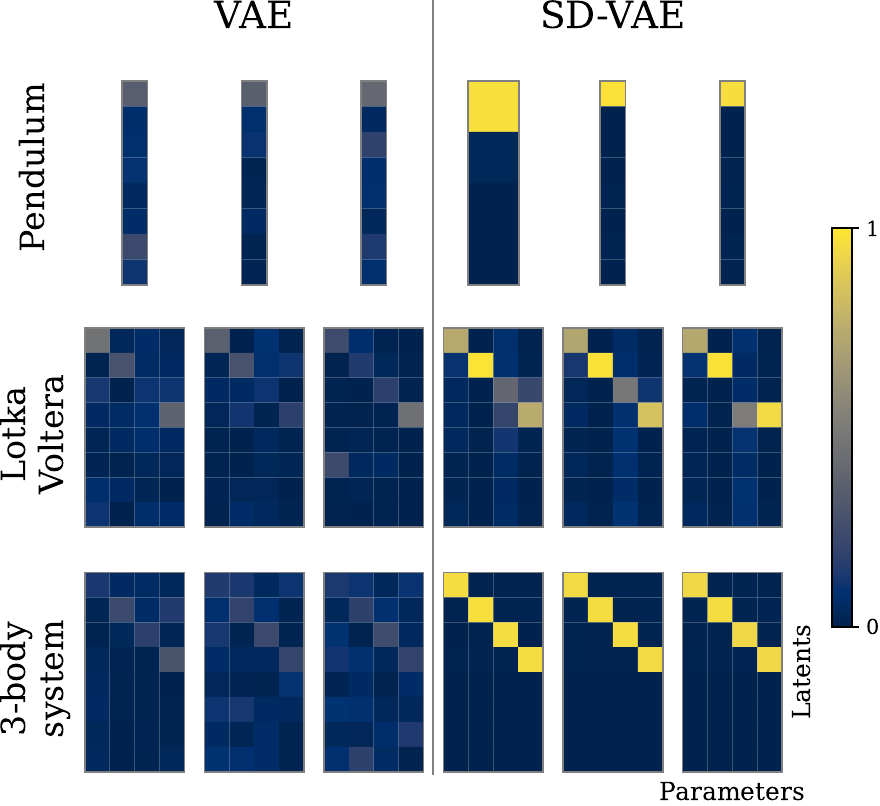}
\caption{\textbf{Disentanglement of representations.} The $x$-axis corresponds to the parameters and the $y$-axis to the latent various. The color scale denotes the value of the importance weights. These values were extracted from the weights of a regressor trained to predict the parameters from the latent values.  High values (yellow) indicate that the latent variable that has high predictive power over the ground-truth value. We present the top 3 models for each method and dataset. The latent space of the SD-VAE is disentangled in highly-predictive and non-predictive parts, while the VAE encoding exhibits no such characteristic. 
}
\label{fig:results_latents_importance_weights_truncated}
\end{figure}

\subsection{Disentanglement of representations}

We want to understand if latent space supervision leads to differences in the learned representations of VAEs and SD-VAEs. For this, we use various metrics to quantify disentanglement. Measuring disentanglement is a challenging task; many metrics have been proposed that do not always correlate well with each other \cite{Locatello2020AEvaluation}. In this work we use \textit{Disentanglement}, \textit{Completeness}, \textit{Informativeness} (DCI) \cite{Eastwood2018AREPRESENTATIONS}, a predictor-based measuring frameworks that analyses three different aspects of disentanglement. Briefly, \textit{Disentanglement} measures how well the factors of variation are factorized in the representation, \textit{Completeness} indicates if each factor is captured by a single latent variable, and \textit{Informativeness} quantifies the amount of information a representation captures about the factors of variation. Recent studies suggest that in practice DCI with random forests is "the best all-around metric" \cite{Zaidi2020MeasuringMetrics}. For completeness, we additional include Mutual Information Gap (MIG) \cite{Chen2018IsolatingVAEs} an information-theoretic metric that quantifies disentanglement as the difference between the mutual information (MI) between the top two latent-factor pairs, and the Attribute Predictability Score (SAP) \cite{Kumar2017VariationalObservations} a metric that works similarly to MIG but uses the importance weight of a learned predictor instead of MI.

Metrics in \cref{tab:resunts_disentanglement_metrics} indicate that SD-VAE produces more disentangled representations than the VAE in all the systems. Specifically, we observe a significant increase in \textit{Disentanglement}, \textit{Completeness} and SAP scores and a more modest increase in MIG. We also observe that the \textit{Informativeness} of both VAE and SD-VAE is close to the maximum ($1$), this suggests that the representation of the VAE also captures information about the parameters but this is spread across the latent dimensions. \textit{Disentanglement} can not be computed for the pendulum since there is only one factor of variation (length).

To compute the DCI metrics, we train a boosted trees regressor to predict the parameters from the latent codes (on the training dataset). The importance weights of the learned regressor demonstrate the predictive power of each latent for each parameter. We visualize the weights of the best SD-VAE and VAE models in \cref{fig:results_latents_importance_weights_truncated}. We use the best 3 models as before (selected by validation MAE). To allow better visual inspection we keep the first 8 latents. To further facilitate the comparison for the VAE we place the highest value of each column at the top diagonal positions ([1,1], [2,2] etc). We provide visualizations of the full latent space with importance weights. We observe that the supervised latents of the SD-VAE have very high predictive power for the system parameters, while the other latents are not significant. In the case of VAE, the predictive power is spread across the whole latent code. In conjunction with the disentanglement metrics, these findings demonstrate that latent space supervision produces highly disentangled representations.

\subsection{Linearity of correlation}
\label{sec:linearity_main_body}

The importance weights in the previous section denote the strong correlation between latents $\mathbf{z}$ and parameters $\bm\xi$. Since trees can capture both linear and non-linear dependencies, the nature of the relationship remains, unclear. To quantify the linearity, we fit linear regression models for each $z_i$, $\xi_j$ pair and compute the absolute Pearson correlation coefficient between the two variables. Pearson $r$ values are visualized in \cref{fig:correlation_latents_parameters}. Results  (see \cref{sec:linear_correlation}) indicate that the relationship between supervised latents and parameters and is strongly linear in most cases. This aligns with our experimental findings that linear scaling works best for the disentanglement loss (see \cref{sec:scaling_parameters}). We exploit this linearity to perform traversals of the latent space in the next section.

\subsection{Latent space traversals}
\label{sec:traversals_main_body}

Being able to traverse between two points in the latent space is a property that indicates meaningful representations. Interpolation in latent space can produce meaningful images in properly disentangled VAEs \cite{Higgins2017Beta-VAE:FRAMEWORK}. While images have easily recognizable visual components, traversals of dynamical systems are harder to portray. Here we study whether interpolating between two points in the latent space of SD-VAE can produce meaningful trajectories. For this, we create a new pendulum dataset containing 100 trajectories with linearly spaced length in the range  $l \in [1.0-1.5]$. The initial conditions are kept constant ($\theta=\frac{pi}{2}, \omega=0$) for all the trajectories to facilitate comparisons %and we use the same noise level as in the training dataset
. We use the encoder of our best SD-VAE model to extract the latent variables for each trajectory. For each trajectory, the encoder produces 4 latent variables $z_1 \dots z_4$.  We linearly interpolate between the latents of the two extreme trajectories ($l=1.0$ and $l=1.5$), driven by our findings in \cref{sec:linear_correlation} that latents and parameters have a higly linear correlation. Next, we feed the real and interpolated latents to the decoder and predict up to 1000 timesteps. We find that the total mean absolute error between prediction and ground truth is $0.29$ with the real latents and $0.33$ with the interpolated one. These results indicate that linear latent space interpolation produces meaningful latent codes. This is further corroborated by plotting the real and interpolated latents together (\cref{fig:autoregressive}). The relationship between the real latents $z_i$ and pendulum length $l$ is highly linear, which further explains with the linear interpolation method works well.

\subsection{Disentangling AEs and stability}

We pose the question of whether supervised disentanglement can also be applied to (deterministic) AEs. For this, we train both AE and SD-AE models and compare them with VAE and SD-VAE models (\cref{tab:results_mae_all_models}). Results indicate that disentangling AEs does not offer much if any improvement in prediction accuracy. Probabilistic models seem better suited to capture the variation in the data distribution. It also illustrates that latent space disentanglement is not trivial and more work is needed to help us understand what works in practice and why. 

We also trained LSTMs and found that their prediction accuracy is subpar compared to the other models. In fact for the Lotka-Volterra system, LSTMs proved to be very unstable: none of the 72 trained LSTMs could predict long-term (800 steps) without diverging. On the note of model stability, this is something to take into consideration when using supervised disentanglement in practice. In AEs supervised disentanglement resulted in a higher percentage of unstable models. SD-VAE was the most stable model in the pendulum and 3-body systems with more than 90\% of the models being stable, but in the Lotka-Volterra systems the VAE training produced more stable models.

%%%%%%%%%%%%%%%%%%%%%%%%%
%%%%%%%%%%%%%%%%%%%%%%%%%
%%%%% RESULTS VIDEO %%%%%
%%%%%%%%%%%%%%%%%%%%%%%%%
%%%%%%%%%%%%%%%%%%%%%%%%%

\begin{figure*}[ht]
  \centering
\includegraphics[width=\linewidth]{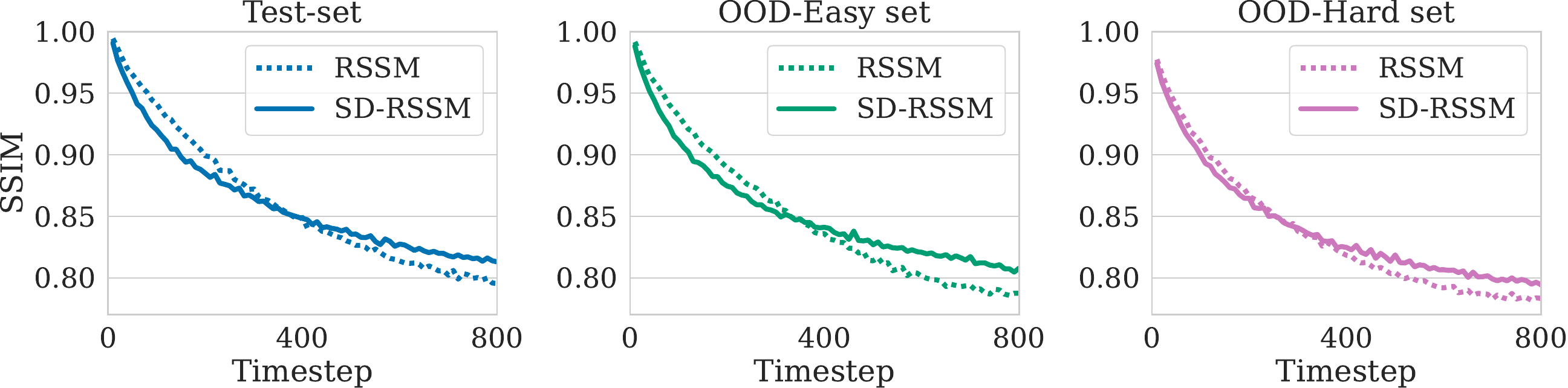}
 \caption{\textbf{Prediction quality on the observation-space pendulum.} Structural Similarity (SSIM) as a function of the predicted timestep. The disentangled SD-RSSM model seems more robust on long-term predictions.}
  \label{fig:results_rssm_ssim}
\end{figure*}

\section{Modelling dynamics in observation-space}

We extend the idea of supervised disentanglement to models that infer the state from high-dimensional observations such as image sequences.
 
\subsection{Datasets}

The dynamical system we use in this experiment is the swinging pendulum, a common benchmark for modelling dynamics in image sequences \citep{Brunton2020MachineMechanics, Barber2021JointSystems}. The data set contains sequences of images of a moving pendulum. The positions of the pendulum are first computed by a numerical simulator and then rendered in image space as frames of dimension $64\times 64$. The length of the pendulum $\ell$, the strength of gravity $g$ and the initial conditions (position  $\theta$, angular velocity $\omega$ ) are set to different values in each trajectory so they differ from each other. Parameters are drawn from a uniform distribution. For the OOD sets we change the distributions of length $l$, and gravity $g$ but keep the same distribution of $\theta$ and $\omega$ as in training. More details about the data set are illustrated in \cref{sec:dataset-pp}. For the simulations, we use an adaptive Runge-Kutta integrator with a timestep of $0.05$ seconds. 

\subsection{Model and Training}%: Recurrent State-Space Model (RSSM)}

In this experiment, we use RSSM described in \cref{sec:methods-rssm}. RSSM is a generative model including both stochastic and deterministic latents. We use supervised disentanglement on the stochastic part, and term that model SD-RSSM.  The RSSM model we use follows the architecture as described in \cite{Hafner2018LearningPixels} \& \cite{Saxena2021ClockworkAutoencoders}. Disentanglement is applied to all four parameters (length $\ell$, gravity $g$, initial position  $\theta$ and velocity $\omega$), but only length and gravity vary between datasets. For training, we use sequences of 100 frames and batch size 100. 
We use an $L_2$ loss for the disentanglement term because preliminary results showed it performs better than $L_1$ and BCE. %All models were trained for 300 epochs with a learning rate of $10^{-3}$ and an Adam optimizer ($b_1=0.9$ and $b_2=0.999$). 
During testing the model uses 50 frames as context. We train 24 RSSM and 24 SD-RSSM models (detailed hyperparameters in \cref{sec:hyperparameters_observation_space_pendulum}). 

%%%%%%%%%%%%%%%%%%%%%%%%%
%%%%%%%%%%%%%%%%%%%%%%%%%
%% RESULTS VIDEO QUAL %%%
%%%%%%%%%%%%%%%%%%%%%%%%%
%%%%%%%%%%%%%%%%%%%%%%%%%

\subsection{Results}

We compare the predictions of RSSM and SD-RSSM using structural similarity (SSIM), a metric that takes into account the qualitative characteristics of the image, something that pixel-wise metrics like MSE and MAE fail to do  \cite{zhou2004ssim}. We select the best RSSM and SD-RSSM models based on the average SSIM over the validation set and plot the SSIM as a function of the timestep (\cref{fig:results_rssm_ssim}) and is widely used for dynamical system prediction from spatiotemporal data \cite{pant2020ssim}. Results show that while for the short-term predictions, the RSSM has higher SSIM, in long-term prediction after about 350 steps SD-RSSM is performing better, in all 3 datasets. Furthermore, as we move from the test-set to the OOD sets, we observe that the SD-RSSM model closes the performance gap in the short-term prediction. Specifically, in the OOD-Hard for predictions up to around 350 step it is almost equivalent to RSSM while at the same time maintaining its long-term (>350) advantage. We hypothesize that SD-RSSM has better long-term performance due to less overfitting to the relatively short training horizon. Similarly, the robustness in increasing distribution shifts could also be explained by less overfitting on the parameters of the training data. We also compared using the peak signal-to-noise ratio (PSNR), drawing similar conclusions (details in \cref{sec:results_extra_observation_space}). Qualitative results show that both models produce accurate short time predictions and also accurately capture the appearance of the pendulum even in long-term predictions. Where they differ is in how well they capture the long-term dynamics indicating that latent space disentanglement is helpful for long-term prediction. Overall, results suggest that supervised disentanglement can be used to model dynamical systems in observation-space sequences, resulting in improved long-term and OOD performance.

\section{Conclusion}
We have shown that using ground-truth parameters to supervise the latent space of VAEs encourages them to learn more disentangled and interpretable representations while at the same time increasing their prediction accuracy and OOD generalization in three dynamical systems. We have, further, shown that supervised disentanglement improves generative models like RSSM trained on observation-space data of a swinging pendulum and leads to better long-term forecasting performance and robustness to OOD shifts. 
% We showed that SD-VAEs, trained with a supervised disentanglement loss, outperform normal VAEs in prediction accuracy and OOD generalization in three dynamical systems. At the same time using supervised disentanglement in models like RSSM improves long-term prediction of a swinging pendulum from observation-space data. 
These results make supervised disentanglement an attractive choice for the generative modelling of system dynamics. In practice, VAE and SD-VAEs should be preferred over their deterministic counterparts. Using simulated data makes the label collection cheap but this is not always possible. Extending our method to the semi-supervised setting, i.e., supervising with few labels, is important for real-world applications where the collection of labels is more expensive but robust prediction of system dynamics remains critical. Further analysis of the method using systems with more complex dynamics is also an important avenue for future work.

\section*{Acknowledgements}

M.L. and S.F. acknowledge financial support from the Departments of Aeronautics and Bioengineering respectively.

\bibliographystyle{icml2023}
% \bibliography{mendeley, main}
\bibliography{main_clean}

%%%%%%%%%%%%%%%%%%%%%%%%%%%%%%%%%%%%%%%%%%%%%%%%%%%%%%%%%%%%%%%%%%%%%%%%%%%%%%%
%%%%%%%%%%%%%%%%%%%%%%%%%%%%%%%%%%%%%%%%%%%%%%%%%%%%%%%%%%%%%%%%%%%%%%%%%%%%%%%
% APPENDIX
%%%%%%%%%%%%%%%%%%%%%%%%%%%%%%%%%%%%%%%%%%%%%%%%%%%%%%%%%%%%%%%%%%%%%%%%%%%%%%%
%%%%%%%%%%%%%%%%%%%%%%%%%%%%%%%%%%%%%%%%%%%%%%%%%%%%%%%%%%%%%%%%%%%%%%%%%%%%%%%
\newpage
\appendix
\onecolumn

\section*{Software and Data}
We provide all the necessary code to reproduce our experiments at https://github.com/stathius/sd-vae. The repository contains code and instructions for generating all the datasets and training all the models presented in this work using the hyperparameters that are clearly presented in the paper. This should significantly help others reproduce our experiments. For any further clarifications, readers are encouraged to contact the corresponding author(s).

\section*{Accessibility}
We have used vector-based figures to increase clarity for zooming-in, a color palette that is easily distinguishable by colorblind people and different line styles. We have also curated arxiv citations to refer to the corresponding conference or journal publications where possible. 

\section{Datasets}

\subsection{Phase space}
\label{sec:dataset_phase_space}
For simulations, we use an adaptive Runge-Kutta integrator with a timestep of $0.01$ seconds. Each simulated sequence has a different combination of parameters. Simulation of the pendulum uses an initial angle $\theta$ which is randomly between $10^{\circ}-170^{\circ}$ while the angular velocity $\omega$ is 0. For the other two systems the initial conditions are always the same to avoid pathological configurations. 

\begin{table*}[thb]
\caption{\textbf{Datasets.} In L-V and 3-body OOD test sets, at least one domain parameter is outside of the parameter range used for training.}
\makebox[\textwidth]{
\small
\begin{tabular}{lcccl}
    \toprule
  & Pendulum            & Lotka-Volterra                & 3-Body                    &  \\
\cmidrule(r){2-4}
ODEs & $\ddot{\theta}+\frac{g}{\ell} \sin \theta=0$ &$\begin{array}{l}
\dot{x}=\alpha x-\beta x y \\ 
\dot{y}=\delta x y-\gamma y
\end{array}$& $
\begin{array}{c}\bar{m}_{i} \frac{d \vec{v}_{i}}{d t}=K_{1} \sum_{j} \frac{\bar{m}_{i} \bar{m}_{j}}{\bar{r}_{i j}^{3}} \overrightarrow{r_{i j}} \\ \frac{d \overrightarrow{\bar{x}}_{i}}{d \bar{t}}=K_{2} \vec{v}_{i}\end{array}$&  \\
Number   of ODEs           & 1                   & 2                  & 6                         &  \\
Independent Variables                  & $\theta$,  $\omega$      & $x$(prey), $y$(predator)   & $\overrightarrow{x}_i, \overrightarrow{v}_i, i=1,2,3$      &  \\
Initial values           & $\begin{array}{c}
\theta \in[10^o-170^o] \\ \omega=0
\end{array}$ & $x=5,y=3$     & $
\begin{array}{l}
\overrightarrow{x_1}=(-1,-1), \overrightarrow{v_1}=(0.0, 0.5)\\ 
\overrightarrow{x_2}=(1, -1), \overrightarrow{v_2}=(0.5 -0.5) \\ 
\overrightarrow{x_3}=(0,1), \overrightarrow{v_3}=(-0.5,  0.0)
\end{array}$ &  \\ 
 Timestep & 0.01 & 0.01 & 0.01 \\
 Sequence length & 2000 & 1000 & 1000 \\
 Noise $\sigma^2$ & 0.05 & 0.05 & 0.01\\
    \midrule
Parameters       & $l$(length)              & $\alpha, \beta, \gamma, \delta$           & $K_2, m_1, m_2, m_3$          &  \\
Train/Val/Test      & $l \in [1.0-1.5]$ & 
$\begin{array}{c}
A = \{\alpha \in [1.95,2.05] \}\\
B = \{\beta\in[0.95,1.05]\} \\ 
C = \{\gamma\in[3.95,4.05]\}\\
D = \{\delta\in [1.95,2.04]\} \\
\Omega_{\text{train}} = (A \times B \times C \times D) \\
\end{array}$ & $\begin{array}{c}
K = \{ K_2 \in[1.95,2.05]\} \\
M1 = \{ m_1 \in[1.95,2.05]\} \\
M2 = \{ m_2 \in[1.95,2.05]\} \\
M3 = \{ m_3 \in[1.95,2.05]\} \\
\Omega_{\text{OOD-Hard}} = \\ (K  \times M1  \times M2  \times M3) \\
\textcolor{white}{.} \\
\end{array}$ &  \\
OOD Test Set Easy    &   $l \in[1.5-1.6]$  &  $\begin{array}{c} 
A = \{\alpha \in [1.94,2.06] \}\\
B = \{\beta\in[0.94,1.06]\} \\ 
C = \{\gamma\in[3.94,4.06]\}\\
D = \{\delta\in [1.94,2.06]\} \\
\Omega_{\text{OOD-Easy}} = \\ (A \times B \times C \times D) \setminus \Omega_{\text{train}} \\
\textcolor{white}{.} \\
\end{array}$& 
$\begin{array}{c}
K = \{ K_2 \in[1.94,2.06]\} \\
M1 = \{ m_1 \in[1.94,2.06]\} \\
M2 = \{ m_2 \in[1.94,2.06]\} \\
M3 = \{ m_3 \in[1.94,2.06]\} \\
\Omega_{\text{OOD-Hard}} = \\ (K  \times M1  \times M2  \times M3) \setminus \Omega_{\text{train}}  \\
\end{array}$ &  \\
OOD Test Set Hard    &  $l \in[0.9-1.0]$ & 
$\begin{array}{c}
A = \{\alpha \in [1.93,2.07] \}\\
B = \{\beta\in[0.93,1.07]\} \\ 
C = \{\gamma\in[3.93,4.07]\}\\
D = \{\delta\in [1.93,2.07]\} \\
\Omega_{\text{OOD-Hard}} = \\ (A \times B \times C \times D) \setminus \\
(\Omega_{\text{train}} \cup \Omega_{\text{OOD-Easy}}) \\
\textcolor{white}{.} \\
\end{array}$& 
$\begin{array}{c}
K = \{ K_2 \in[1.93,2.07]\} \\
M1 = \{ m_1 \in[1.93,2.07]\} \\
M2 = \{ m_2 \in[1.93,2.07]\} \\
M3 = \{ m_3 \in[1.93,2.07]\} \\
\Omega_{\text{OOD-Hard}} = \\ (K  \times M1  \times M2  \times M3) \setminus \\
(\Omega_{\text{train}} \cup \Omega_{\text{OOD-Easy}}) \\
\end{array}$ &  \\
\midrule
Number of sequences \\ 
Train/Val/Test  & \multicolumn{3}{c}{8000/1000/1000} \\
OOD Test Set Easy& \multicolumn{3}{c}{1000}  \\
OOD Test Set Hard & \multicolumn{3}{c}{1000}   \\
\bottomrule
\end{tabular}}
\label{tab:datasets}
\end{table*}

\subsection{Visualizing dataset shift}

Visualizing the distribution shift between datasets is not always straightforward. Especially in cases like dynamical system trajectories where there is usually not much familiarity with their visual representation in comparison for example to natural images. To facilitate qualitative comparisons we we depict the datasets from the three dynamical systems. We provide plots for all the dynamical systems and each dataset in separate figures so that the differences become more apparent. Apart from the phase space diagrams we also provide trajectories across time, offering another way to discern the difference in dynamics.

\begin{figure*}[ht]
    \centering
    \includegraphics[width=0.3\linewidth]{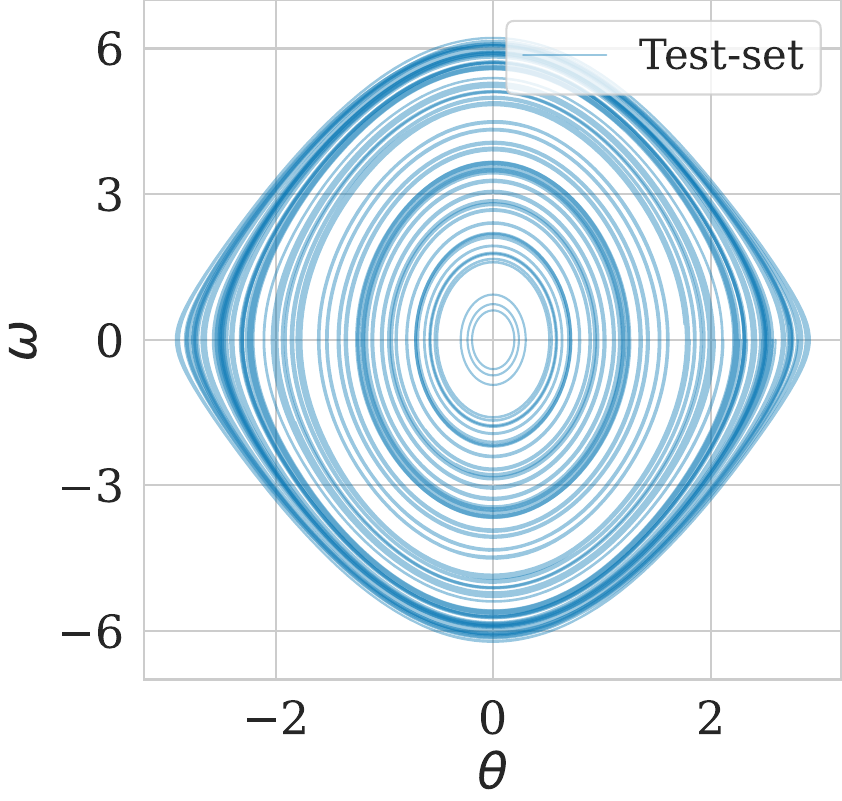}
    \hfill
    \includegraphics[width=0.3\linewidth]
    {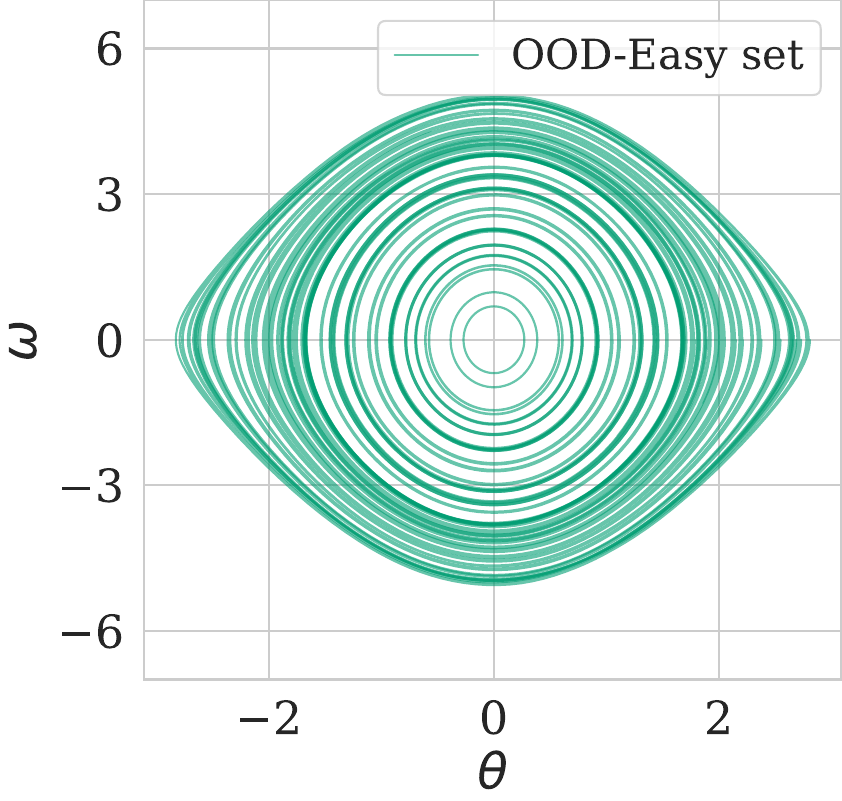}
    \hfill
    \includegraphics[width=0.3\linewidth]{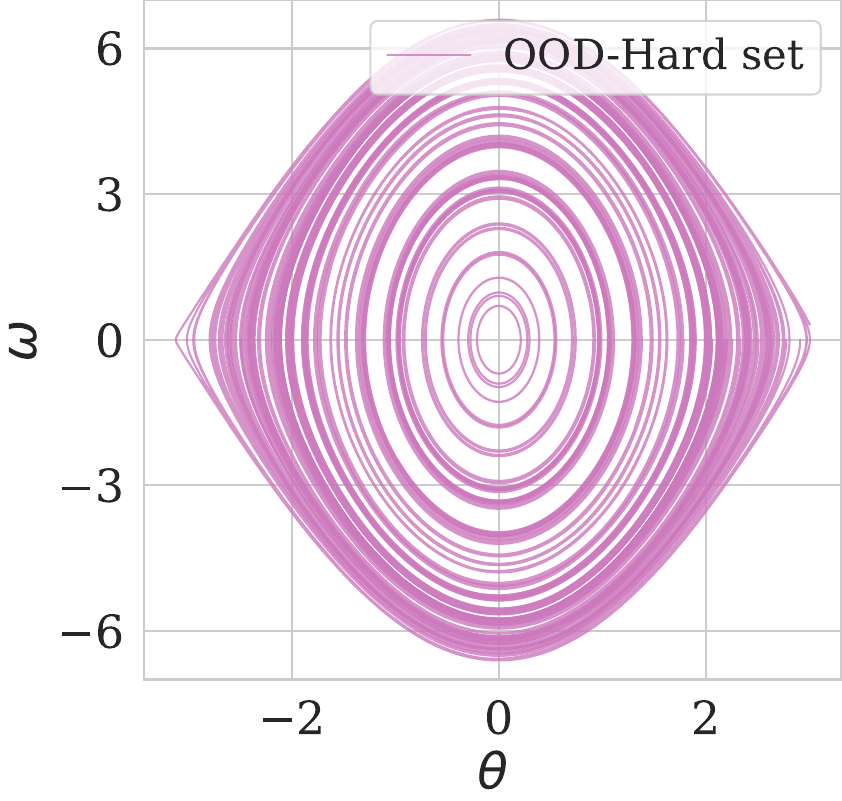}

    \includegraphics[width=0.3\linewidth]{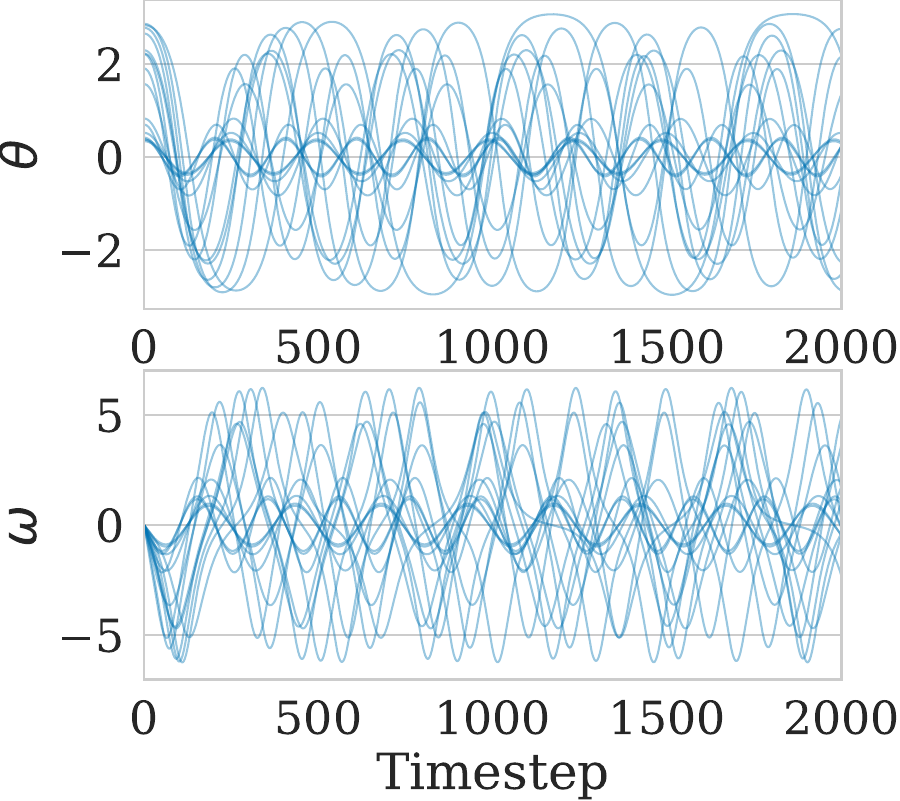}
    \hfill
    \includegraphics[width=0.3\linewidth]{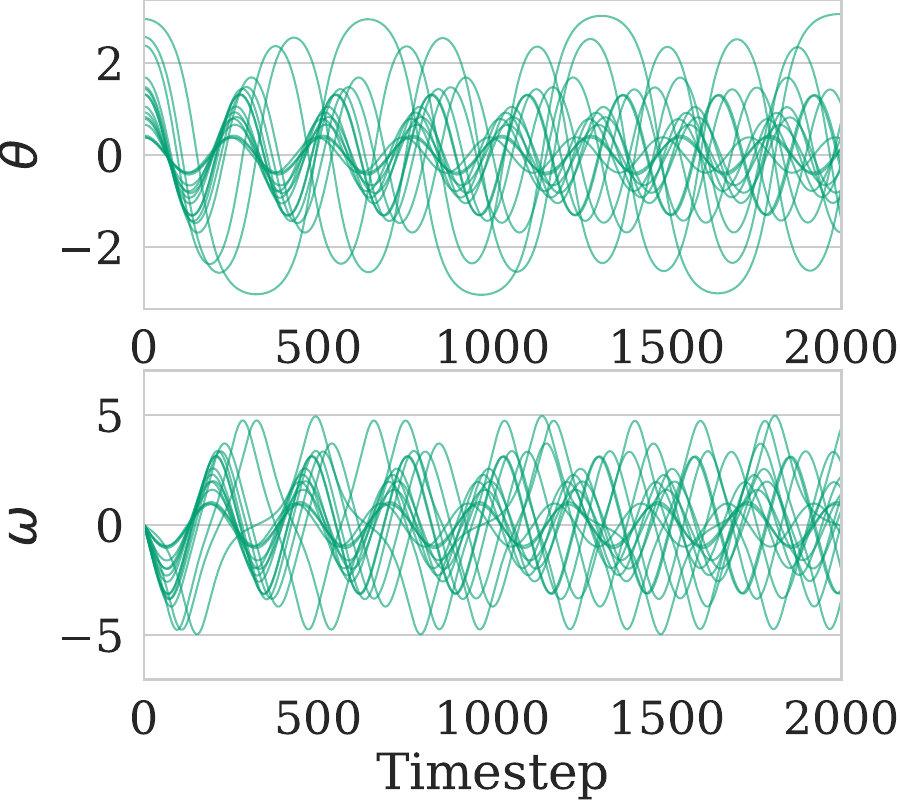}
    \hfill
    \includegraphics[width=0.3\linewidth]{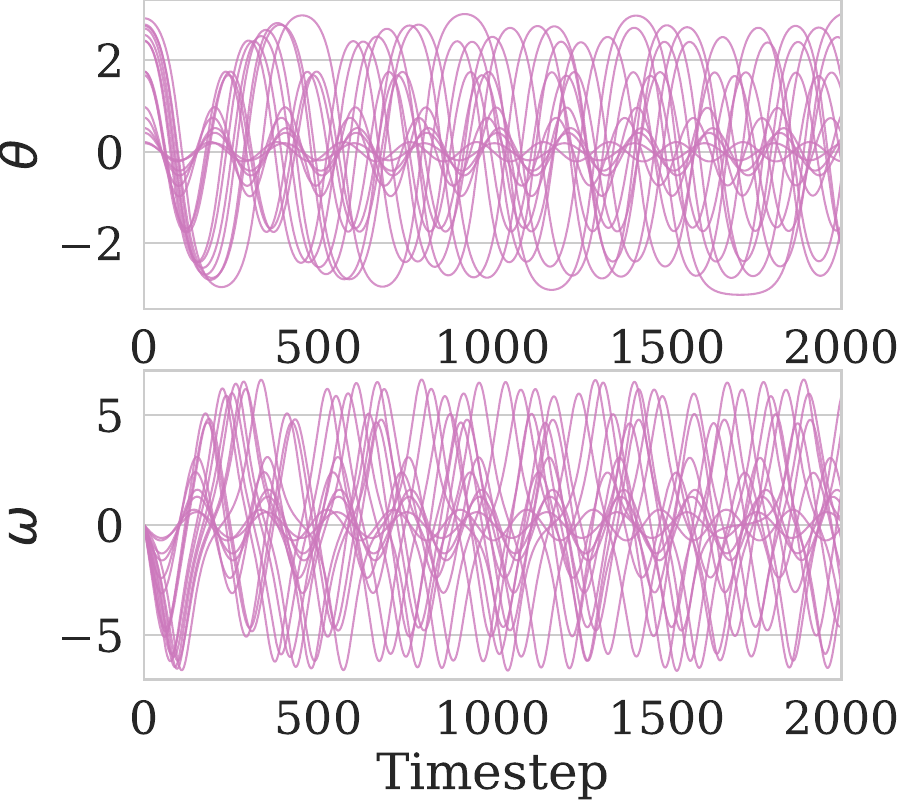}
\caption{Phase space diagrams \textbf{(top)}  and evolution over time (\textbf{bottom}) for random samples from the pendulum datasets. The OOD-Hard test set exhibits higher variation in the trajectories of $\theta, \omega$ as can be seen in the bottom row.}
  \label{fig:pend-trajectories}
\end{figure*}

\begin{figure*}[h]
    \centering
  \begin{subfigure}{.4\linewidth}
\includegraphics[width=\linewidth]{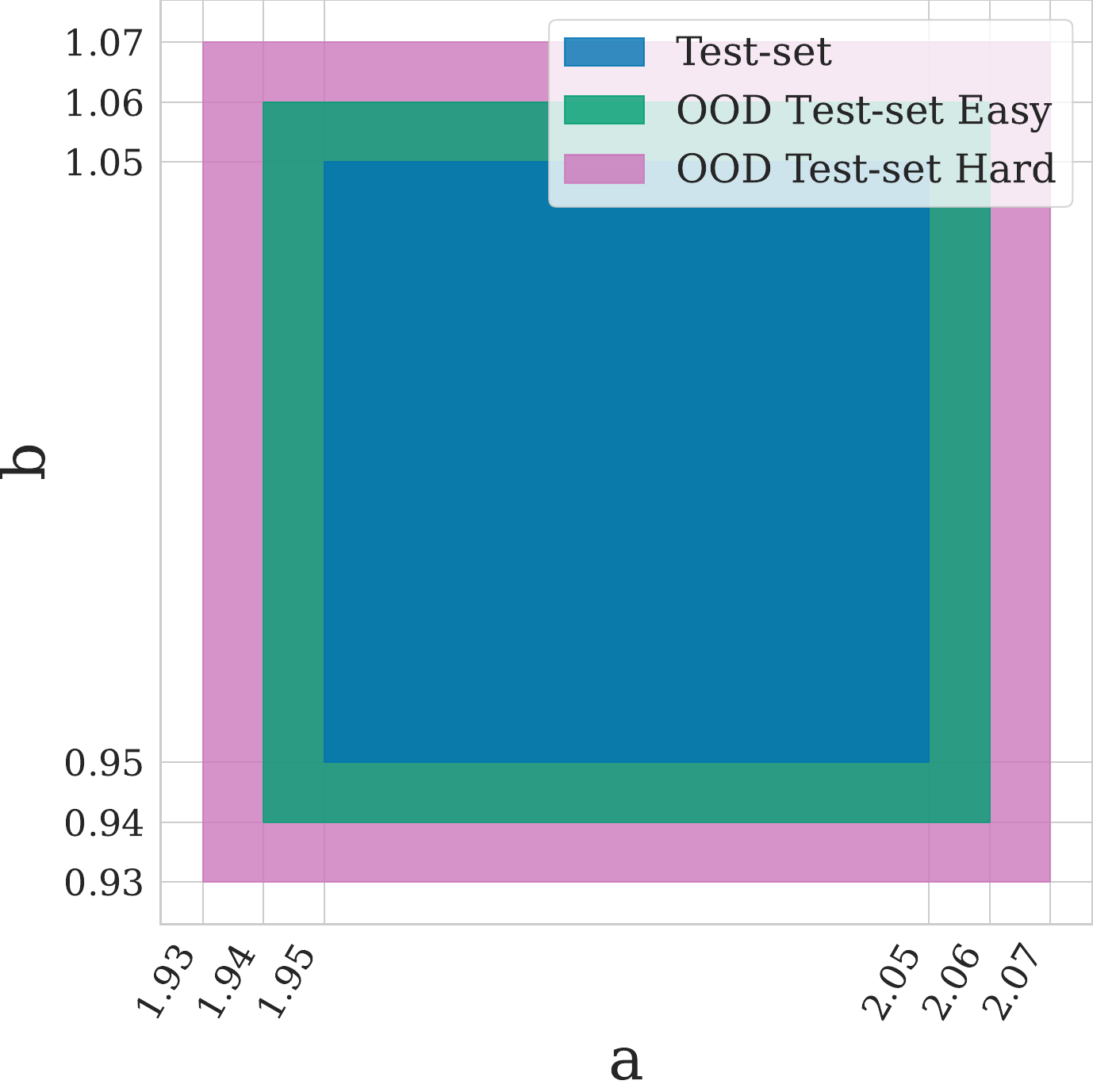}
  \end{subfigure}
%   \hfill
    \begin{subfigure}{.4\linewidth}
    \includegraphics[width=\linewidth]{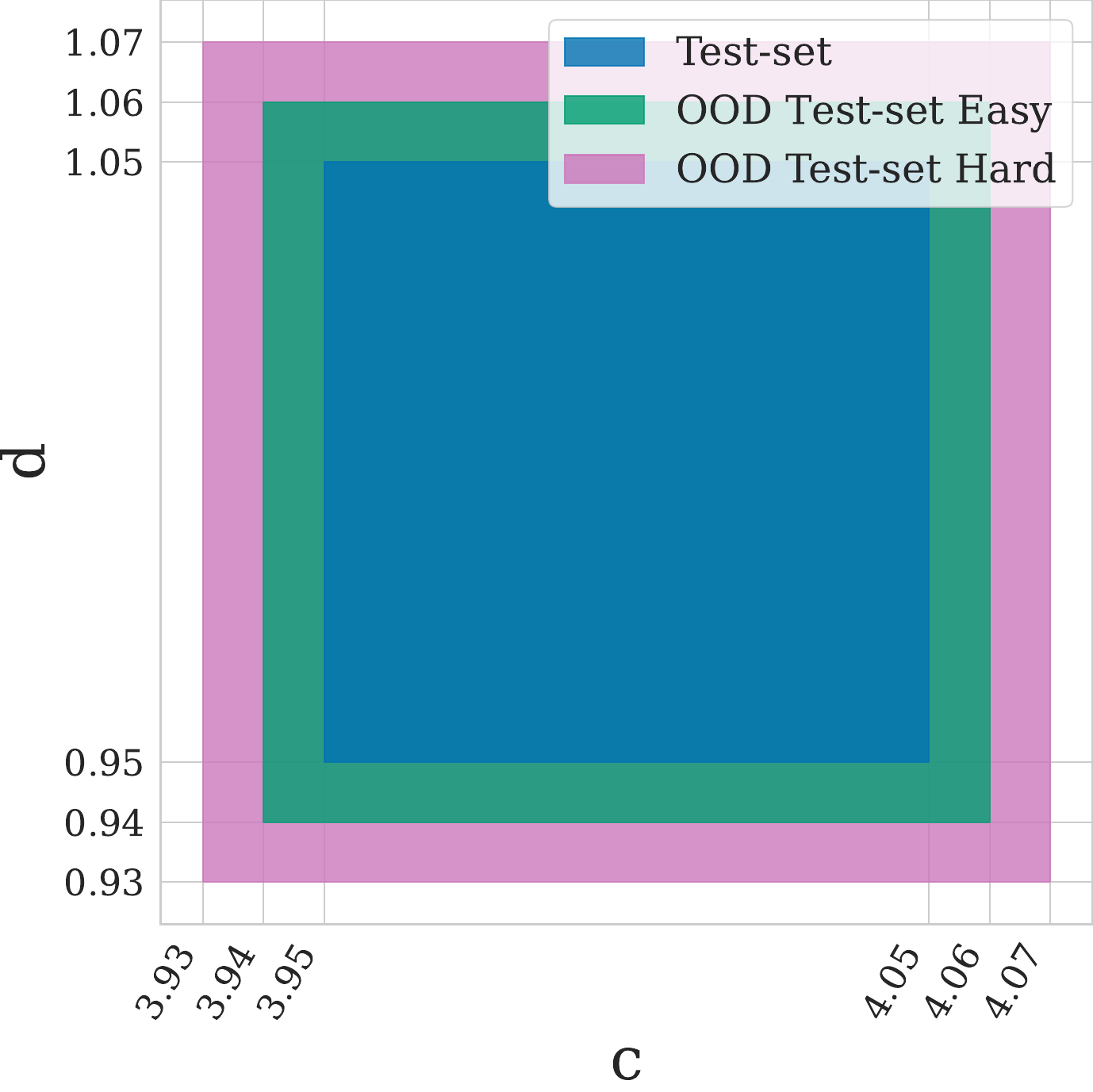}
  \end{subfigure}
\caption{Example illustration of the parameter distribution for the LV test sets. The regions do not overlap, colors represent regions not boundaries. The OOD-Easy test (green) set does not include any of the parameter configurations of the training and original test set (blue). Respectively, the OOD-Hard dataset (magenta) does not include none of the OOD-Easy or the original test set configurations. The parameter space of the blue region is almost half as big at the green area (again without any overlap), signifying a significant OOD shift).
}
  \label{fig:data-lv}
\end{figure*}

\begin{figure*}[ht]
    \centering
    \includegraphics[width=0.3\linewidth]{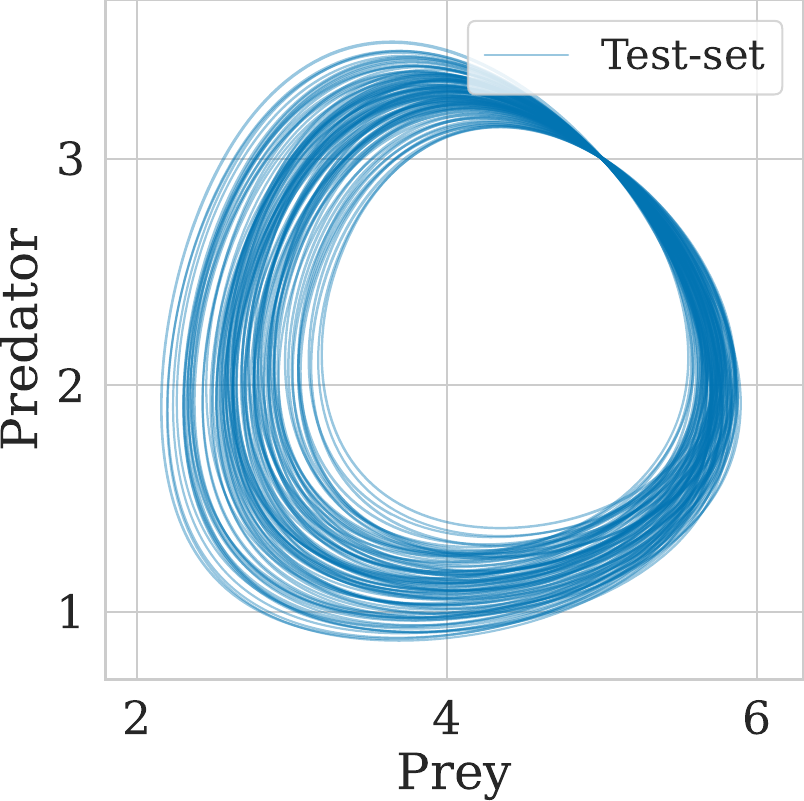}
    \hfill
    \includegraphics[width=0.3\linewidth]
    {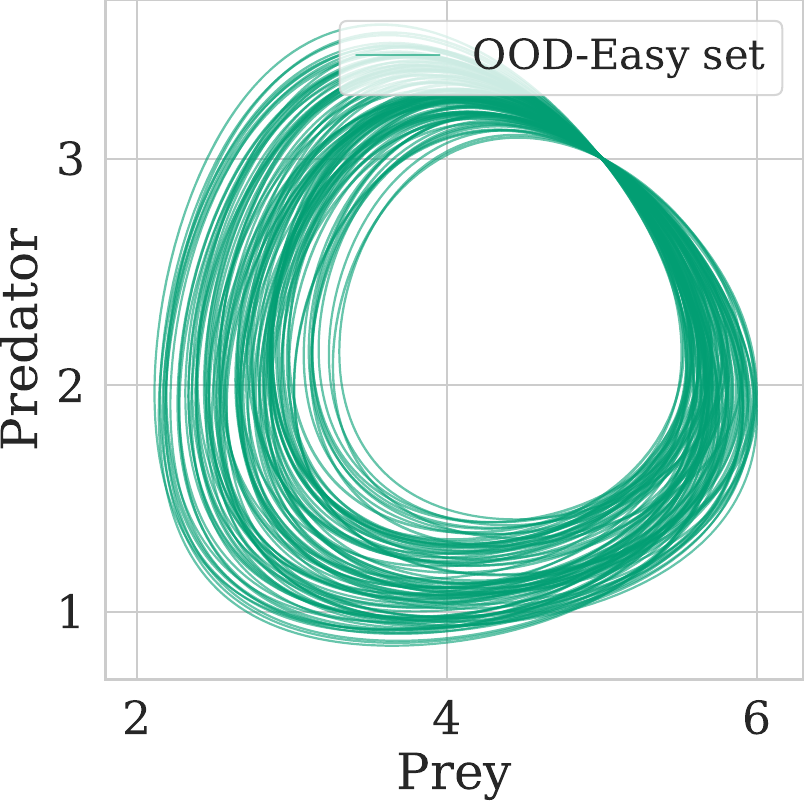}
    \hfill
    \includegraphics[width=0.3\linewidth]{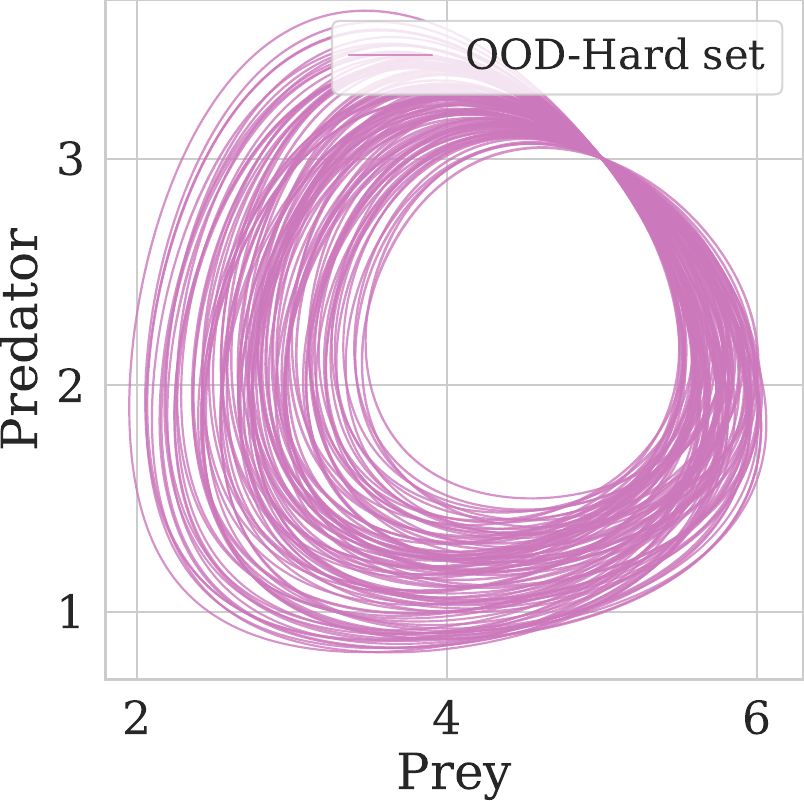}

    \includegraphics[width=0.3\linewidth]{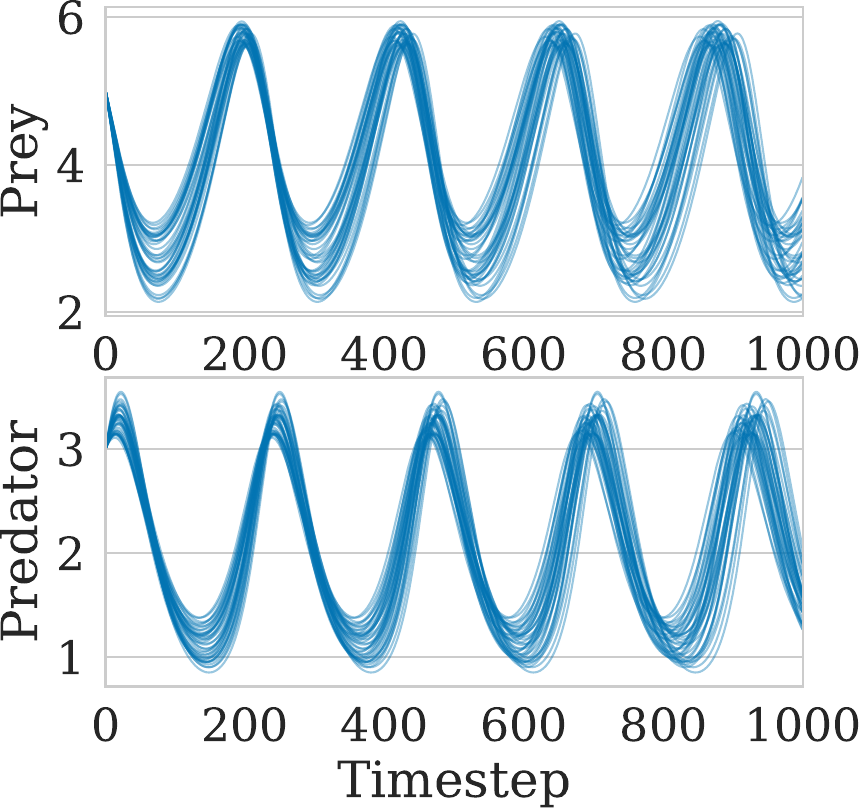}
    \hfill
    \includegraphics[width=0.3\linewidth]{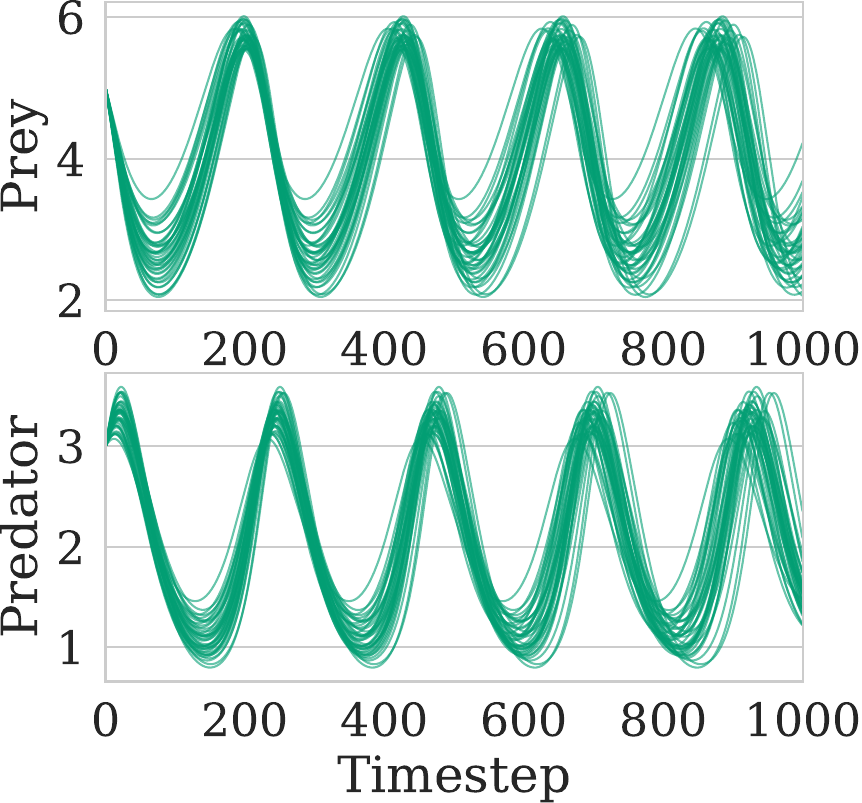}
    \hfill
    \includegraphics[width=0.3\linewidth]{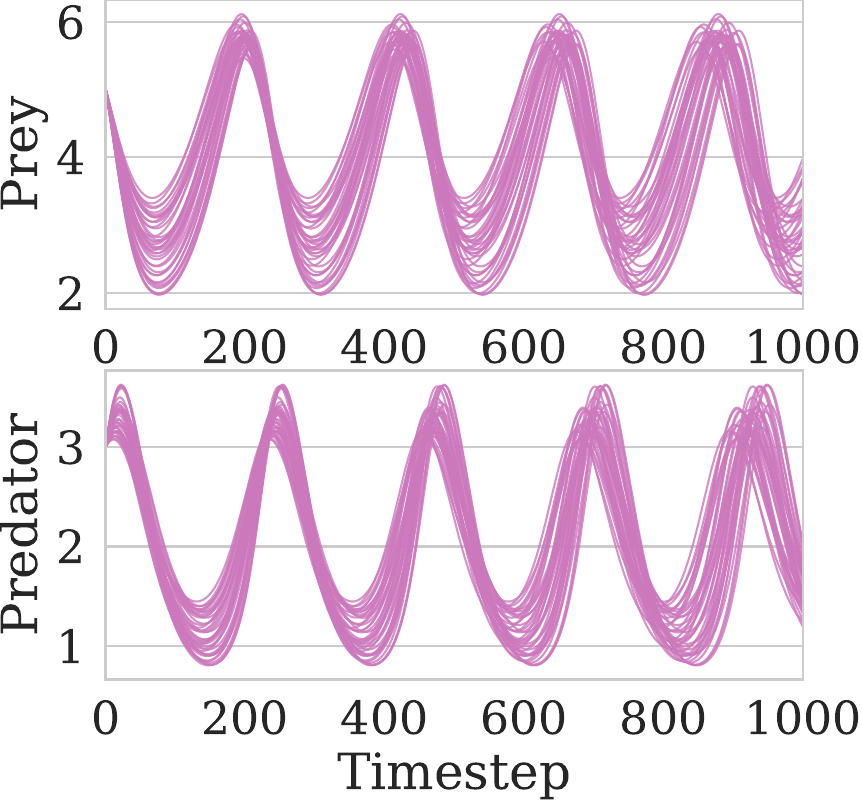}
\caption{Phase space diagrams \textbf{(top)}  and evolution over time (\textbf{bottom}) for random samples from the Lotka-Volterra datasets. The OOD test sets have an increasingly wider coverage of the domain in the phase-space and time. }
  \label{fig:lv-trajectories}
\end{figure*}

\begin{figure*}[ht]
    \centering
    \includegraphics[width=0.3\linewidth]{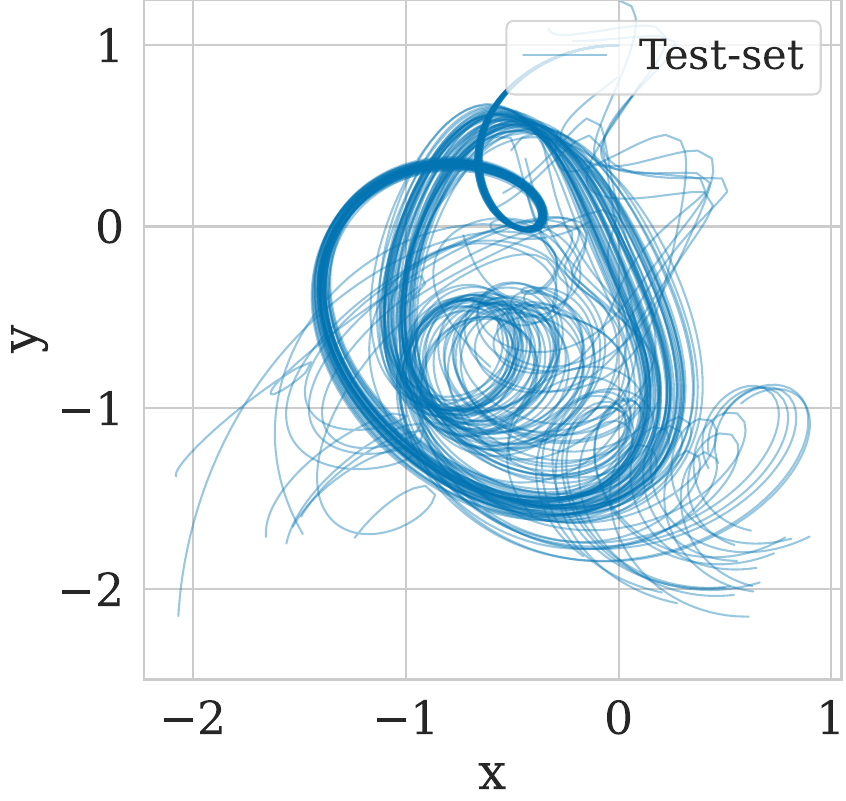}
    \hfill
    \includegraphics[width=0.3\linewidth]
    {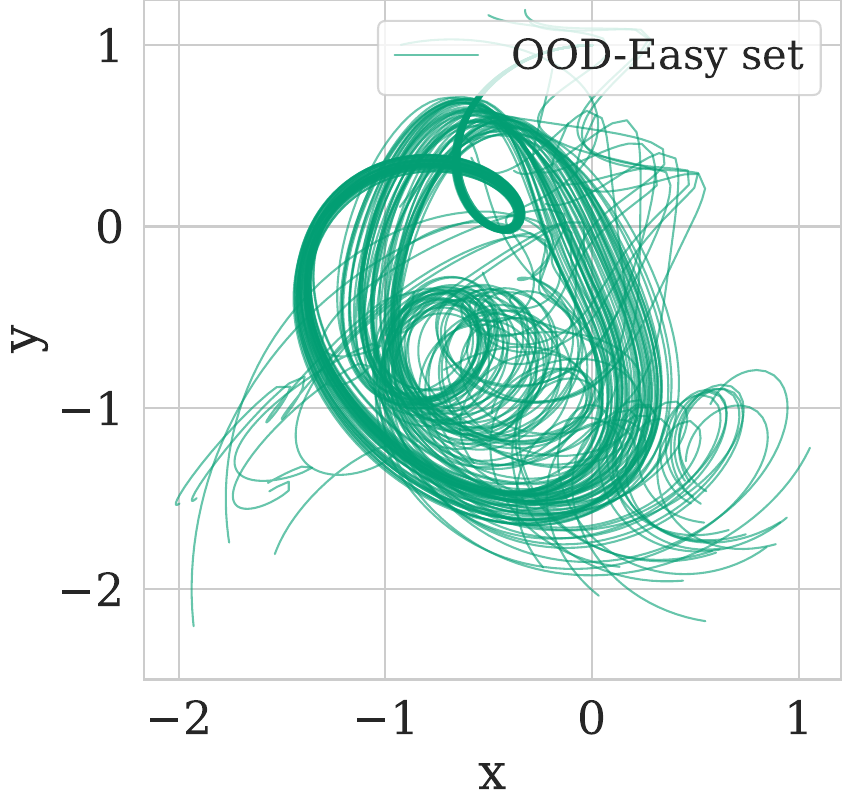}
    \hfill
    \includegraphics[width=0.3\linewidth]{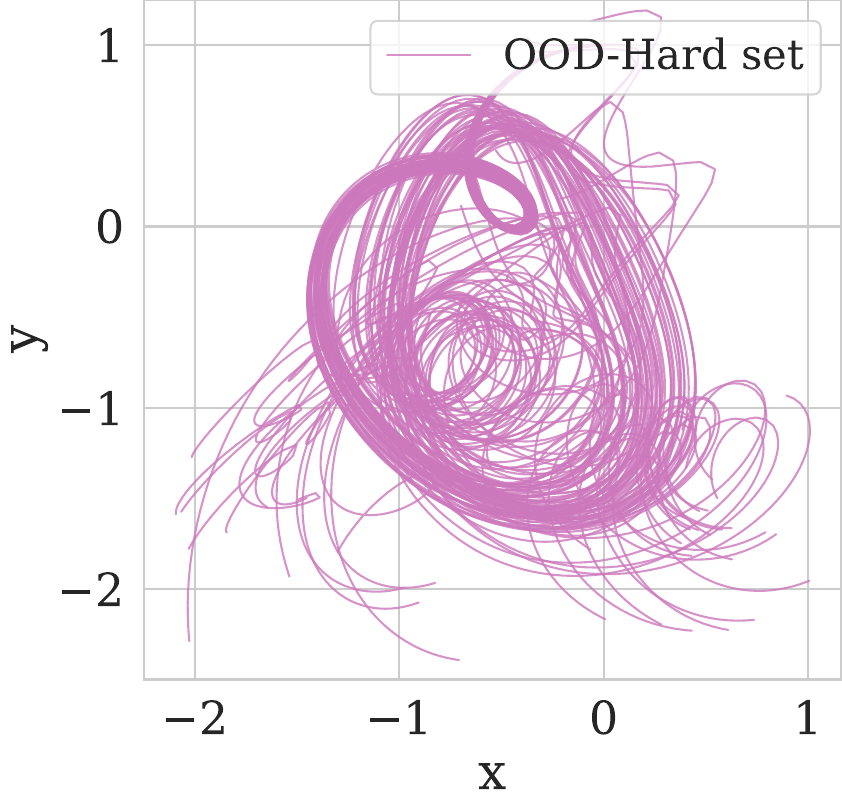}

    \includegraphics[width=0.3\linewidth]{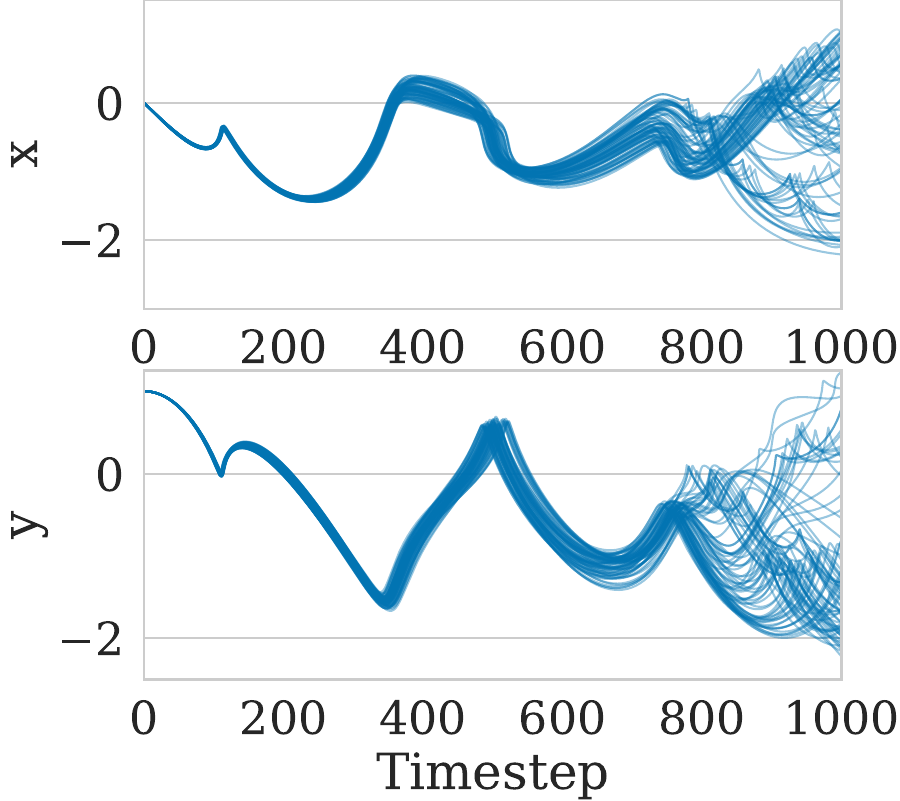}
    \hfill
    \includegraphics[width=0.3\linewidth]{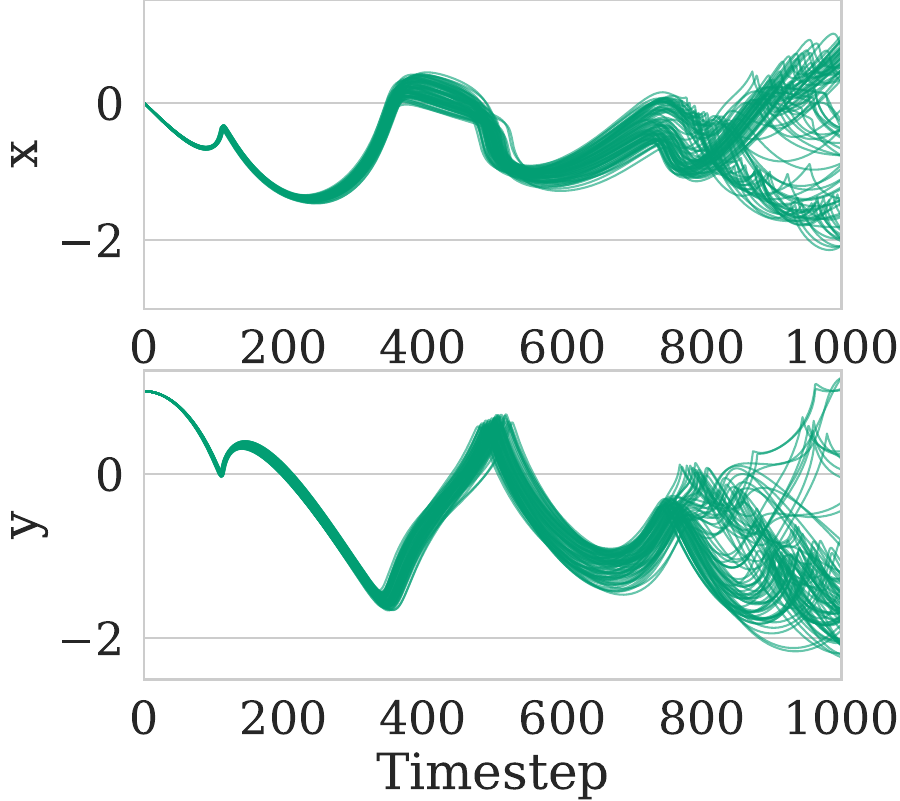}
    \hfill
    \includegraphics[width=0.3\linewidth]{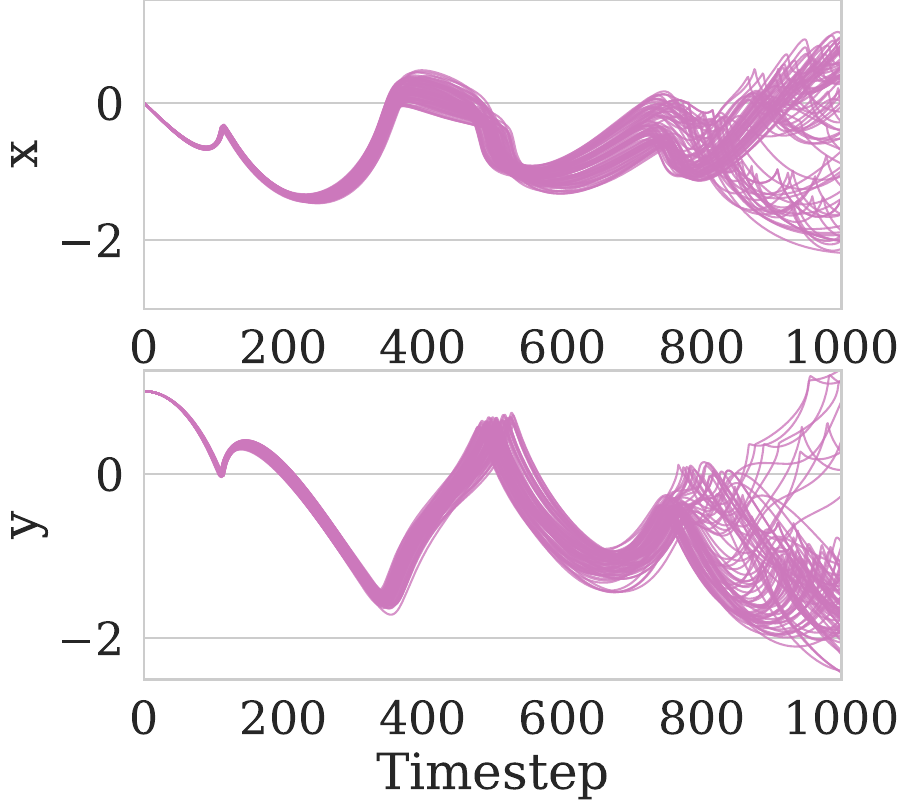}
\caption{Cartesian coordinates \textbf{(top)} and evolution over time (\textbf{bottom}) for random samples from the 3-body system datasets. We only plot the first body to avoid cluttering. The OOD test sets include a wider range of possible trajectories. This is evident by the higher coverage of the domain in the cartesian coordinates plot.}
  \label{fig:3body-trajectories}
\end{figure*}

\FloatBarrier

\subsection{Observation-space}\label{sec:dataset-pp}

\begin{figure}[th]
\centering
\includegraphics[width = 0.8\hsize]{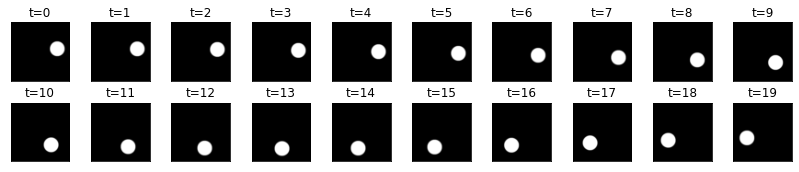}
\caption{Example image sequence from the observation-space pendulum data set}
\label{fig:pendulum}
\end{figure}

\begin{figure}[th]
  \centering
  \begin{subfigure}{.35\linewidth}
\includegraphics[width=\linewidth]{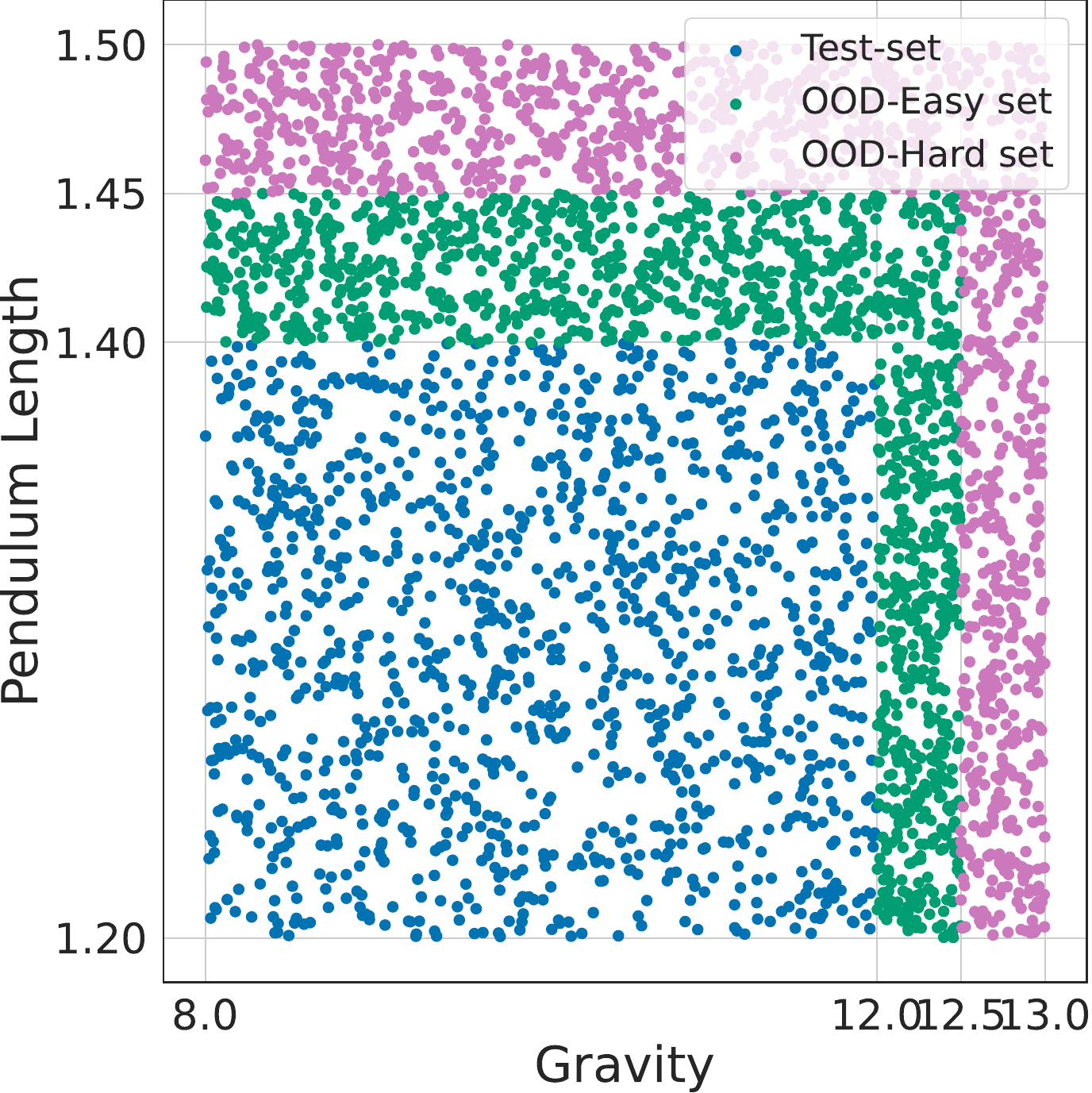}
  \end{subfigure}
      \hspace{1cm}
    \begin{subfigure}{.35\linewidth}
    \includegraphics[width=\linewidth]{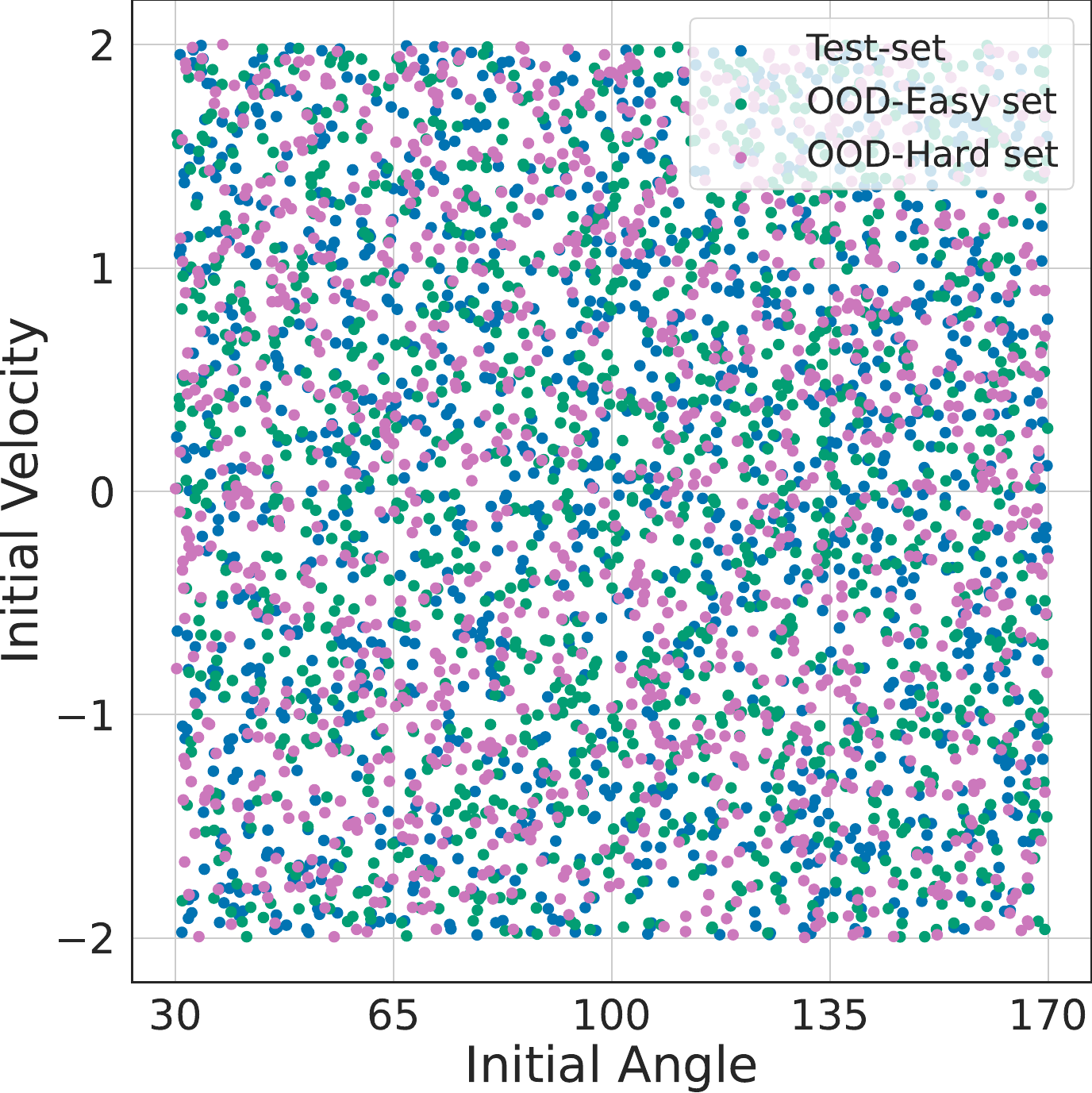}
  \end{subfigure}
        \hspace{1cm}
\caption{\textbf{Parameter distribution for the observation-space pendulum test-sets.} \textbf{Left} For the in-distribution test-set we draw the pendulum length and gravity from the same distribution as during training. The OOD test-sets represent distribution shifts of increasing magnitude, where parameters are drawn from totally different space which has zero overlap with the training and in-distribution test-set. \textbf{Right} The initial angle and angular velocity are drawn from the same uniform distribution for all test-sets. }
  \label{fig:data-pp}
\end{figure}

This data set contains image sequences of a moving pendulum under different conditions. The positions of the pendulum are first computed by a numerical simulator and then rendered in pixel space as frames of dimension $64\times 64$. An example image sequence is shown in \cref{fig:pendulum}. For the simulations, we use an adaptive Runge-Kutta integrator with a timestep of $0.05$ seconds. The length of the pendulum, the strength of gravity and the initial conditions (position, momentum) are set to different values so that each trajectory slightly differs from the others. The initial angle and initial velocity are drawn from the same uniform distribution for all data sets. The initial angle ranges from $30^{\circ}$ to $170^{\circ}$ and the initial velocity ranges from  $-2$ to $2$ rad/s.
For training, validation and in-distribution testing set, the gravity fall in the range $8.0 - 12.0$ $\text{m}^2/\text{s}$ , and the pendulum length lies between $1.20 -1.40$ m. In the easy OOD testing set, the gravity is between $12.0 - 12.5$ $\text{m}^2/\text{s}$ and the pendulum length is between $1.40 - 1.45$ m, while in the hard OOD testing set, the gravity is $12.5 - 13.0$ $\text{m}^2/\text{s}$ and the pendulum length is $1.45 - 1.50$ m.  The distributions of these parameters are shown in  \cref{fig:data-pp}.

\FloatBarrier

\section{Motivating disentanglement}

Disentanglement for dynamical system prediction is motivated both from previous experimental results and theoretically.

\subsection{Experimental motivation}

It is well established that disentangled representations can improve downstream tasks performance and are less prone to overfitting \cite{bengio2013}. For example, in image generation, disentangled representations enable more controlled synthesis of images with desired attributes while in image reconstruction or inpainting, they can help fill in missing parts of an image while preserving the existing attributes \cite{Higgins2017Beta-VAE:FRAMEWORK, Locatello2020AEvaluation}.

Disentangled representations can also be beneficial for time-series prediction by separating appearance from the underlying dynamics \cite{Li2018DisentangledAutoencoder} or by  separating trends, seasonal patterns, noise, and other relevant factors \cite{li2022disentanglement}.

Disentanglement for dynamical systems has not been studied as extensively but previous research demonstrated VAEs can fully recover the parameters of a physical system in their latent space \cite{Iten2018DiscoveringNetworks}. This is something we also corroborate in this work, where we see that the latent space of the plain VAE contains almost all the information of the system parameters (as indicated by Informativeness on \cref{tab:resunts_disentanglement_metrics}). This is a strong indication that parameter inference is very important for prediction and models learn it implicitly. Latent space supervision acts as a regularizer, making the parameter inference task explicit. If this leads to disentangled representations, and results in  \cref{tab:resunts_disentanglement_metrics} suggests it does, the increased prediction performance should be expected. 

\subsection{Theoretical motivation}
\label{sec:app:theoretical}

Disentanglement can also be motivated by a probabilistic view of the evolution of dynamical systems. We assume a class, $C$, of a deterministic dynamical system, $S_C$, parameterized by unobserved parameters $\bm\xi_C$. Our aim is to predict the evolution of the system state, up to some time in future, $t+n$, given a number of observations of the system state up to some point in time $t$. 
The initial conditions of the dynamical system, $\bm{I}_C$, are also unobserved at inference, and constitute a form of uncertainty. Given this form of uncertainty, we can consider the inference problem under a probabilistic framework as estimating the distribution $P(\mathbf{x}_{t:t+n} | \mathbf{x}_{<t}; \bm\xi_C, \bm{I}_C)$ where $\bm\xi_C, \bm{I}_C$ are not observed. Our best options for solving the prediction problem are:
\begin{itemize}
    \item Assign priors on $\bm\xi_C, \bm{I}_C$, and marginalize over them to obtain an estimate of the marginal 
$P(\mathbf{x}_{t:t+n} \mid  \mathbf{x}_{<t})$
    \item  Estimate $\bm\xi_C$ and $\bm{I}_C$ and directly model the conditional
\end{itemize}

Both of these two approaches can be modelled with neural networks, but given the wide nature of divergence in the trajectories of a system for different $\bm\xi_C, \bm{I}_C$ (they may be considered quasi-chaotic),  it is hard to both  assign a proper prior and efficiently marginalize. If, on the other hand, the model can identify $\bm\xi_C, \bm{I}_C$ correctly, we'd be better of with the second modelling choice. Disentanglement can help with better system identification.

To see how disentanglement can be beneficial in this case, we consider the nature of the probability distribution $P(\mathbf{x}_{t:t+n}|\mathbf{x}_{<t})$ and illustrate it with an example on the simple pendulum. For this example, we assume that $\bm\xi_C=l$ the length of the pendulum, while all other parameters and initial conditions are constant. The marginal $P(\mathbf{x}_{t:t+n}|\mathbf{x}_{<t})$ remains unknown, but if we condition on pendulum length, $P(\mathbf{x}_{t:t+n}|\mathbf{x}_{<t}, l)$ is a Gaussian distribution (since the model is deterministic and we assume Gaussian observation noise). In VAE terms this procedure is modelled by the decoder as $P( \mathbf{x}_{t+n}| \mathbf{z}_{<t})$. With supervised disentanglement we can separate the latent vector in two parts (i) $ \mathbf{z}_{<t}$, which captures the dynamics, and (ii) $\mathbf{z}_l$, which captures the information about the pendulum length. This leads to a conditional distribution $P(\mathbf{x}_{t:t+n}|  \mathbf{z}_{<t},  \mathbf{z}_l)$ which better resembles the functional structure of the real conditional distribution. Assuming that the model is able to capture well the deterministic dynamics after training, this should be a better modelling choice and increase prediction performance.

% \textcolor{orange}{Parameter-Conditioned Sequential Generative Modeling of Fluid Flows, Jeremy Morton et al.}

\section{Training objective}

\subsection{Derivation of SD-VAE loss}
\label{sec:vae_loss}
% We will first derive the general form of \cref{eq:sdvae_loss} without the dependence on time and then explain. \
VAEs are trained by maximizing the Evidence Lower Bound 
over the dataset:
$$ 
\operatorname{ELBO} := \mathbb{E}_{\mathbf{x}}\left[\mathbb{E}_{q_{\phi}(\mathbf{z} \mid \mathbf{x})}\left[\log p_{\theta}(\mathbf{x} \mid \mathbf{z})\right]-D_{\mathrm{KL}}\left(q_{\phi}(\mathbf{z} \mid \mathbf{x}) \| p(\mathbf{z})\right)\right]
$$

We can enforce a structure on the latent space of VAEs using constrained optimization. Rewriting the objective in the Langragian form, under the Karush-Kuhn-Tucker conditions, the constrains become regularization terms. The majority of the methods using this approach can be subsumed in the following objective (see \citet{Tschannen2018RecentLearning} for a comprehensive review): 

$$\operatorname{ELBO}(\phi, \theta)+\beta \mathbb{E}_{\mathbf{x}} R_{1}\left(q_{\phi}(\mathbf{z} \mid \mathbf{x})\right)+\delta \mathbb{E}_{\mathbf{x}, \mathbf{z}} R_{2}\left(q_{\phi}(\mathbf{z} \mid \mathbf{x}), \mathbf{z}\right)
$$

We use $R_1 = D_{\mathrm{KL}}\left(q_{\phi}(z \mid x) \| p(z)\right)$ a common choice for enabling unsupervised disentanglement that was originally proposed in beta-VAE \citep{Higgins2017Beta-VAE:FRAMEWORK}. Contrary to many other approaches for the second regularizer we use a supervised term $R_2 = \mathcal{L}_{\bm\xi}(\mathbf{z}_{1:N_{\mathbf{\xi}}}, \bm\xi) = \| \mathbf{z}_{1:N_{\mathbf{\xi}}} - \bm\xi \|_2$ where $\mathbf{\xi}$ the real ODE parameters as described in \cref{sec:methods_vaes_for_modelling_dynamics}. While many dynamical VAE methods use a different latent for each time step \citep{Li2018DisentangledAutoencoder}, our model can be seen as performing multi-step prediction from a single latent vector. Putting the above together we arrive at the formulation of \cref{eq:sdvae_loss}:

\begin{equation*}
\begin{aligned} 
\mathcal{L}_{\phi, \theta}(\mathbf{x_{\leq t}})&= \mathbb{E}_{Q_{\phi}(\mathbf{z} \mid \mathbf{x}_{\leq t})}\left[\log P_{\theta}(\mathbf{x}_{t<,\leq t+n}\mid \mathbf{z}) + \delta \mathcal{L}_{\bm{\xi}}(\mathbf{z}_{1:N_{\bm{\xi}}}, \bm{\xi})\right] -\beta D_{KL}\left(Q_{\phi}(\mathbf{z} \mid \mathbf{x}_{\leq t})|| P(\mathbf{z})\right)
\end{aligned}
\end{equation*}

\subsection{Scaling of the parameter} 
\label{sec:scaling_parameters}

For our experiments (both in phase and observation space) we scale the ground-truth parameter in the $[0,1]$ range:
 \begin{equation}
\hat\xi_i = \frac{\xi - \min(\xi_i)}{\max(\xi_i) - \min(\xi_i)}
\end{equation}

where $\xi_i$ are the domain parameters and their corresponding minimum and maximum values of domain parameters from the training set. During training we use the output of $\hat\xi$ as the target for the regression loss.

\subsection{SD-RSSM loss}
\label{sec:sd-rssm-loss}
The SD-RSSM is built upon the original RSSM with the addition of regression loss term which enhances the latent space disentanglement. Since the RSSM has latent variables for each time-step, we apply a disentanglement loss on all of them.

\begin{equation}
\begin{aligned}
\mathcal{L}_{SD-RSSM}(\bm{o}_{\leq t})=&\sum_{t=1}^{T}\left(\underbrace{\mathbb{E}_{q(\bm{s_{t}} \mid \bm{o_{\leq t})}}[\ln p(\bm{o_{t}} \mid \bm{s_{t})})]}_{\text{reconstruction}}- \right. \nonumber
\underbrace{
\mathbb{E}_{q(\bm{s_{t-1}} \mid \bm{o_{\leq t-1}})}\left[\mathrm{KL}[q(\bm{s_{t} }\mid \bm{o_{\leq t}}) \| p(\bm{s_{t}} \mid \bm{s_{t-1}})]\right]
}_{\text{prediction}}  %\nonumber
\\
&\left.+\underbrace{\delta \mathbb{E}_{q(\bm{s_{t}} \mid \bm{o_{\leq t}})}\left[\left\|
\bm{\xi} - \bm{s_{t}}^{(1:N_{{\xi}})}  \right\|_{2}
\right]}_{\text{supervised disentanglement loss}}\right)
\end{aligned}
\end{equation}

Where $\bm{o_t}$ is the observations, $\bm{s_t}$ the stochastic latent variables at time $t$, $\bm{\xi}$ are the $k$ dimensional domain parameters and $\delta$ tunes the supervised disentanglement strength.

\section{Training and hyperparameters}

\subsection{Phase Space Experiments}
\label{sec:hyperparameters_phase_space}

Typically our training sequences are at least 1000 steps long. As a form of data augmentation, for each batch we select a random starting point $t$ within the sequence.

An Adam optimizer with $b_1=0.9$ and $b_2=0.999$ and a scheduler for the learning rate are employed. The maximum number of epochs was set to 2000 but we also do early stopping using a validation set which led to significantly less epochs. 

\begin{table*}[!htb]
\caption{\textbf{Pendulum hyperparameters}} 
\label{tab:hp-pendulum}
  \centering
  \makebox[\textwidth]{
%\small
\begin{tabular}{lccccc}
    \toprule
 & AE & SD-AE & VAE & SD-VAE & LSTM \\
      \cmidrule(r){2-6}
Input Size & \multicolumn{5}{c}{10, 50} \\
Output Size & \multicolumn{5}{c}{1, 10} \\
Hidden Layers & \multicolumn{4}{c}{{[}400, 200{]}} &  50,100,200 \\
Latent Size & \multicolumn{4}{c}{4, 8, 16} & -  \\
Nonlinearity & \multicolumn{4}{c}{Leaky ReLU} & Sigmoid \\
Num. Layers & - & - & - & -  & 1,2,3\\ 
Learning rate &  \multicolumn{5}{c}{$10^{-3}$}  \\
Batch size & 16, 32& 16 & 16, 32 & 16 & 16, 64  \\
Sched. patience & 20, 30, 40 & 20,30 & 20 & 20 & 30  \\
Sched. factor & 0.3 & 0.3 & 0.3 & 0.3 & 0.3 \\
Gradient clipping & No & 1.0 & 1.0  \\
Layer norm (latent) & No & No & Yes & Yes & No\\
Teacher Forcing &-&-&-&-& Partial \\
Decoder $\gamma$ & - & - & $10^{-3}, 10^{-4}, 10^{-5}$ & $10^{-3}, 10^{-4}$ & - \\
Sup. scaling & - & Linear & - & Linear & -   \\
Supervision $\delta$ & - & 0.1, 0.2, 0.3 & - & 0.01, 0.1, 0.2 & -\\
\cmidrule(r){2-6}
\# of experiments & 72 & 72 & 72 & 72 & 72\\
\bottomrule
\end{tabular}}
\end{table*}

\FloatBarrier

\begin{table*}[ht]
\caption{\textbf{Lotka-Volterra hyperparameters}} 
\label{tab:hp-lv}
  \centering
  \makebox[\textwidth]{
%\small
\begin{tabular}{lcccccc}
    \toprule
 & AE & SD-AE & VAE & SD-VAE & LSTM \\
      \cmidrule(r){2-6}
Input Size & \multicolumn{5}{c}{50} \\
Output Size & \multicolumn{5}{c}{10} \\
Hidden Layers & \multicolumn{4}{c}{{[}400, 200{]}} &  50,100 \\
Latent Size & \multicolumn{4}{c}{8, 16, 32} & -  \\
Nonlinearity & \multicolumn{4}{c}{Leaky ReLU} & Sigmoid \\
Num. Layers & - & - & - & -  & 1,2,3\\ 
Learning rate &  $10^{-3}, 10^{-4}$ &  $10^{-3}, 10^{-4}$ & $10^{-3}, 10^{-4}$ & $10^{-3}$ &  $10^{-3}$\\
Batch size & 16, 32, 64 & 16, 32 & 16, 32 & 16 & 10, 64, 128 \\
Sched. patience & 20, 30 & 20, 30 & 20 & 20 & 20, 30  \\
Sched. factor & 0.3, 0.4 & 0.3 &  0.3 & 0.3 & 0.3\\
Gradient clipping & No & No & 0.1, 1.0 & 0.1, 1.0 & No  \\
Layer norm (latent) & No & No & No & No & No \\
Teacher Forcing &-&-&-&-& Partial, No \\
Decoder $\gamma$ & - &  - & $10^{-4}, 10^{-5}, 10^{-6}$ & $10^{-4}, 10^{-5}, 10^{-6}$  &  - \\ % exps for 10^-6 in vae did not run 
Sup. scaling & -  & Linear & - & Linear  &  - \\
Supervision $\delta$ & - & 0.1, 0.2, 0.3 & - & 0.01, 0.1, 0.2, 0.3 & -\\
\cmidrule(r){2-6}
\# of experiments & 72 & 72 & 72  & 72 & 72\\
\bottomrule
\end{tabular}}
\end{table*}

\FloatBarrier

\begin{table*}[!htb]
\caption{\textbf{3-body system hyperparameters}} 
\label{tab:hp-3body}
\centering
\makebox[\textwidth]{
\small
\begin{tabular}{lccccc}
\toprule
 & AE & SD-AE & VAE & SD-VAE & LSTM \\
      \cmidrule(r){2-6}
Input Size & \multicolumn{5}{c}{50} \\
Output Size & \multicolumn{5}{c}{10} \\
Hidden Layers & \multicolumn{4}{c}{{[}400, 200{]}} &  50,100 \\
Latent Size & \multicolumn{4}{c}{8, 16, 32} & -  \\
Nonlinearity & \multicolumn{4}{c}{Leaky ReLU} & Sigmoid \\
Learning rate &  $10^{-3}, 10^{-4}$ & $10^{-3}, 10^{-4}$ & $10^{-3}, 10^{-4}$ & $10^{-3}$ \\
Batch size & 16, 32 & 16 & 16 & 16 & 16, 64, 128 \\
Sched. patience & 30, 40, 50, 60 & 30, 40, 50, 60  & 30, 40, 50, 60 & 30, 40, 50, 60 & 20, 30  \\
Sched. factor & 0.3, 0.4 & 0.3 & 0.3, 0.4 & 0.3, 0.4 & 0.3\\
Gradient clipping & No & No & No & No & No  \\
Layer norm (latent) & No & No & No  & No& No \\
Decoder $\gamma$ & - &  - &  $10^{-5}, 10^{-6}$ & $10^{-5}, 10^{-6}$ & -  \\
Sup. scaling & - & Linear & - & Linear, Sigmoid & -\\
Supervision $\delta$ & - & 0.05, 0.1, 0.2, 0.3 & - & 0.1, 0.2 & - \\
\cmidrule(r){2-6}
\# of experiments & 96 & 96 & 96 & 96 & 96\\
\bottomrule
\end{tabular}
}
\end{table*}
\FloatBarrier

\begin{table}[ht]
  \caption{\textbf{Number of experiments with phase space data.} Each experiment corresponds to a distinct configuration of hyperparameters.}
  \label{tab:numexperiments}
\centering
\makebox[\textwidth]{
\small
\begin{tabular}{lcccccccc}
    \toprule
 & AE & SD-AE & VAE & SD-VAE & LSTM & Total \\
     \cmidrule(r){1-7}
Pendulum & 72 & 72 & 72 &  72 & 72 & 360 \\
L-V & 72 & 72 & 72 &  72  &  72 & 360 \\
3-body & 96 & 96 & 96 &  96 & 96 &  480 \\
     \cmidrule(r){1-7}
 & \multicolumn{3}{l}{} & \multicolumn{2}{r}{Total experiments} & 
 \textbf{1200}\\
     \bottomrule
\end{tabular}}
\end{table}

\FloatBarrier

\subsection{Observation-space experiments}
\label{sec:hyperparameters_observation_space_pendulum}

\begin{table*}[!htb]
\centering
\caption{\textbf{Hyperparameters for the RSSM and SD-RSSM models}} 
\makebox[\textwidth]{
% \small
\begin{tabular}{lcc}
\toprule
& RSSM & SD-RSSM  \\
\cmidrule(r){2-3} 
Batch Size  & \multicolumn{2}{c}{50, 100}   \\
Decoder std. &  \multicolumn{2}{c}{$1.0, 2.0$} \\
Train Input Length & \multicolumn{2}{c}{50, 100}  \\
Supervision $\delta$ & - &  $0.01, 0.1, 1$ \\
Seeds & 3 & 1 \\
\cmidrule(r){2-3}
\# of experiments & 24 & 24\\
\bottomrule
\end{tabular}
}
\label{tab:hp-rssm}
\end{table*}
% \FloatBarrier

\section{Additional Results for Phase Space Experiments}
\label{sec:results_extra_phase_space}

\begin{table*}[!ht]
\centering
    \caption{\textbf{SD-VAE exhibits stronger disentanglement properties than the plain VAE according to many common metrics}. Figures are computed over the best 3 models.}
\begin{tabular}{lcc|cc|cc}
\toprule
     & \multicolumn{2}{c}{Pendulum} & \multicolumn{2}{c}{Lotka-Volterra} & \multicolumn{2}{c}{3 body system} \\
     &                VAE &             SD-VAE &                VAE &             SD-VAE &                VAE &             SD-VAE \\
\midrule
Disentanglement  &    - &   - &  $0.27  \pm  0.06$ &  $\mathbf{0.53  \pm  0.06}$ &  $0.20  \pm  0.00$ &  $\mathbf{0.90  \pm  0.00}$ \\
Completeness &   $0.17  \pm  0.06$ &  $\mathbf{0.90  \pm  0.00}$ &  $0.20  \pm  0.00$ &  $\mathbf{0.57  \pm  0.06}$ &  $0.13  \pm  0.06$ &  $\mathbf{0.90  \pm  0.00}$ \\
Informativeness  &  $0.94  \pm  0.01$ &  $\mathbf{0.99  \pm  0.00}$ &  $\mathbf{1.00  \pm  0.00}$ &  $\mathbf{1.00  \pm  0.00}$ &  $\mathbf{1.00  \pm  0.00}$ &  $\mathbf{1.00  \pm  0.00}$ \\
SAP  &  $0.03  \pm  0.04$ &  $\mathbf{0.87  \pm  0.02}$ &  $0.04  \pm  0.01$ &  $\mathbf{0.21  \pm  0.04}$ &  $0.01  \pm  0.01$ &  $\mathbf{0.67  \pm  0.04}$ \\
MIG  &  $0.01  \pm  0.00$ &  $\mathbf{0.17  \pm  0.01}$ &  $0.00  \pm  0.00$ &  $\mathbf{0.03  \pm  0.00}$ &  $0.00  \pm  0.00$ &  $\mathbf{0.08  \pm  0.01}$ \\
\bottomrule
\end{tabular}
\end{table*}

\begin{table*}[!ht]
\centering
\caption{\textbf{Model stability} Percentage of models that diverge during testing in all some trajectories (out of all the trained models)}
\label{tab:results_divergent_models}
\begin{tabular}{lccc}
\toprule
{} & Pendulum & Lotka-Volterra & 3-body system \\
\midrule
LSTM   &      86\%&           100\% &           53\% \\
AE    &      42\%&            14\% &           50\% \\
SD-AE &      69\% &            29\% &           58\% \\
VAE    &       3\% &            10\% &           15\% \\
SD-VAE &       2\% &            48\% &            9\% \\
\bottomrule
\end{tabular}
\end{table*}

\FloatBarrier

\begin{figure}[!htb]
  \centering
   \caption{\textbf{Mean Absolute Error (MAE)} between model predictions and ground-truth trajectories.}  

    \textbf{Pendulum} \par

        \hspace{0.0cm}

  \begin{subfigure}{.33\linewidth}
    \centering    
\includegraphics[width=\linewidth]{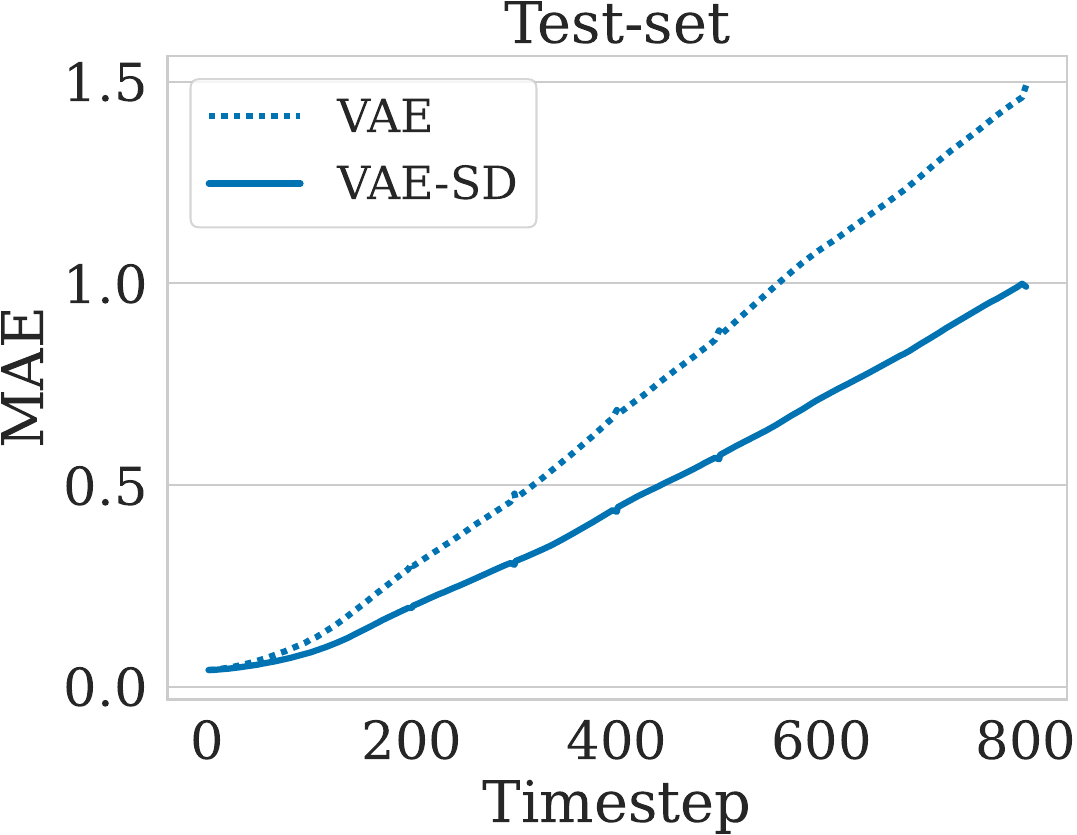}
  \end{subfigure}
  \hfill
    \begin{subfigure}{.33\linewidth}
    \centering
  \includegraphics[width=\linewidth]{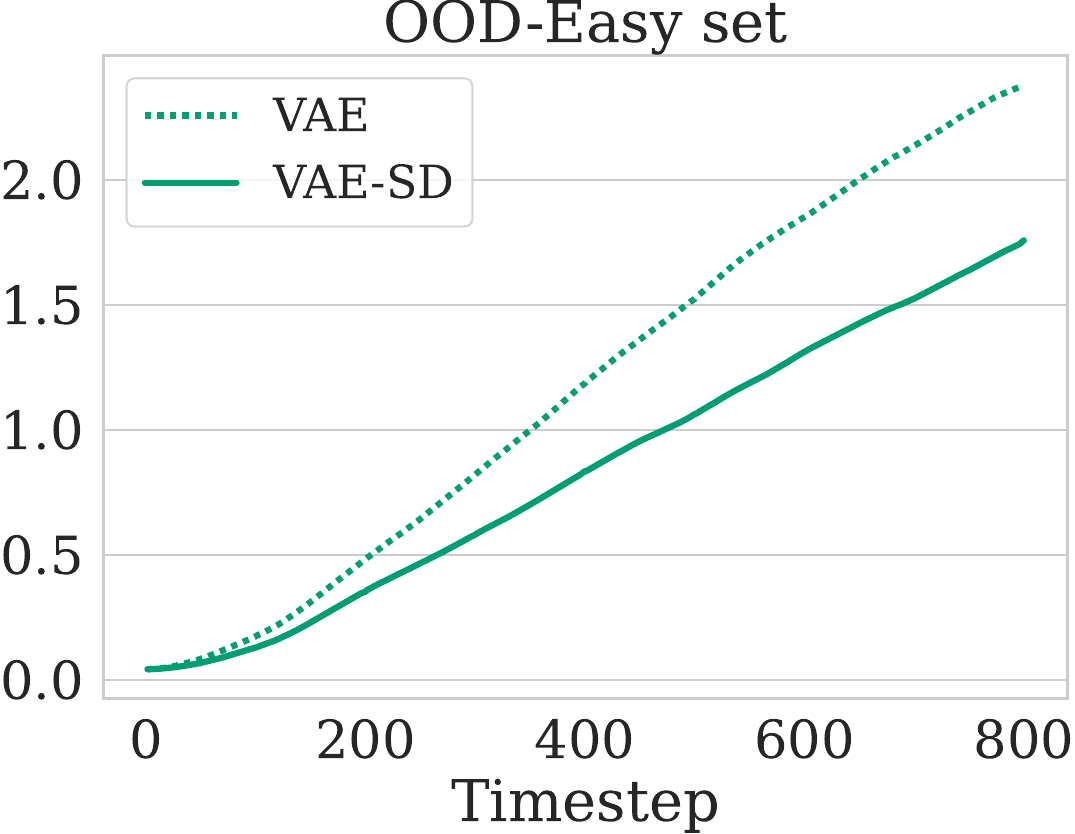}
  \end{subfigure}
    \hfill
    \begin{subfigure}{.33\linewidth}
    \centering
  \includegraphics[width=\linewidth]{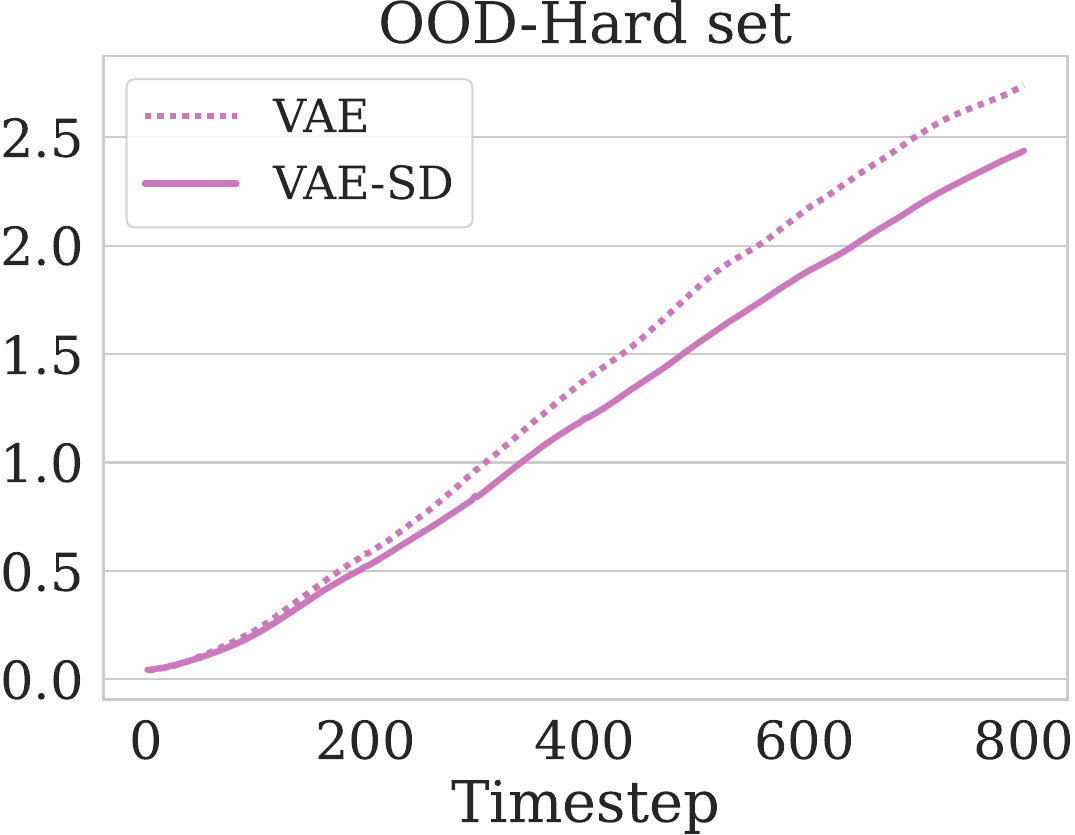}
\end{subfigure}

    \hspace{1.0cm}

\textbf{ Lotka-Volterra} \par

        \hspace{0.0cm}

  \begin{subfigure}{.33\linewidth}
    \centering
\includegraphics[width=\linewidth]{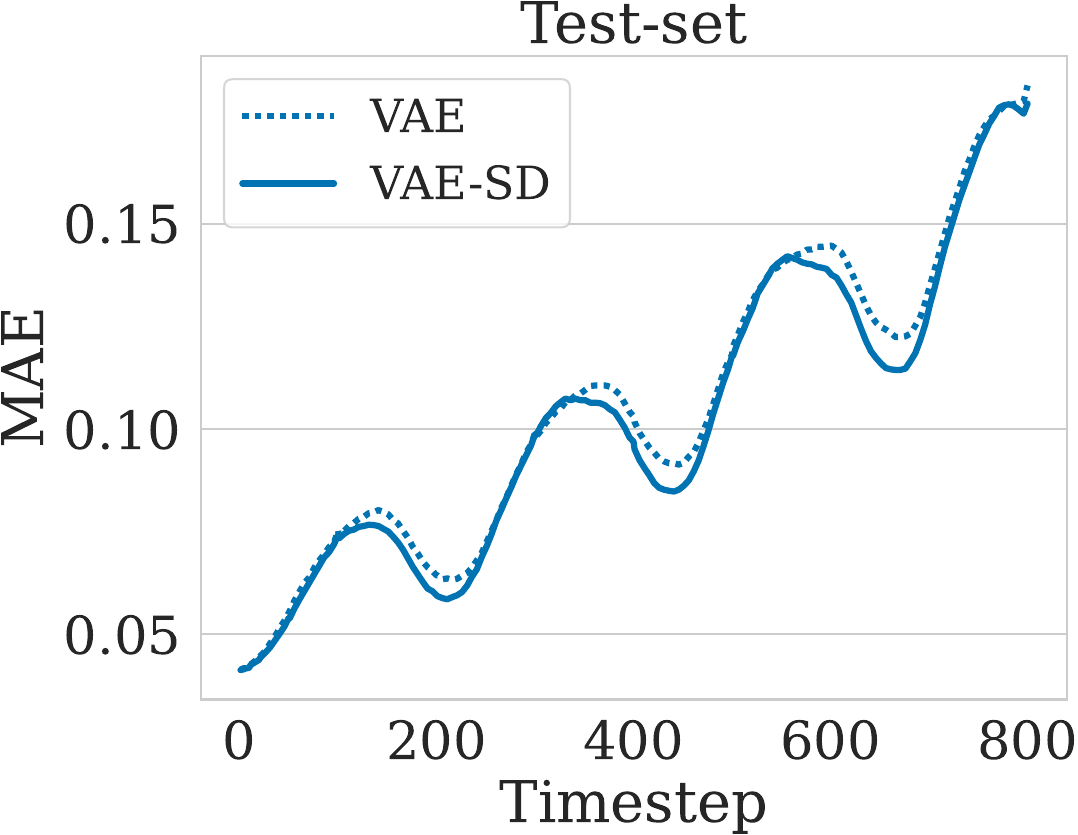}
  \end{subfigure}
  \hfill
    \begin{subfigure}{.33\linewidth}
    \centering
  \includegraphics[width=\linewidth]{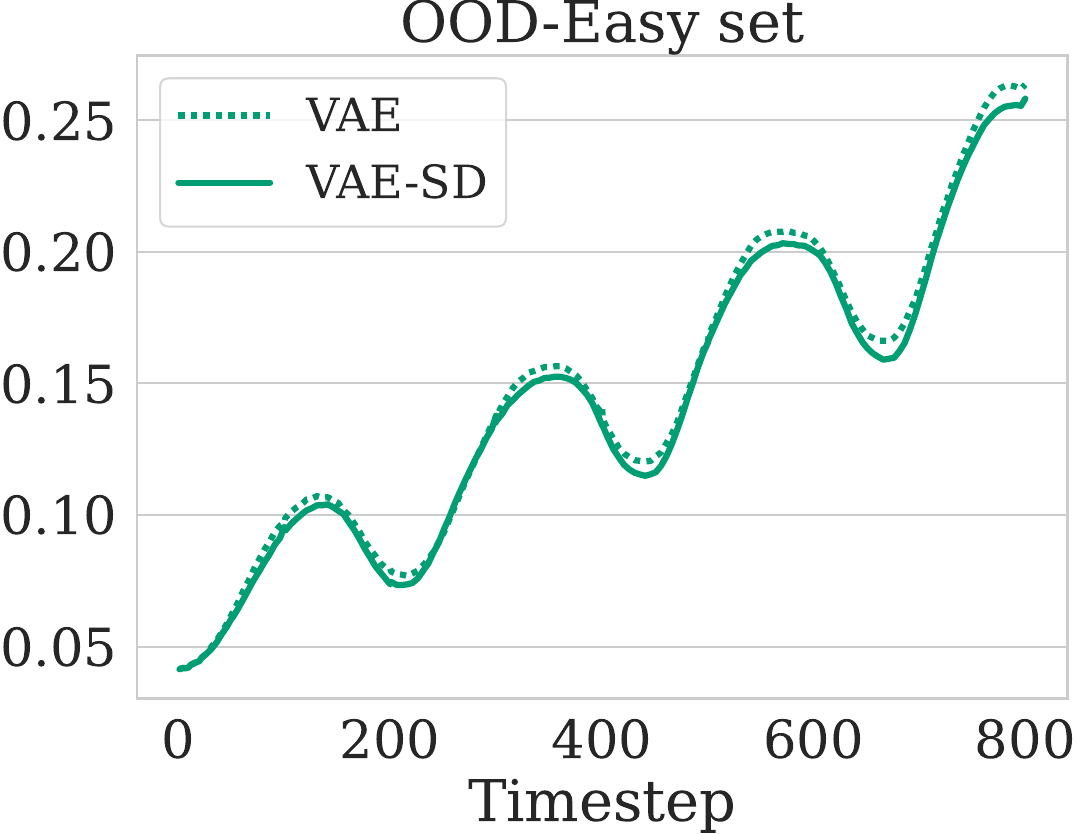}
  \end{subfigure}
    \hfill
    \begin{subfigure}{.33\linewidth}
    \centering
  \includegraphics[width=\linewidth]{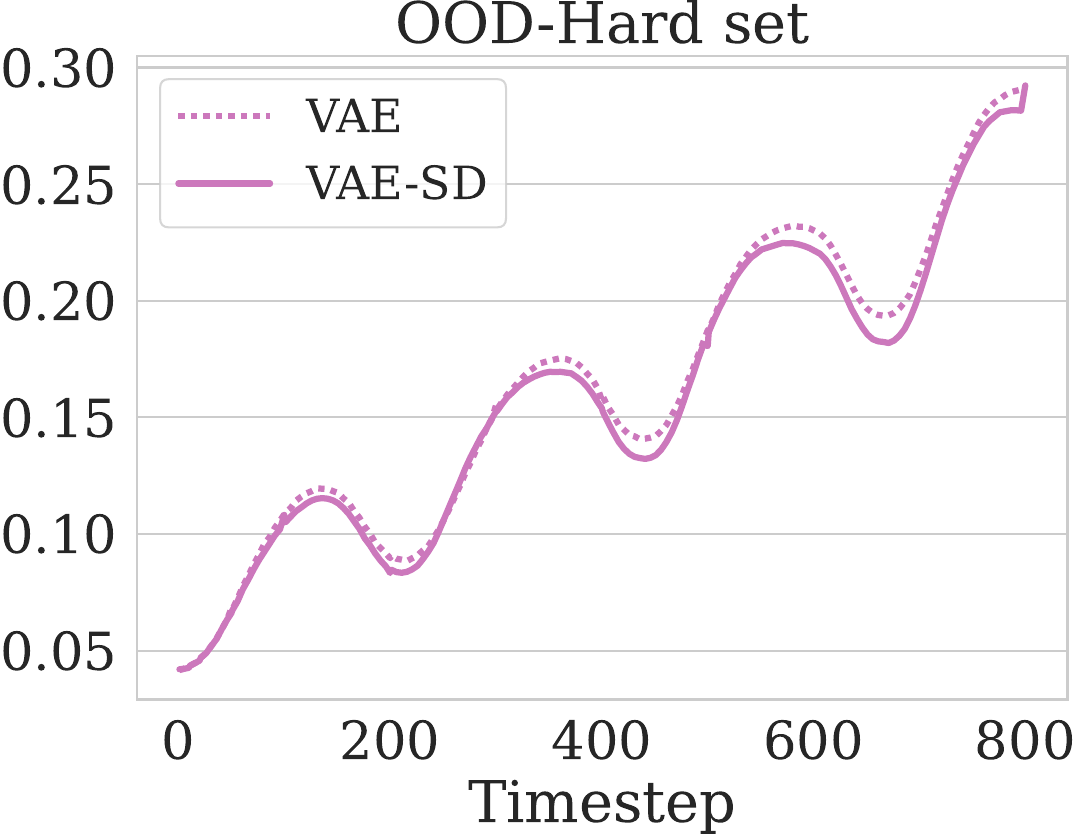}
\end{subfigure}

    \hspace{1.0cm}

\textbf{3 body system} \par

        \hspace{0.0cm}

  \begin{subfigure}{.33\linewidth}
    \centering
\includegraphics[width=\linewidth]{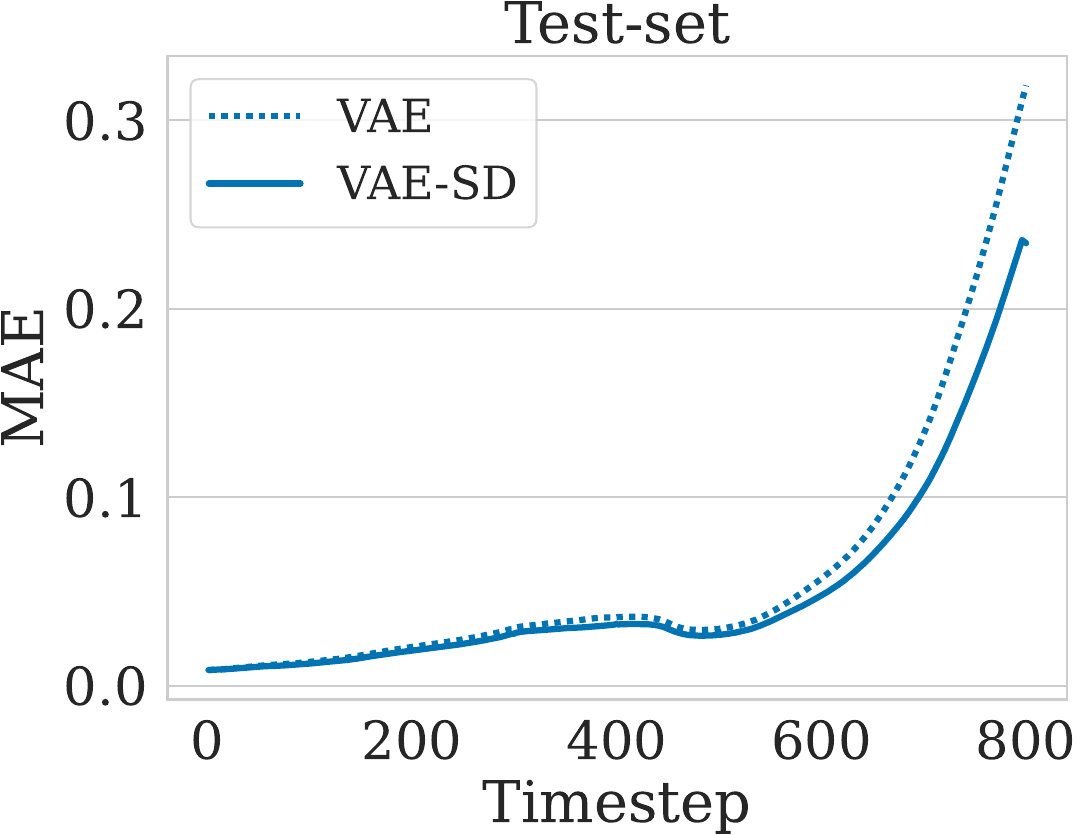}
  \end{subfigure}
  \hfill
    \begin{subfigure}{.33\linewidth}
    \centering
  \includegraphics[width=\linewidth]{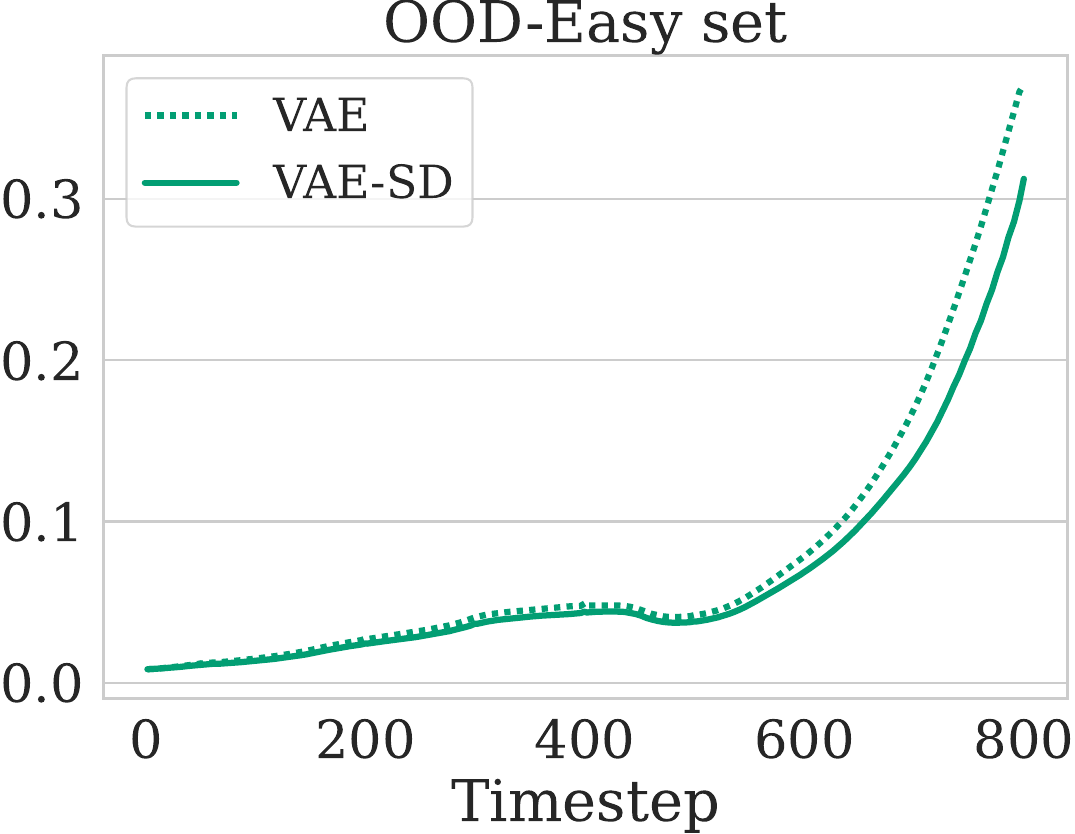}
  \end{subfigure}
    \hfill
    \begin{subfigure}{.33\linewidth}
    \centering
  \includegraphics[width=\linewidth]{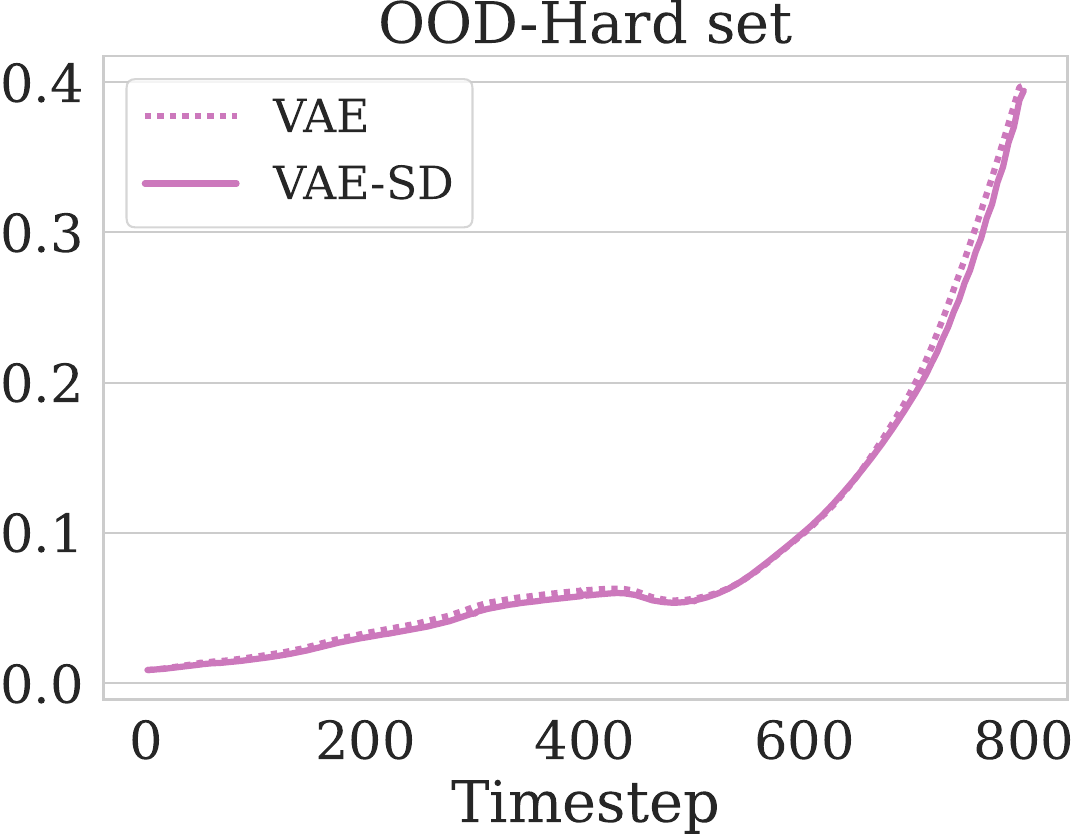}
\end{subfigure}

  \label{fig:results_mae_separate}
\end{figure}

\FloatBarrier

%%%%%%% %%%%%%% %%%%%%% %%%%%%% 
%%%%%%%% RESULTS QUAL  %%%%%%%%
%%%%%%% %%%%%%% %%%%%%% %%%%%%% 
\begin{figure*}[!htb]
% PEND
\begin{subfigure}{.32\linewidth}
\begin{tikzpicture}[spy using outlines={circle,gray,magnification=2.5,size=2.0cm, connect spies}]
\node {\includegraphics[width=\linewidth]{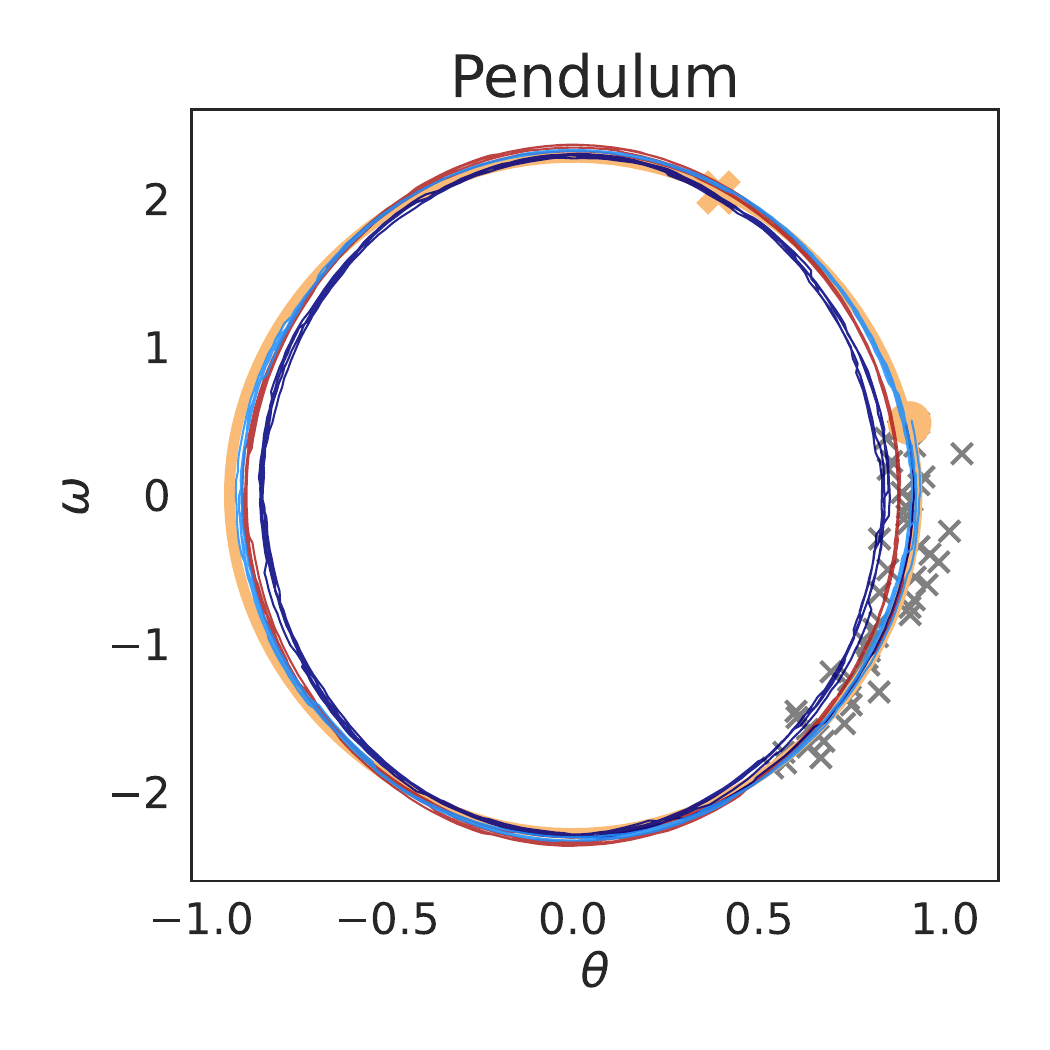}};
\spy on (-1.0,-0.75) in node [left] at (1.2,0.2);
\end{tikzpicture}
\end{subfigure}
\hfill
% LV
\begin{subfigure}{.32\linewidth}
\begin{tikzpicture}[spy using outlines={circle,gray,magnification=2.5,size=2.0cm, connect spies}]
\node {\includegraphics[width=\linewidth]{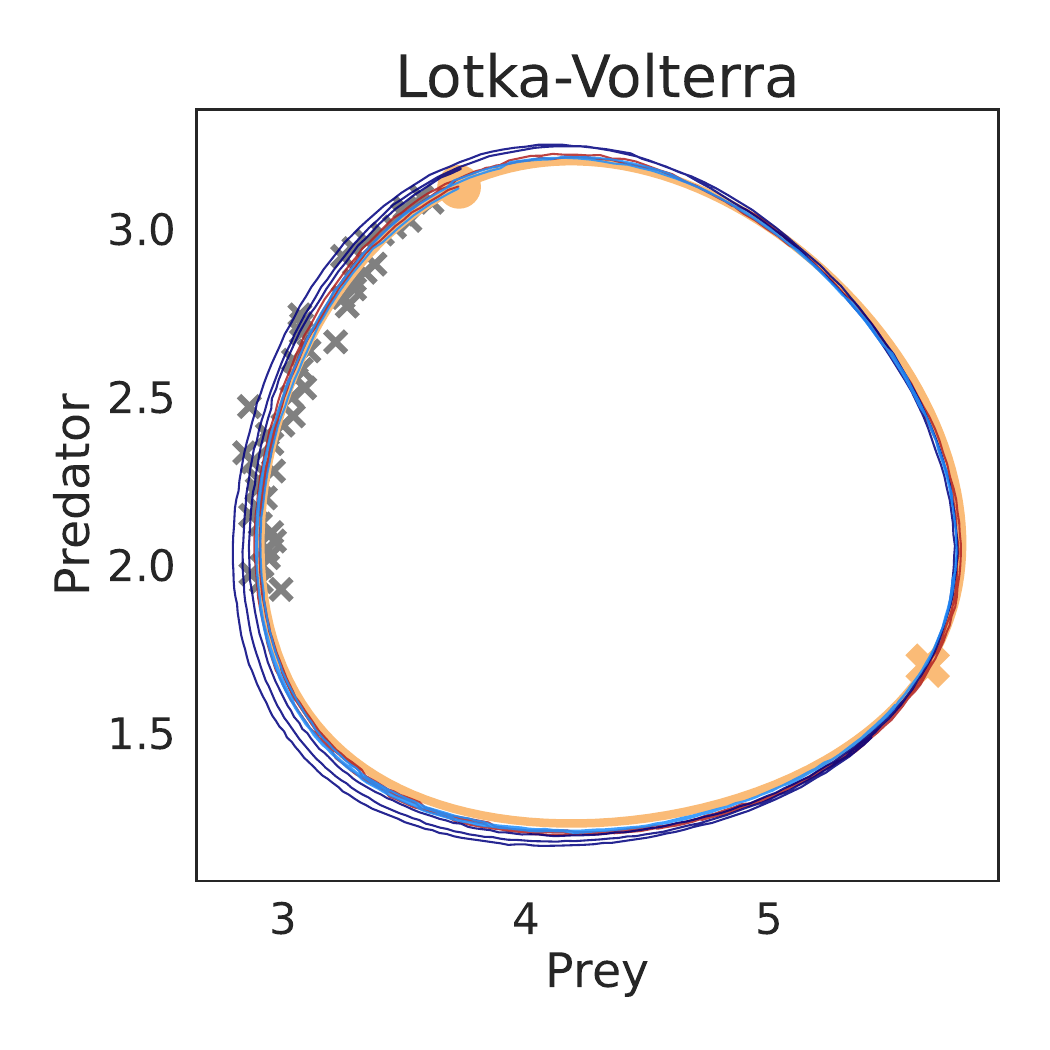}};
\spy on (-1,-0.75) in node [left] at (1.5,0.2);
\end{tikzpicture}
\end{subfigure}
\hfill
% 3BODY
\begin{subfigure}{.32\linewidth}
\begin{tikzpicture}[spy using outlines={circle,yellow,magnification=2.0,size=0cm}]
\node {\includegraphics[width=\linewidth]{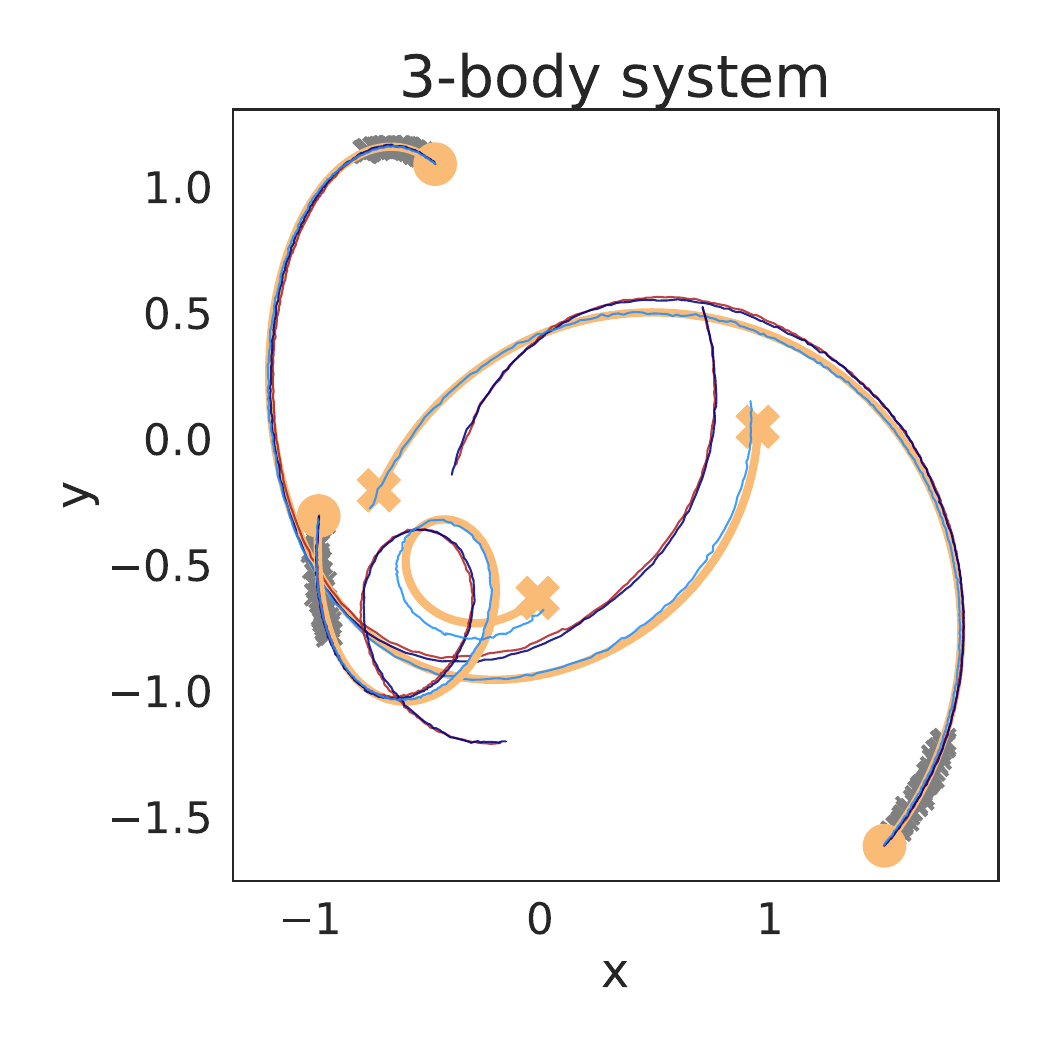}};
\spy on (-0.2,-1.05) in node [left] at (1.7,-0.7);
\end{tikzpicture}
\end{subfigure}

\begin{subfigure}{.32\linewidth}
\begin{tikzpicture}[spy using outlines={circle,gray,magnification=2.5,size=2.0cm, connect spies}]
\node {\includegraphics[width=\linewidth]{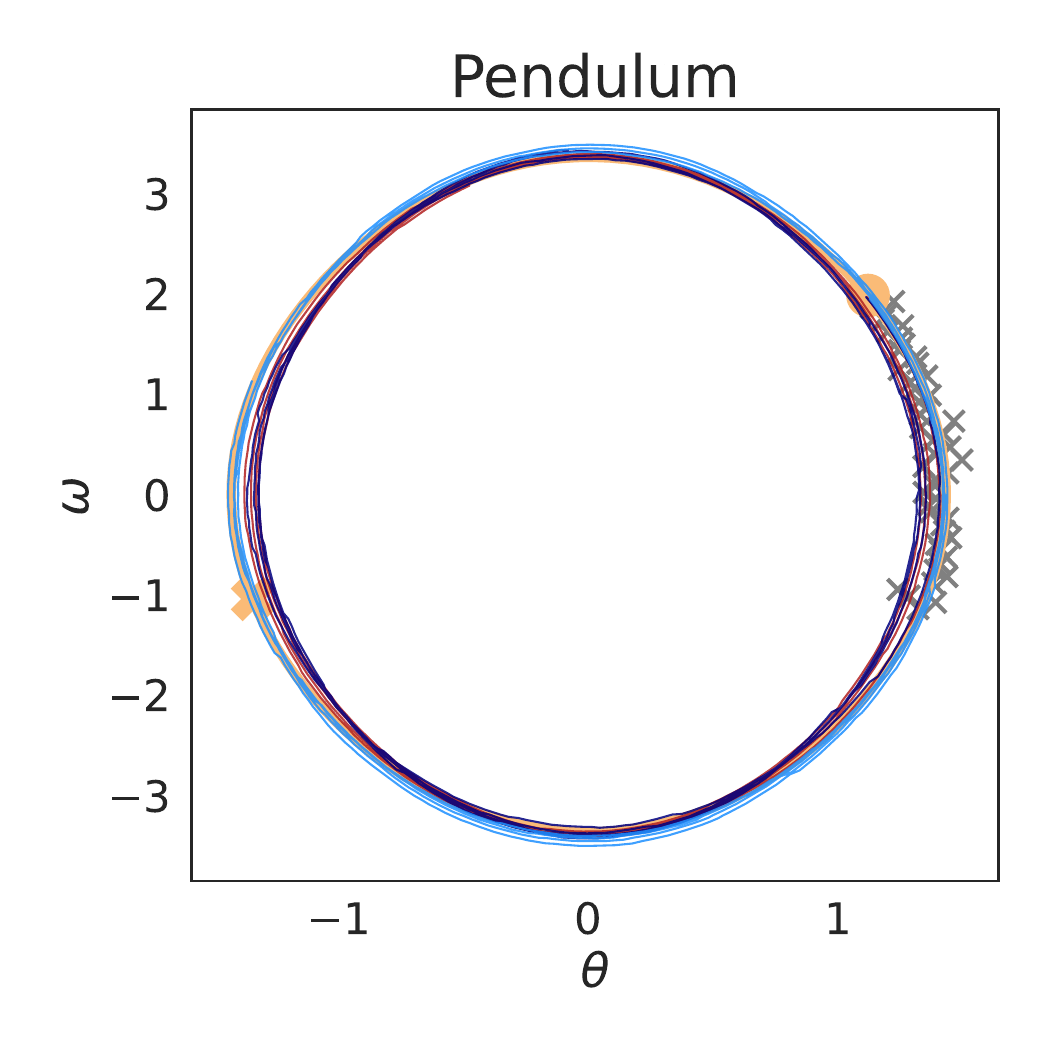}};
\spy on (1.65,-0.25) in node [left] at (1.2,0.2);
\end{tikzpicture}
\end{subfigure}
% LV
\begin{subfigure}{.32\linewidth}
\begin{tikzpicture}[spy using outlines={circle,gray,magnification=2.5,size=2.0cm, connect spies}]
\node {\includegraphics[width=\linewidth]{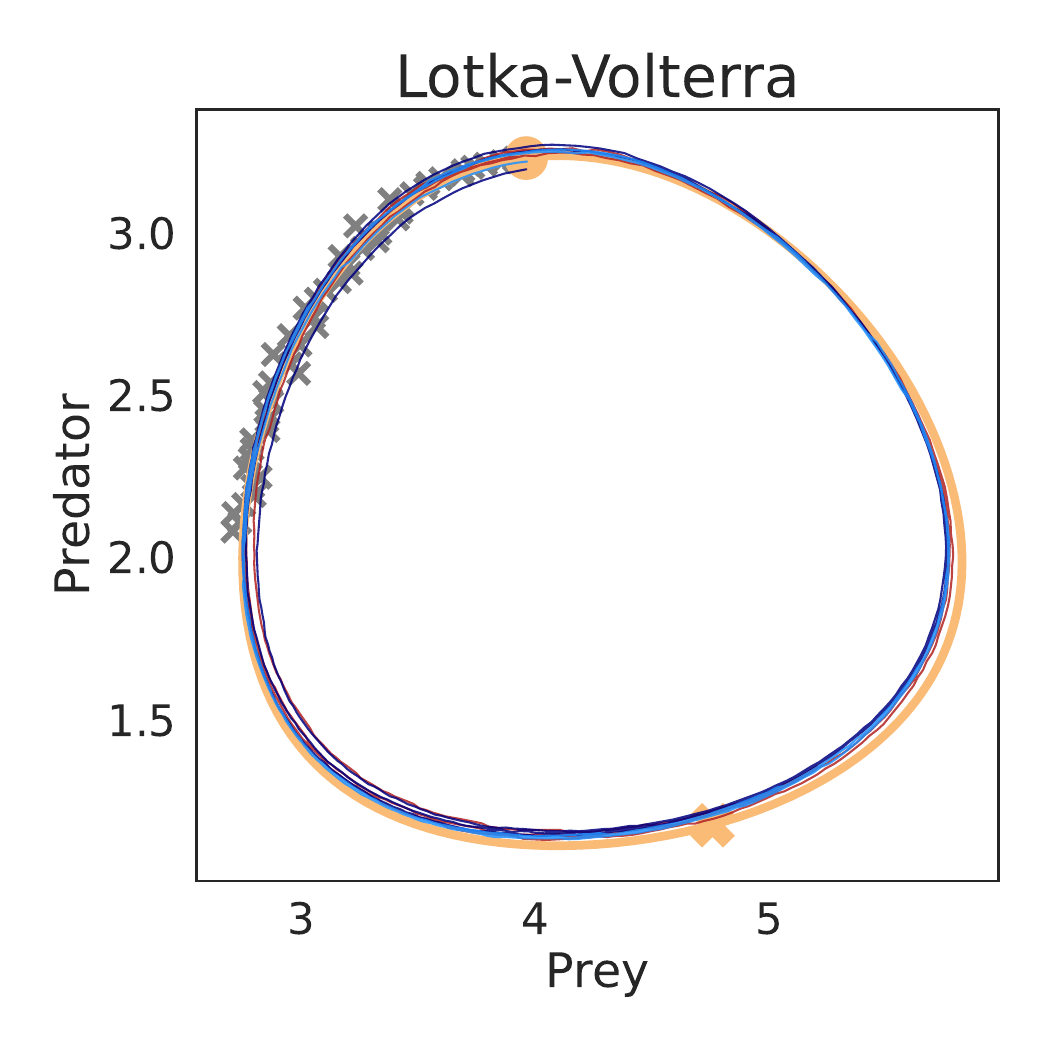}};
\spy on (-1,-0.55) in node [left] at (1.5,0.2);
\end{tikzpicture}
\end{subfigure}
% 3BODY
\begin{subfigure}{.32\linewidth}
\begin{tikzpicture}[spy using outlines={circle,gray,magnification=2.0,size=1.5cm, connect spies}]
\node {\includegraphics[width=\linewidth]{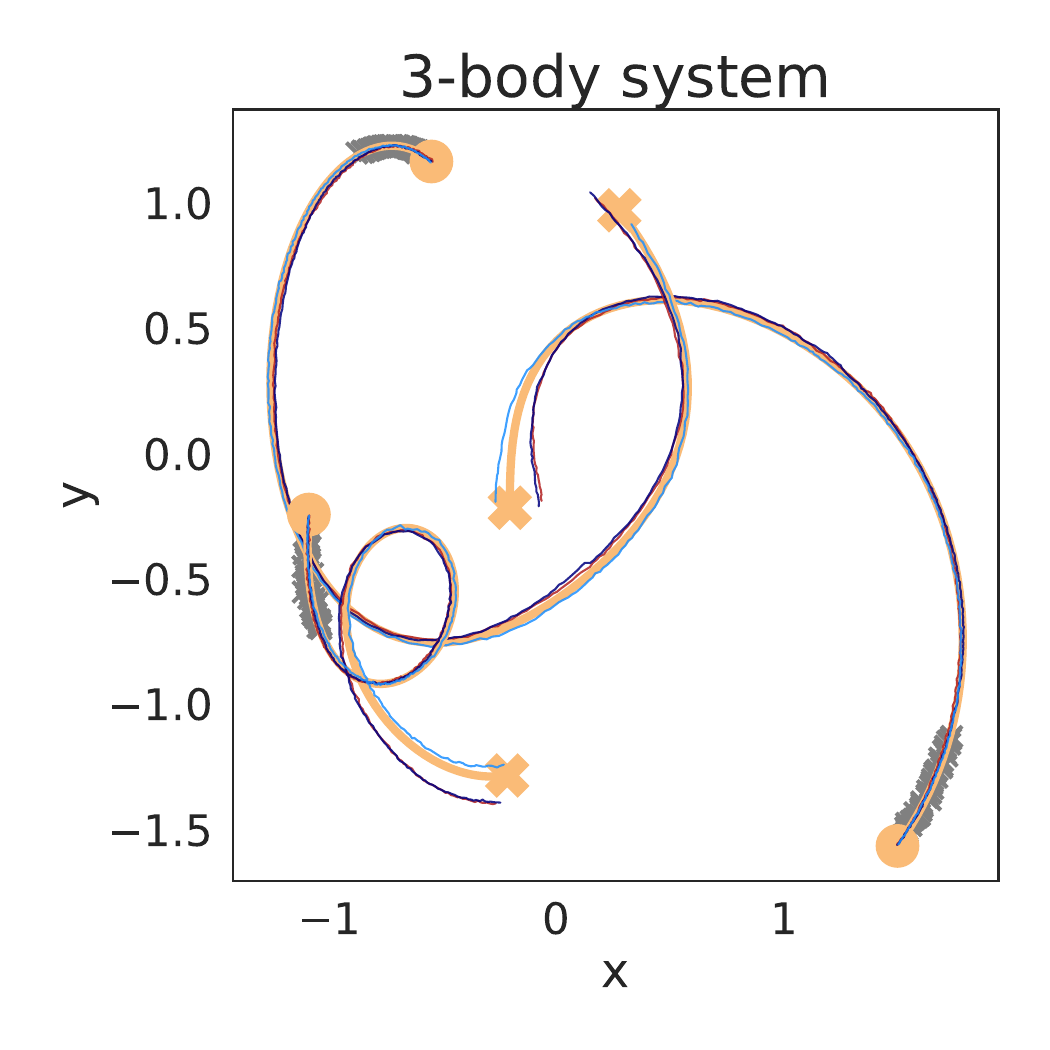}};
\spy on (-0.2,-1.05) in node [left] at (1.7,-0.7);
\end{tikzpicture}
\end{subfigure}
\begin{subfigure}{\linewidth}
\centering
  \includegraphics[width=0.9\linewidth]{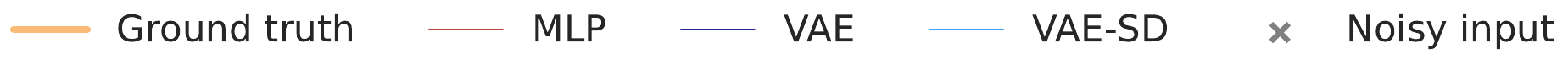}
\end{subfigure}

  \caption{\textbf{Model predictions in phase space}. Trajectories are taken from the OOD-Hard set of each system. The trajectory observations are noisy, denoted by the grey `$\times$'  markers. The orange circle and the orange and bold `$\times$'  markers denote the start and end of the ground-truth trajectories respectively}
  \label{fig:qual-all}
\end{figure*}

\FloatBarrier

\subsection{Linear correlation between $\bm{z}$ and $\bm{\xi}$} 
\label{sec:linear_correlation}
\begin{figure*}[!htb]
\centering
\caption{\textbf{Correlation between latent values and system parameters.} \textbf{Top} The importance weights of a random forest regressor trained to predict the ground-truth parameter values from the latents computed on the training set. High  importance weights indicate a latent variable that has high predictive power over the ground-truth value. \textbf{Bottom} Pearson correlation between (absolute) between the system paramters and the latent variables.}
  \begin{subfigure}{.45\linewidth}
    \centering
\includegraphics[width=1\hsize]{images/latents/importance_weights_truncated.pdf}
\caption{Importance weights. Ordered and truncated latents.}
  \end{subfigure}
\hfill
  \begin{subfigure}{.45\linewidth}
    \centering
\includegraphics[width=1\hsize]{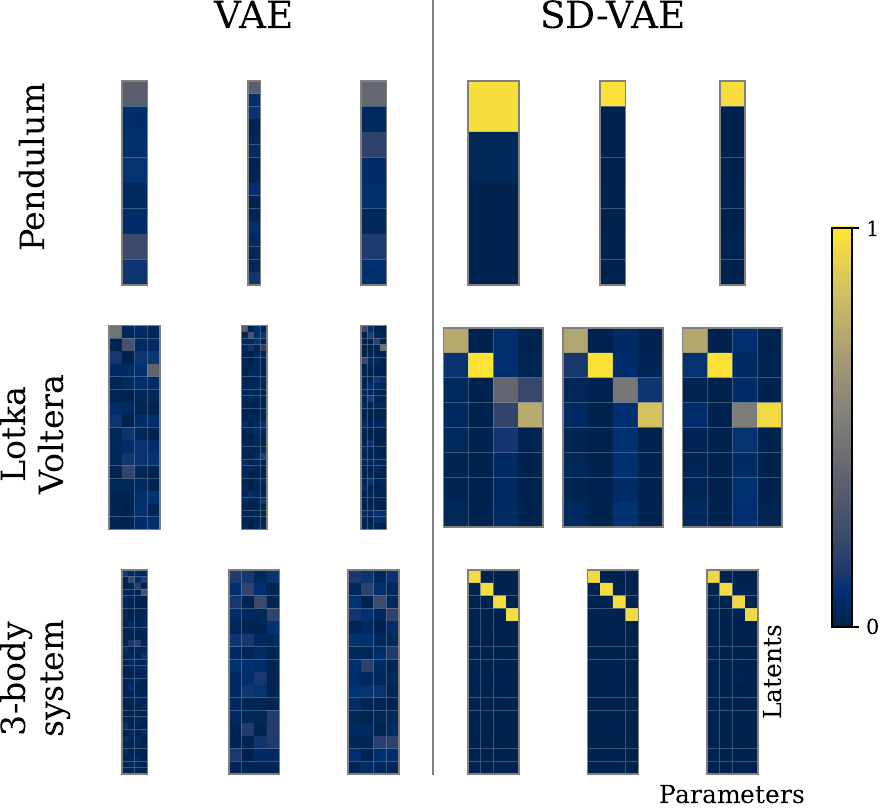}
\caption{Importance weights. All latents.}
  \end{subfigure}
% \vspace{2cm}
    \begin{subfigure}{.45\linewidth}
    \centering
\includegraphics[width=1\hsize]{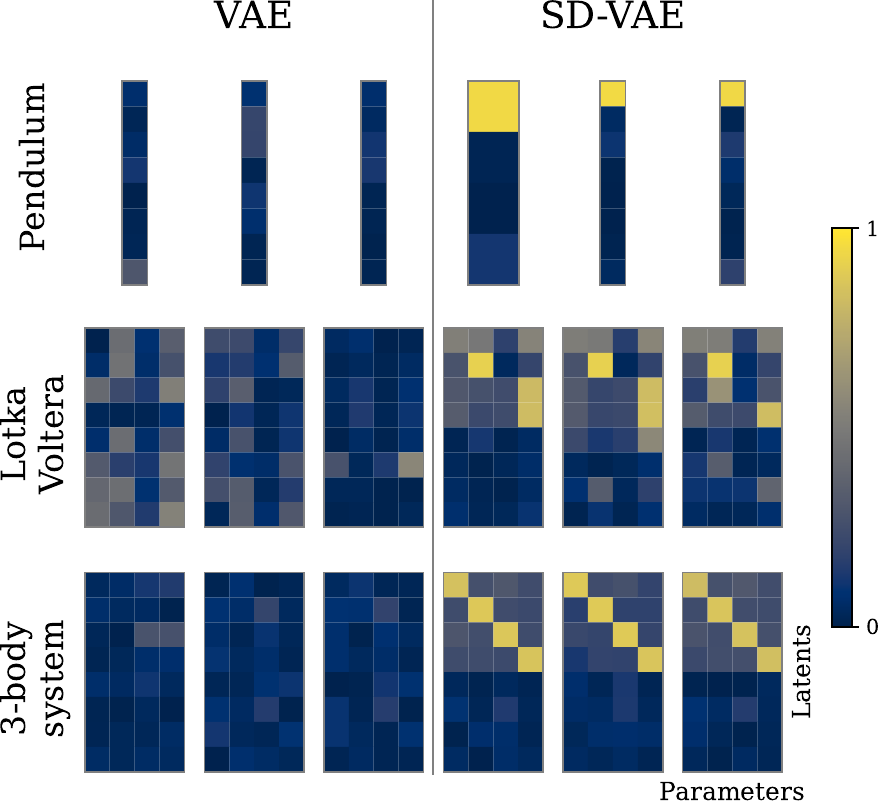}
\caption{Pearson correlation. Ordered and truncated latents.}
  \end{subfigure}
    \hfill
    \begin{subfigure}{.45\linewidth}
    \centering
\includegraphics[width=1\hsize]{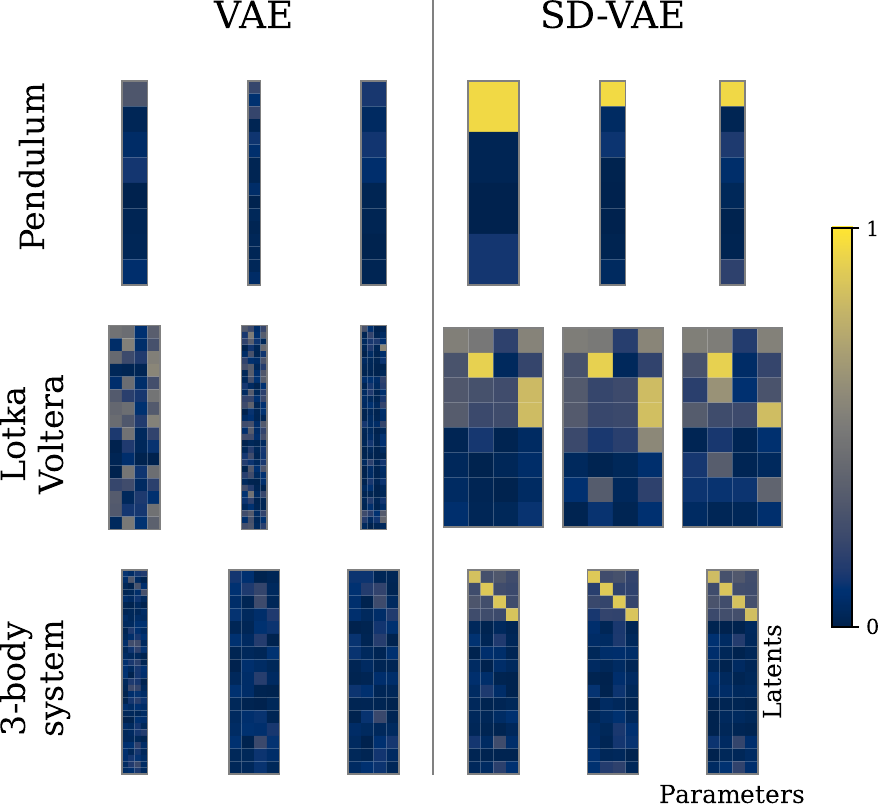}
\caption{Pearson correlation. All latents.}
\end{subfigure}
  \label{fig:correlation_latents_parameters}
\end{figure*}

To capture the relationship between latents $z$ and parameters $\xi$ we train boosted trees regressors with $z$ as predictors and $\xi$ as targets. The weights of the trained trees indicate whether there is a strong correlation between latents and real parameters. These results, depicted in \cref{fig:results_latents_importance_weights_truncated} and \cref{fig:correlation_latents_parameters} (a,b), show that supervised latents have very high predictive power for their respective real parameters. Nevertheless, trees can capture both linear and non-linear dependencies. It remains, unclear from this analysis if the relationship is linear or not. To futher clarify the relationship, we fit linear regression models between all $z_i$, $\xi_j$ pairs. For each fit we compute the absolute Pearson correlation coefficient $r$ that captures the linearity between the two variables. A value of $r$ close to $1$ denotes a highly linear correlation. Pearson $r$ values are visualized in \cref{fig:correlation_latents_parameters} (c,d). We also provide numerical values of $r$ for the supervised latents:

\begin{itemize}
    \item \textbf{\textbf{Pendulum}}: $r_l = 0.94$
\item \textbf{Lotka-Volterra }: $r_a=0.52, r_b=0.91, r_c=0.23, r_d=0.82$
\item \textbf{3-Body system}: $r_K=0.84, r_{m_1}=0.87, r_{m_2}=0.87, r_{m_3}=0.85$
\end{itemize}

These results show that the relationship between supervised latents and is highly linear in most cases. This aligns with our experimental findings that linear scaling works best for the disentanglement loss (see \cref{sec:scaling_parameters}). We furthermore, exploit this linearity to perform traversals of the latent space in \cref{sec:latent_space_traversals}.

\FloatBarrier

\subsection{Latent space traversals}
\label{sec:latent_space_traversals}

Being able to traverse between two point in the latent space and obtain a meaningful representation is a highly desirable property. Simple linear interpolation in latent space can produce meaningful images in properly disentangled VAEs \cite{Higgins2017Beta-VAE:FRAMEWORK}. While images have easily recognizable visual components, similar traversals for dynamical systems are harder to portray. Here we study whether interpolating between two points in the latent space of SD-VAE can produce meaningful trajectories. First, we create a new pendulum dataset containing 100 trajectories with linearly spaced pendulum length in the range  $l \in [1.0-1.5]$. The initial conditions are kept constant ($\theta=\frac{pi}{2}, \omega=0$) for all the trajectories to facilitate comparisons and we use the same noise level as in the training dataset. We use the encoder of our best SD-VAE model to extract the latent variables for each trajectory. For each trajectory the encoder produces 4 latent variables $z_1 \dots z_4$.  We interpolate between the latents of the the two extreme trajectories ($l=1.0$ and $l=1.5$). The interpolation we use is linear, driven by our findings that latents and parameters have a strongly linear correlation (\cref{sec:linearity_main_body} and \cref{sec:linear_correlation}). Next, we feed the real and interpolated latents to the decoder and predict up to 1000 timesteps. We find that the total mean absolute error between prediction and ground truth is $0.29$ with the real latents and $0.33$ with the interpolated one. These results indicate that linear latent space interpolation produces meaningful latent codes. This is further corroborated by plotting the real and interpolated latents together. As we can see in \cref{fig:autoregressive} the relationship between the real latents $z_i$ and pendulum length $l$ is highly linear, which further explains with the linear interpolation method works well.

\begin{figure*}[!htp]
\centering
\includegraphics[width=1.\hsize]{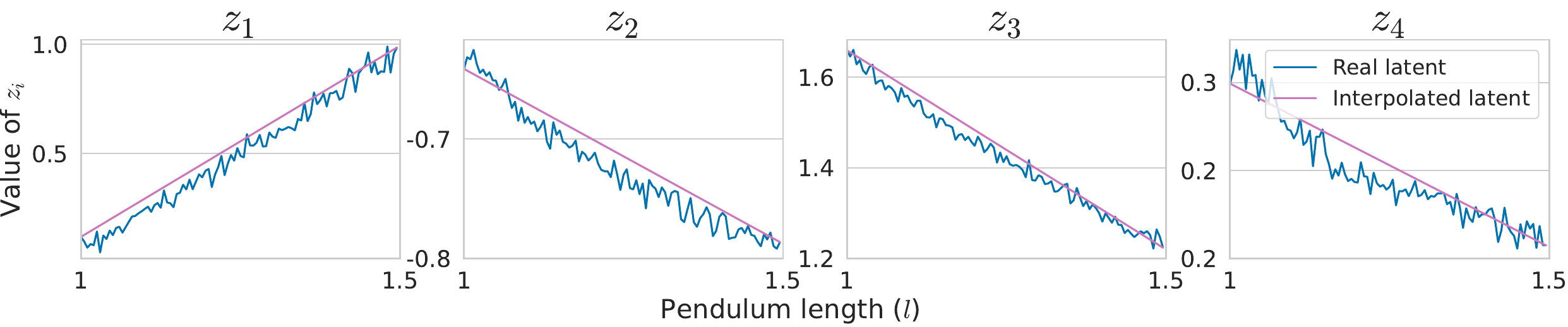}
\caption{\textbf{Traversals of the latent space of SD-VAE} on the pendulum system. We pass trajectories corresponding to different pendulum length through the encoder to obtain the real values of $z_i$. The interpolation is done between the latent of minimum($l=1.0$) and maximum($l=1.5$) length. For this experiment the initial conditions were kept constant $\theta=\frac{\pi}{2}, \omega=0$}
\label{fig:latent-traversal}
\end{figure*}

 \FloatBarrier

% \newpage
\section{Additional Results for Phase Space Experiments}
\label{sec:results_extra_observation_space}

\begin{table*}[!htb]
\centering
\caption{\textbf{Model comparison in observation-space pendulum.} Metrics are calculated between ground truth and prediction of the models at exactly 800 timesteps in the future.}
\label{tab:results_pixel_pendulum_800_ssim_psnr}
\begin{tabular}{lccc|ccc}
\toprule
& \multicolumn{3}{c}{SSIM}  & \multicolumn{3}{c}{PSNR} \\
&  Test-set &  OOD-Easy &  OOD-Hard &  Test-set &  OOD-Easy &  OOD-Hard \\
 \cmidrule(r){2-4} \cmidrule(r){5-7} 
RSSM         &  0.795 &0.787 &0.783 2 &13.36 &12.71 &12.26 \\
SD-RSSM     &\textbf{0.813 }&\textbf{0.808} &\textbf{0.794 }&\textbf{14.16} &\textbf{13.82} &\textbf{12.90} \\
\bottomrule
\end{tabular}
\end{table*}

\begin{figure*}[!htb]
  \centering
\includegraphics[width=\linewidth]{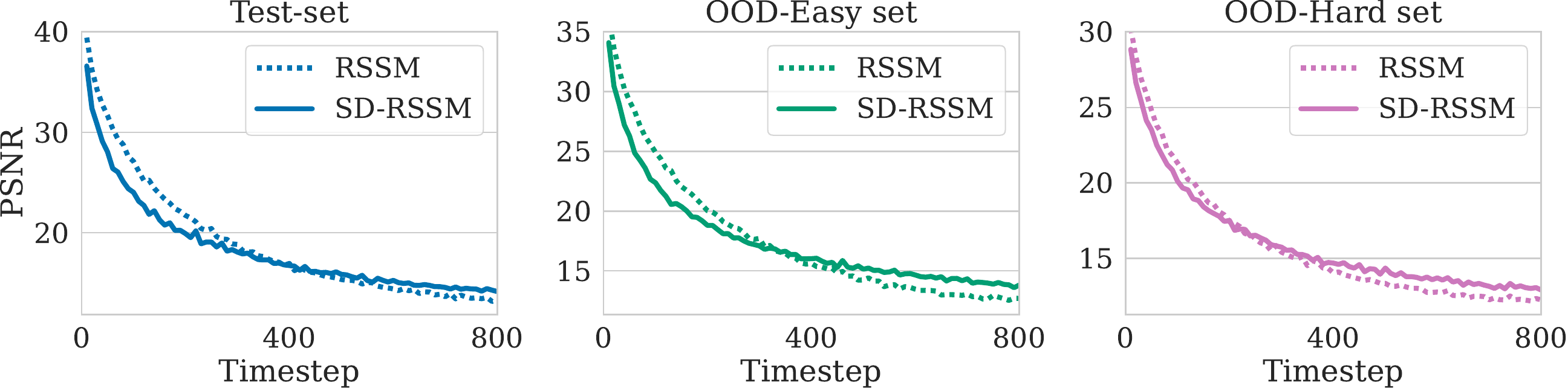}
 \caption{\textbf{Prediction quality on the observation-space pendulum.} PSNR as a function of the distance predicted into the future (x axis)} 
  \label{fig:results_rssm_psnr}
\end{figure*}

\begin{figure*}[!htp]
\centering
\includegraphics[width=0.94\hsize]{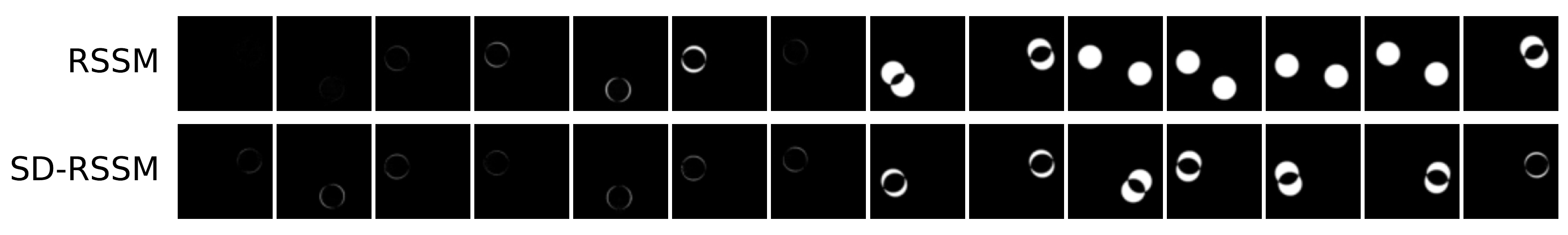}
\includegraphics[width=0.94\hsize]{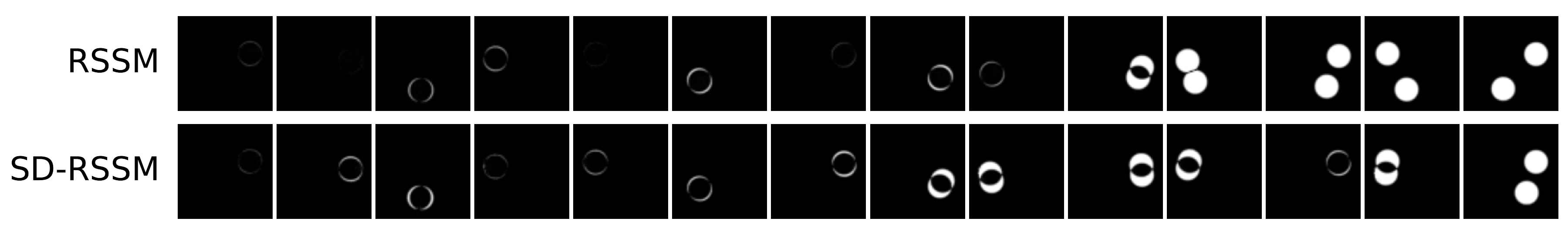}
\includegraphics[width=0.94\hsize]{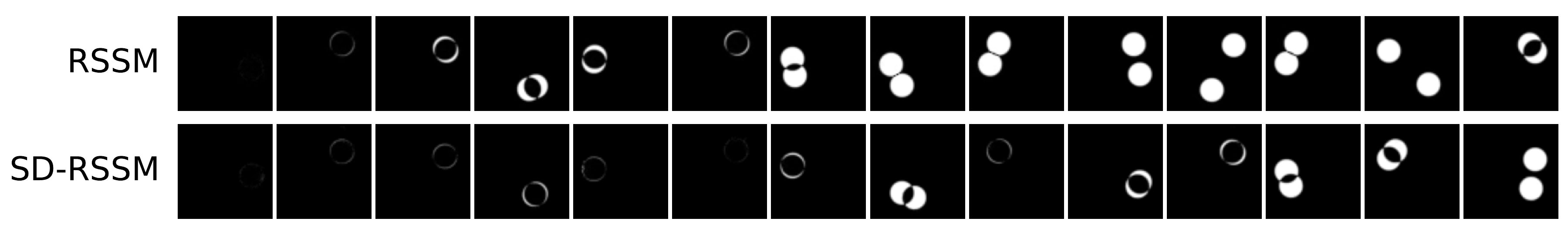}
\caption{\textbf{Samples of predicted trajectories.} Absolute difference between ground truth and predictions on the  test-set of the pendulum data set.}
\label{fig:qual-pendulum1}
\end{figure*}

\end{document}